\def\paperID{10769}
\def\confName{CVPR}
\def\confYear{2025}

\def\paperTitle{\TitleEmoji: Random Image Transformations as a Universal Anti-hallucination Lever in Large Vision Language Models\vspace{-4mm}}

\def\authorBlock{
    Sangmin Woo\thanks{Equal contribution} \qquad
    Jaehyuk Jang\footnotemark[1] \qquad
    Donguk Kim\footnotemark[1] \qquad
    Yubin Choi \qquad
    Changick Kim\\
    KAIST \\
    {\tt\small \{smwoo95, jhyuk, kdu3613, choibinbin, changick\}@kaist.ac.kr} \\
    {\tt\textbf{ Project: \url{https://sangminwoo.github.io/RITUAL/}}}
}

\newif\ifreview 
\newif\ifarxiv \newcommand{\arxiv}{\arxivtrue}
\newif\ifcamera 
\newif\ifrebuttal 

\arxiv
\pdfoutput=1
\documentclass[10pt,twocolumn,letterpaper]{article}
\ifreview \usepackage[review]{cvpr} \fi
\ifarxiv \usepackage[pagenumbers]{cvpr} \fi
\ifrebuttal \usepackage[rebuttal]{cvpr} \fi
\ifcamera \usepackage{cvpr} \fi

\usepackage{times}
\usepackage{epsfig}
\usepackage{float}
\usepackage{placeins}
\usepackage{stfloats}
\usepackage{enumitem}
\usepackage{tabularx}
\usepackage{xstring}
\usepackage{xspace}
\usepackage[hang,flushmargin]{footmisc}

\usepackage[utf8]{inputenc} % allow utf-8 input
\usepackage[T1]{fontenc}    % use 8-bit T1 fonts
\usepackage{url}            % simple URL typesetting
\usepackage{booktabs}       % professional-quality tables
\usepackage{amsfonts}       % blackboard math symbols
\usepackage{nicefrac}       % compact symbols for 1/2, etc.
\usepackage{microtype}      % microtypography
\usepackage{xcolor}         % colors
\usepackage{amsmath, amssymb, overpic, textpos}
\usepackage{graphicx}
\usepackage{multirow}
\usepackage{color, colortbl}
\usepackage{tabulary}
\usepackage{algorithmicx}
\usepackage{algpseudocode}
\usepackage{listings}
\usepackage{multicol}
\usepackage{xspace}
\usepackage{wrapfig}
\usepackage{graphbox}
\usepackage{arydshln}
\usepackage{algorithm}
\usepackage{adjustbox}
\usepackage{kotex}
\usepackage{caption}
\usepackage{subcaption}
\usepackage{setspace}
\usepackage{makecell}
\usepackage{tabularray}

\definecolor{citecolor}{HTML}{0071BC}
\definecolor{linkcolor}{HTML}{ED1C24}
\usepackage[pagebackref=true,breaklinks=true,colorlinks,citecolor=citecolor,linkcolor=linkcolor,bookmarks=false]{hyperref}

\usepackage{amsmath,amsfonts,bm}

\def\eqref#1{equation~\ref{#1}}
\def\1{\bm{1}}

\def\vs{{\bm{s}}}

\DeclareMathAlphabet{\mathsfit}{\encodingdefault}{\sfdefault}{m}{sl}
\SetMathAlphabet{\mathsfit}{bold}{\encodingdefault}{\sfdefault}{bx}{n}

\usepackage[capitalize]{cleveref}
\crefname{section}{Sec.}{Secs.}
\Crefname{section}{Section}{Sections}
\Crefname{table}{Table}{Tables}
\crefname{table}{Tab.}{Tabs.}
\usepackage{pifont}
\newcommand{\cmark}{\ding{51}}
\newcommand{\xmark}{\ding{55}}

\newcommand*\rot{\rotatebox{90}}

\makeatletter
\DeclareRobustCommand\onedot{\futurelet\@let@token\@onedot}
\def\@onedot{\ifx\@let@token.\else.\null\fi\xspace}

\def\eg{\emph{e.g}\onedot}

\def\etc{\emph{etc}\onedot} \def\vs{\emph{vs}\onedot}

\makeatother
\newlength\savewidth

\newcolumntype{x}[1]{>{\centering\arraybackslash}p{#1pt}}
\newcolumntype{y}[1]{>{\raggedright\arraybackslash}p{#1pt}}
\newcolumntype{z}[1]{>{\raggedleft\arraybackslash}p{#1pt}}

\newlength\secmargin
\newlength\subsecmargin
\newlength\subsubsecmargin
\newlength\paramargin
\newlength\abovetabcapmargin
\newlength\belowtabcapmargin
\newlength\abovefigcapmargin
\newlength\belowfigcapmargin
\definecolor{Green}{rgb}{0.2, 0.7, 0.1}
\definecolor{Gray}{HTML}{B0B0B0} % A9A9A9
\definecolor{Blue}{HTML}{4A90E2} % 69A7D2
\definecolor{Yellow}{HTML}{FFA700} % CCCC00 E5E500 FFD700
\definecolor{prompt_blue}{HTML}{1f78b4}
\definecolor{prompt_red}{HTML}{d45c43}
\definecolor{plus}{HTML}{0071bc}
\definecolor{minus}{RGB}{153,10,10}
\definecolor{SecondBest}{HTML}{E0F0FA}
\definecolor{Best}{HTML}{BAD8F2}

\newcommand{\up}{\bf \fontsize{10}{42} \color{plus}{$\uparrow$}}
\newcommand{\down}{\bf \fontsize{10}{42}\selectfont \color{minus}{$\downarrow$}}

\newcommand\ritualtitlefont[1]{\smash{{\usefont{T1}{cinzeldecorativebold}{m}{n}#1}}}
\newcommand\ritualfont[1]{\smash{{\usefont{T1}{cinzeldecorative}{m}{n}#1}}}
\newcommand{\ritualtitle}{\ritualtitlefont{RITUAL}}
\newcommand{\ritual}{\ritualfont{RITUAL}}
\newcommand{\ritualplus}{\ritualfont{RITUAL+}}
\newcommand{\ritualplustitle}{\ritualtitlefont{RITUAL+}}

\newcommand{\Title}{\ritualtitle\xspace}
\newcommand{\Ours}{\ritual\xspace}
\newcommand{\Oursplus}{\ritualplus\xspace}
\newcommand{\TitlePlus}{\ritualplustitle\xspace}
\newcommand{\Emoji}{\raisebox{-0.2em}{\includegraphics[height=1em]{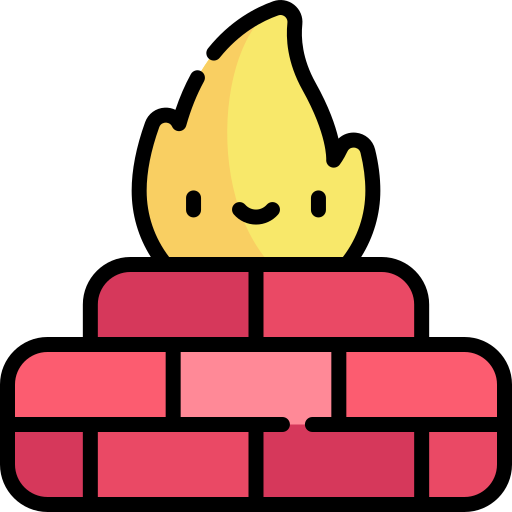}}}
\newcommand{\TitleEmoji}{\Emoji\xspace\Title}

\newcommand{\OursEmoji}{\Emoji\xspace\Ours}

\ifcamera \usepackage[accsupp]{axessibility} \fi

% ...

\ifarxiv  \fi

\newcommand{\R}[1]{{%
    \textbf{%
        \ifstrequal{#1}{1}{\textcolor{red}{R#1}}{%
        \ifstrequal{#1}{2}{\textcolor{blue}{R#1}}{%
        \ifstrequal{#1}{3}{\textcolor{magenta}{R#1}}{%
        \ifstrequal{#1}{4}{\textcolor{teal}{R#1}}{%
                           \textcolor{cyan}{R#1}%
        }}}}%
    }%
}}
\usepackage{xr-hyper}

\makeatletter
\newcommand*{\addFileDependency}[1]{
  \typeout{(#1)}
  \@addtofilelist{#1}
  \IfFileExists{#1}{}{\typeout{No file #1.}}
}

\makeatother
\newcommand*{\myexternaldocument}[1]{
    \externaldocument{#1}
    \addFileDependency{#1.tex}
    \addFileDependency{#1.aux}
}

\definecolor{cvprblue}{rgb}{0.21,0.49,0.74}
\usepackage[capitalize]{cleveref}
\crefname{section}{Sec.}{Secs.}
\crefname{table}{Table}{Tables}
\crefname{figure}{Fig.}{Figs.}

\ifarxiv \crefname{appendix}{App.}{Apps.}
\else \crefname{appendix}{Suppl.}{Suppls.} \fi
\unless\ifarxiv \myexternaldocument{_supplementary} \fi

\begin{document}
%% TITLE
\title{\paperTitle}
\author{\authorBlock}
\maketitle

\begin{abstract}
Recent advancements in Large Vision Language Models (LVLMs) have revolutionized how machines understand and generate textual responses based on visual inputs, yet they often produce "hallucinatory" outputs that misinterpret visual information, posing challenges in reliability and trustworthiness.
We propose \TitleEmoji, a simple decoding method that reduces hallucinations by leveraging randomly transformed images as complementary inputs during decoding, adjusting the output probability distribution without additional training or external models.
% \Ours combines them with the original image to adjust the probability distribution of generated outputs, operates without additional training or external models.
%
Our key insight is that random transformations expose the model to diverse visual perspectives, enabling it to correct misinterpretations that lead to hallucinations. Specifically, when a model hallucinates based on the original image, the transformed images---altered in aspects such as orientation, scale, or color---provide alternative viewpoints that help recalibrate the model's predictions. By integrating the probability distributions from both the original and transformed images, \Ours effectively reduces hallucinations.
% This method not only enriches the model’s decision-making process but also enhances robustness without compromising performance on the original task.
%
To further improve reliability and address potential instability from arbitrary transformations, we introduce \TitlePlus, an extension that selects image transformations based on self-feedback from the LVLM.
Instead of applying transformations randomly, \Oursplus uses the LVLM to evaluate and choose transformations that are most beneficial for reducing hallucinations in a given context. This self-adaptive approach mitigates the potential negative impact of certain transformations on specific tasks, ensuring more consistent performance across different scenarios.
%
% Experiments demonstrate that \Ours and \Oursplus significantly outperform existing contrastive decoding methods across several object hallucination benchmarks.
Experiments demonstrate that \Ours and \Oursplus significantly reduces hallucinations across several object hallucination benchmarks.
% , including POPE, CHAIR, and MME.
% Moreover, our method shows consistent improvements across multiple LVLM architectures, highlighting its broad applicability and practical value.
% By providing a straightforward, training-free solution that enhances both accuracy and reliability, \Ours advances the development of more trustworthy LVLMs and offers a valuable tool for real-world applications where dependability is crucial.
\end{abstract}

\addtocontents{toc}{\protect\setcounter{tocdepth}{0}}
\vspace{\secmargin}\section{Introduction}\vspace{\secmargin}
\label{sec:intro}

\begin{figure}[t!]
    \centering
    \vspace{-1mm}
    \includegraphics[width=\linewidth]{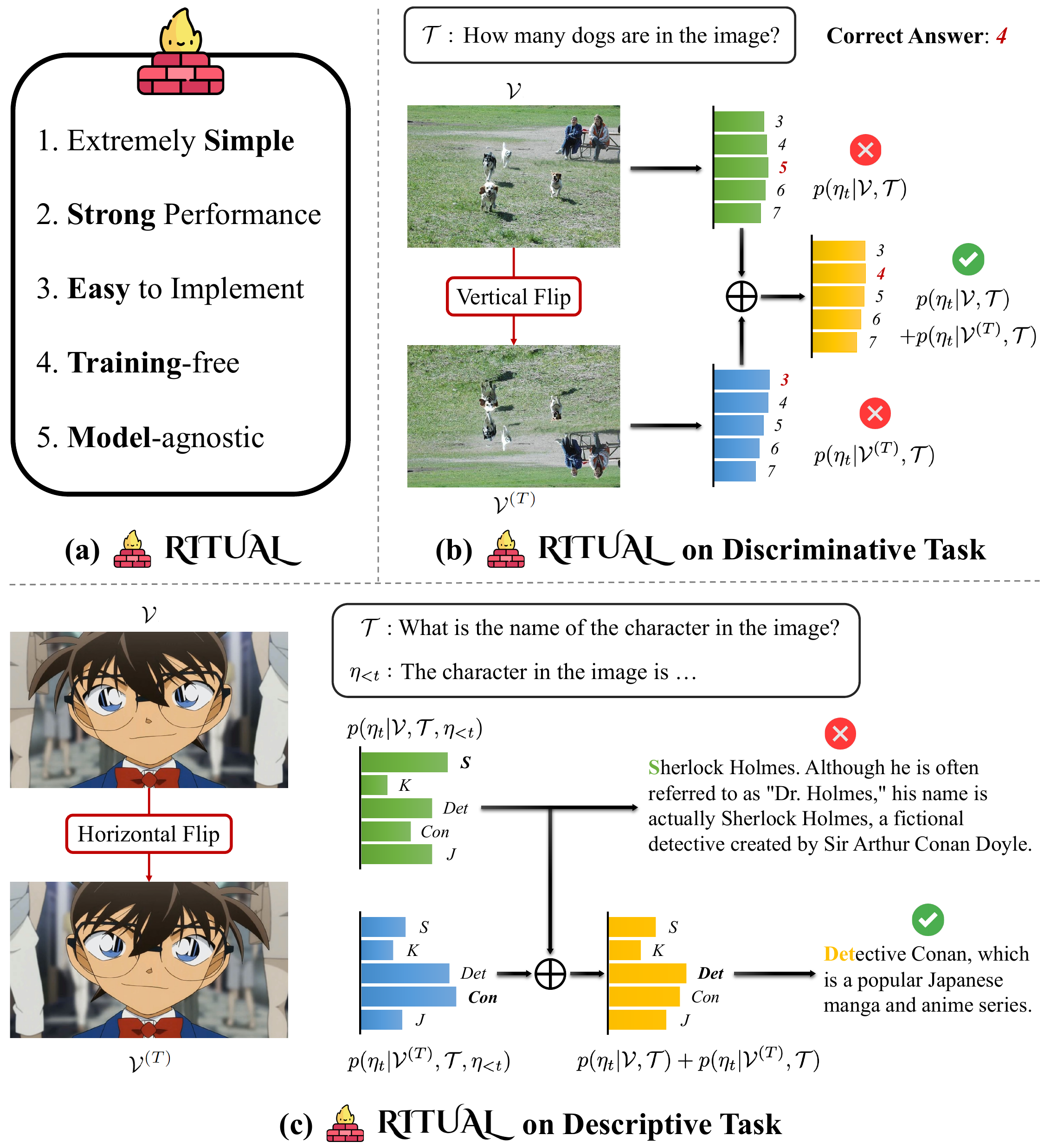}
    \vspace{\abovefigcapmargin}
    \caption{
    \textbf{\TitleEmoji: A simple yet effective anti-hallucination approach for LVLMs.}
    Our \Ours method leverages basic image transformations (\eg, vertical and horizontal flips) to enhance LVLM accuracy without external models or training.
    By integrating transformed and original images, \Ours significantly reduces hallucinations in both discriminative tasks and descriptive tasks.
    Using both versions together enables the model to refine predictions, reducing errors and boosting correct responses.
    }
    \label{fig:teaser}
    \vspace{\belowfigcapmargin}
\end{figure}
Large Vision-Language Models (LVLMs)~\citep{dai2024instructblip,zhu2023minigpt,liu2023visual,liu2023improved,bai2023qwen} have emerged as a pivotal technology, enabling machines to interpret complex visual scenes and generate contextually appropriate textual descriptions.
These models integrate and process inputs from both visual and linguistic domains, offering unprecedented possibilities in applications ranging from video content creation~\citep{videoworldsimulators2024} to assistive technologies~\citep{team2023gemini,OpenAI2023}.

% Despite their potential, LVLMs are often criticized for generating "hallucinatory" content---outputs that appear plausible but do not faithfully reflect the visual inputs.
Despite these advancements, LVLMs still face a fundamental challenge: the tendency to produce "hallucinations"~\citep{li2023evaluating,zhao2023beyond,wang2023llm,huang2023opera}---outputs that are inconsistent with the actual content of the visual input.
This gap in reliability and trustworthiness is particularly concerning for sensitive applications such as medical diagnosis~\citep{zhou2023survey, liu2023qilin}, surveillance~\citep{wu2024vadclip,hasan2024vision}, and autonomous driving~\citep{li2024automated}.

% Related Work (Broad)

Hallucinations in LVLMs often arise due to the model’s overreliance on certain visual cues or its inability to generalize effectively across diverse visual perspectives.
% Traditional approaches to mitigate hallucinations typically involve complex training regimes, large-scale data augmentation, or external contrastive models. However, these solutions introduce considerable overhead in terms of computational resources and may not generalize well across different LVLM architectures or application domains. This raises the need for a lightweight and architecture-agnostic approach that can effectively reduce hallucinations without necessitating retraining or additional external models.
% The challenge primarily arises from the difficulty in maintaining alignment between the visual inputs and textual outputs, given the complexity of training such models to accurately interpret and narrate visual data.
% Although several strategies have been developed to mitigate these issues, they often
Existing approaches to mitigate hallucinations often require complex training regimes~\citep{jiang2023hallucination, zhou2023analyzing, gunjal2023detecting,liu2023mitigating,sun2023aligning,wang2023vigc,yin2023woodpecker,lu2024evaluation, zhai2024hallecontrol,yue2024more}, sophisticated feedback mechanisms~\citep{yin2023woodpecker, yu2023rlhf, kim2024if, sun2023aligning}, or reliance on auxiliary models~\citep{zhao2024mitigating, wan2024contrastive, deng2024seeing, yang2024pensieve, li2023visual}, which can complicate deployment and scalability.

% Related Work (Contrastive Decoding)
% Contrastive decoding~\citep{leng2023mitigating,favero2024multi, wang2024mitigating}, a notable approach in this context, refines the model outputs by contrasting the conditional probability of textual responses given the original visual input versus a distorted visual input.
% This aims to alleviate language biases or statistical priors, ensuring that responses are more grounded in the actual images, thereby reducing deviations from the visual truth.
% While beneficial, contrastive decoding does not fully resolve the misalignments between visual data and textual descriptions and can sometimes lead to the reinforcement of incorrect patterns.

% Our Proposal
% We present a simple, training-free approach termed \Ours, which leverages random image transformations to complement the original probability distributions of textual responses.
% Our method is distinct from contrastive decoding~\citep{leng2023mitigating,favero2024multi, wang2024mitigating}, which attributes the causes of hallucinations to language bias or statistical priors.
% Instead, \Ours suggests that the source of hallucinatory content might actually reside within the images themselves, advocating for a multifaceted view of visual inputs.
% The conceptual comparison is shown in~\cref{fig:teaser}.
We present a simple, training-free approach termed \TitleEmoji, which leverages random image transformations to complement the original image and enhance models' robustness (see~\cref{fig:teaser}).
% \Ours is designed to address the issue of visual hallucination by employing a dual-input strategy that integrates both the original and a randomly transformed image.
% The final prediction is an ensemble of the individual predictions generated from both the original and augmented images.
Our core insight is that by exposing the model to diverse visual transformations---such as changes in orientation, scale, and color---during decoding, it can better discern the true contents of the original image and reduce the likelihood of generating hallucinatory outputs.
Specifically, \Ours introduces these transformed images as complementary inputs during the decoding process, allowing the LVLM to adjust its output probability distribution by integrating alternative visual perspectives.
\Ours employs a dual-input strategy that integrates both the original and a randomly transformed image, and the final prediction is an ensemble of the individual predictions generated from both the original and augmented images.
% Importantly, these image transformations are applied only during the inference phase, not during training.
This simple yet effective approach does not require additional training or external models and is readily compatible with existing LVLMs. 
To further enhance reliability, we propose \TitlePlus, an adaptive extension of \Ours that leverages self-feedback from the LVLM to guide the selection of transformations.
Rather than applying random transformations indiscriminately, \Oursplus employs the LVLM itself to evaluate and choose the transformations that are most likely to mitigate hallucinations in a specific context.
This self-adaptive mechanism mitigates the potential for detrimental transformations, which may inadvertently introduce instability in the model’s predictions, ensuring that our method performs consistently across a range of tasks and scenarios.

% Experimental Findings
Our experiments evaluate \Ours and \Oursplus across several benchmarks, including POPE~\citep{rohrbach2018object}, CHAIR~\citep{li2023evaluating}, and both MME-Hallucination and MME-Fullset~\citep{fu2024mme}.
Despite its simplicity, our approach effectively reduces hallucination across these benchmarks, significantly enhancing the general capabilities of LVLMs.
Moreover, \Ours and \Oursplus consistently outperform existing contrastive decoding baselines~\citep{leng2023mitigating,favero2024multi, chuang2023dola} on all tested benchmarks, achieving superior performance with comparable latency.
\vspace{\secmargin}\section{Related Work}\vspace{\secmargin}
\label{sec:related_work}

% \jh{
% All of the LVLMs mentioned above are affected by hallucination issues, and we conduct experiments on the representative models, LLAVA and InstructBLIP.
% }

\vspace{\paramargin}\paragraph{Hallucinations in LVLMs.}
LVLMs are susceptible to visual hallucinations, in which the generated text descriptions include objects or details entirely irrelevant from the given image.
A range of methods has been introduced to address the issue by additional training~\citep{gunjal2023detecting,liu2023mitigating,sun2023aligning,wang2023vigc,yin2023woodpecker,lu2024evaluation,jiang2023hallucination, zhou2023analyzing, zhai2024hallecontrol,yue2024more}.
While these approaches offer promise, they often face practical limitations due to their dependence on additional data and extensive training periods.
In response to these limitations, training-free approaches have gained traction.
These models aim to refine the model output by self-feedback correction~\citep{lee2023volcano, yin2023woodpecker}, providing additional knowledge using auxiliary models~\citep{wan2024contrastive, deng2024seeing, zhao2024mitigating, yang2024pensieve, kim2024if},
and contrastive decoding~\citep{leng2023mitigating,favero2024multi, zhang2024debiasing,wang2024mitigating}, which refines the model outputs by contrasting the conditional probability of textual responses given the original visual input versus a distorted visual input.
Our work adopts a unique approach by applying random image transformations to complement the original image.
This provides a wide range of visual contexts, aiming to mitigate hallucinatory visual explanations without the complexities of extra models, additional training, or data requirements.

\vspace{-1mm}

\vspace{\paramargin}\paragraph{Image augmentations for model robustness.}
Image augmentations~\citep{shorten2019survey,perez2017effectiveness} have long been recognized as a crucial technique for improving model robustness, particularly in computer vision and multimodal tasks.
By introducing variations in input data, augmentations help models generalize better to unseen scenarios, reduce overfitting, and improve performance in the presence of noise or ambiguous inputs.
In the training phase, data augmentation techniques~\citep{cubuk2018autoaugment,taylor2017improving}, such as those used in SimCLR~\citep{chen2020simple} and BYOL~\citep{grill2020bootstrap}, enhance the diversity of training data by applying transformations like rotations, flips, and crops.
This encourages the model to learn more generalizable features, improving performance on unseen data.
At inference time, test-time augmentation (TTA)~\citep{zhang2022memo, shanmugam2021better,perez2021enhancing} further improves model robustness. TTA applies multiple transformations to the input image during testing, generating varied predictions which are then averaged or ensembled to produce a more reliable output.
By exposing the model to diverse perspectives of the same input, TTA reduces sensitivity to noise and ambiguity, stabilizes predictions on difficult cases, and serves as a cost-effective ensembling method without requiring additional model training.
Our approach builds on these concepts by using random image transformations during inference to provide a broader visual context, reducing hallucinations in vision-language models. By combining predictions from both the original and transformed images, our method enhances robustness.

\vspace{-1mm}

\vspace{\secmargin}\section{Approach: \TitleEmoji} 
\begin{figure}[t!]
    \centering
    \vspace{-1mm}
    \includegraphics[width=\linewidth]{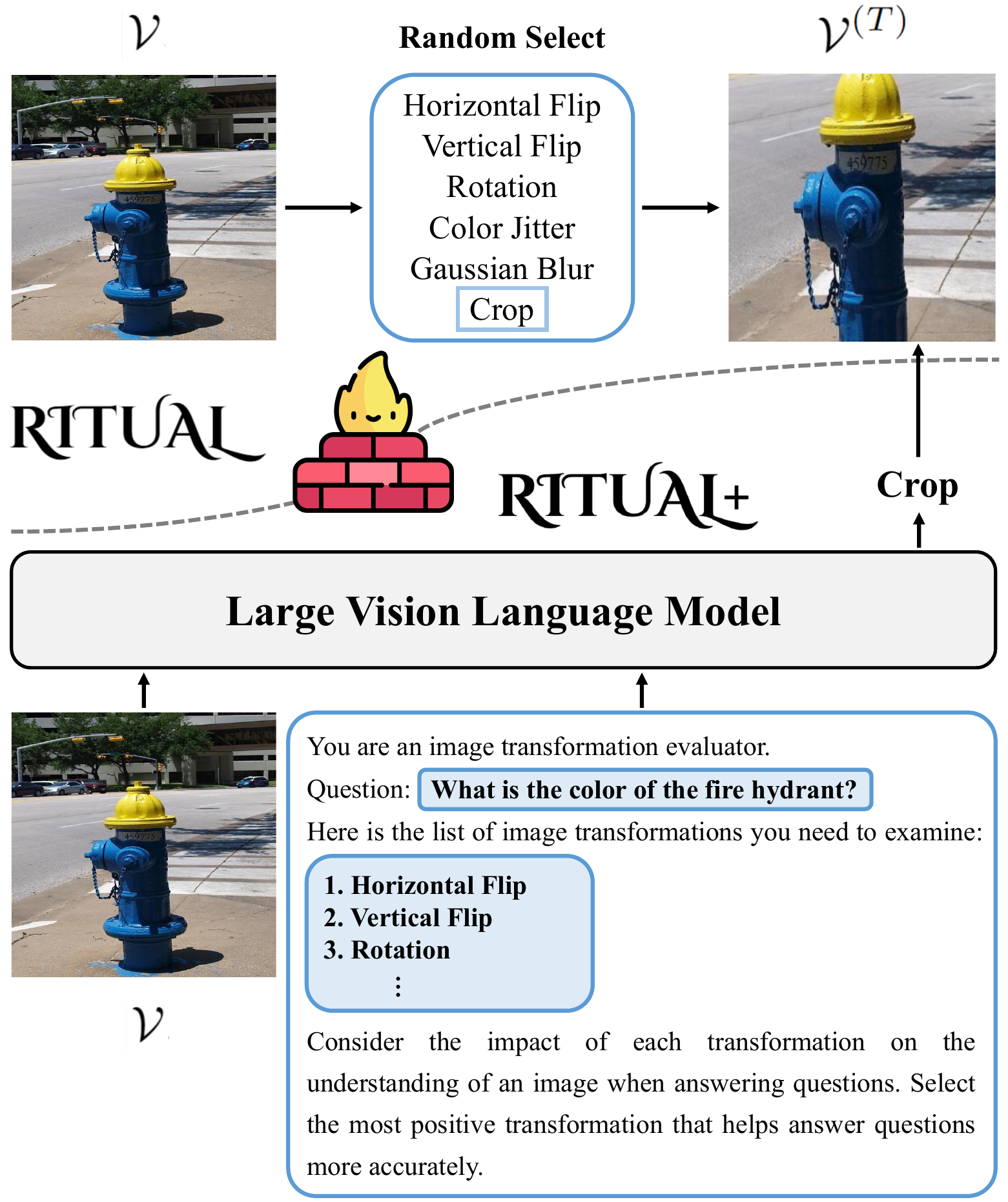}
    \vspace{\abovefigcapmargin}
    \caption{
    \textbf{Overview of \TitleEmoji and \TitlePlus.}
    In \Ours, the original image $\mathcal{V}$ undergoes  random transformations, generating a transformed image $\mathcal{V}^{(T)}$.
    In \Oursplus, the model evaluates various potential transformations and selects the most beneficial one to improve answer accuracy within the given context, further refining reliability.
    These transformed images serve as complementary inputs, enabling the model to incorporate multiple visual perspectives to reduce hallucinations.
    % At each timestep $t$, LVLM auto-regressively samples a response $\eta_t$ given a visual input, a textual query $\mathcal{T}$, and previously generated tokens $\eta_{<t}$.
    % from a multinomial distribution \todo{(equation refer)}.
    % When conditioned on the original image $\mathcal{V}$, the probabilities for \textcolor{blue}{Blue} (\textit{correct}) and \textcolor{red}{Red} (\textit{hallucinated}) responses are similar, which can lead to the hallucinated response being easily sampled.
    % \Ours leverages an additional probability distribution conditioned on the transformed image $\mathcal{V^{(T)}}$, where the likelihood of hallucination is significantly reduced.
    % Consequently, the response is sampled from a linear combination of the two probability distributions, ensuring more accurate and reliable outputs.
    % \todo{ritual plus 추가}
    % Note that LVLMs are not explicitly trained on the transformed images.
    }%
    \label{fig:overview}
    \vspace{\belowfigcapmargin}
\end{figure}

% \todo{and \Oursplus}}
\vspace{\secmargin}
\label{sec:approach}
We present a simple yet effective decoding method that is training-free and operates without the need for external models.
% that can be applied in an online manner during token generation.
% Our method is training-free, does not require external models
% or a costly self-feedback mechanism, and remains compatible with existing contrastive decoding techniques~\citep{leng2023mitigating,favero2024multi}.
An overview of our method is illustrated in~\cref{fig:overview}.

\vspace{\subsecmargin}\subsection{LVLM Formulation}
% We describe the generative process for Language-Vision Language Models (LVLMs) through three phases: vision-language alignment, model forwarding, and response generation.

\vspace{\paramargin}\paragraph{Vision-Language Alignment.}
LVLM takes a visual input and a textual query as inputs, where the visual input provides contextual visual information to assist the model in generating a relevant response to the textual query. 
Initially, a vision encoder (\eg, ViT~\citep{dosovitskiy2020image}, CLIP~\citep{radford2021learning}, \etc) processes raw images to extract visual features.
These features are then projected into the language model’s input space using a vision-language alignment module (\eg, Q-Former~\citep{li2023blip}, linear projection~\citep{liu2023visual}, \etc), resulting in a set of visual tokens, $\mathcal{V} = \{\nu_0, \nu_1, \dots, \nu_{N-1}\}$.
Concurrently, the textual inputs are tokenized into $\mathcal{T} = \{\tau_N, \tau_{N+1}, \dots, \tau_{N+M-1}\}$.
The visual and textual tokens are concatenated to form an input sequence of length $N+M$.
% \begin{equation}
% [ \mathcal{V}, \mathcal{T} ] = \{ \eta_i \}_{i=0}^{L-1},
% \end{equation}
% where $L = N + M$ and $[\cdot, \cdot]$ denotes concatenation.

\vspace{\paramargin}\paragraph{Model Forwarding.}
The LVLM, parametrized by $\theta$, processes the concatenated sequence of visual and textual tokens.
This process is formalized as:
\begin{equation}
\begin{aligned}
\mathcal{H} &= \mathrm{LVLM}_{\theta}([\mathcal{V}, \mathcal{T}]),
\label{eq:lvlm}
\end{aligned}
\end{equation}
where $\mathcal{H}$ denotes the sequence of output hidden states from the final layer of LVLM.
These hidden states $\mathcal{H}$ are used to compute the logits (or probabilities) for predicting the next tokens.

\vspace{\paramargin}\paragraph{Response Generation.}
The LVLM generates responses auto-regressively, employing a causal attention mask to ensure each subsequent token is predicted based solely on the preceding tokens.
Each response token is generated by sampling from the following probability distribution:
\begin{equation}
\eta_t \sim p_{\theta}(\eta_t | \mathcal{V}, \mathcal{T}, \eta_{<t}).
\label{eq:original_sampling}
\end{equation}
where $\eta_t$ denotes the response token being generated at timestep $t$, and $\eta_{<t}$ indicates the sequence of tokens generated up to timestep $(t - 1)$.
This generative process is iteratively continued, appending each newly predicted token to the sequence, until the termination of the sequence.
% By default, Greedy Decoding is used.
By default, standard multinomial sampling is used.
Alternatively, decoding strategies such as Beam Search~\citep{wiseman2016sequence}, Nucleus Sampling~\citep{holtzman2019curious}, or DoLa~\citep{chuang2023dola} can be employed.

% \begin{wrapfigure}[8]{r}{0.6\textwidth}
%     \centering
%     \vspace{-35pt}
%     \includegraphics[width=\linewidth]{figures/overview.pdf}
%     % \vspace{-7pt}
%     \caption{
%     \textbf{Overview of \TitleEmoji.}
%     }
%     \label{fig:overview}
%     \vspace{\belowfigcapmargin}
% \end{wrapfigure}

% 매 timestep마다 각각의 prob dist가 업데이트 되고 이 multinomial distribution에서 각 response token을 sampling함 .
% original image에 condition되면 blue와 red가 비슷함.
% transformed image에 condition되면 hallucination 확률을 확 줄임.
% 두 probability distribution을 equation에 따라 combine함 (equation refer).
% \jh{logit scale 수정함, 원래 red랑 yellow or blue랑 비슷한 정도로 바꾸고 pdf}

\vspace{\subsecmargin}\subsection{Anti-hallucinating LVLMs with \Title}
Visual hallucinations in LVLMs can occur during the decoding phase when tokens are selected based on erroneous probability distributions that do not align with the visual inputs.
Our approach aims to mitigate these visual hallucinations with a simple yet effective modification to the input handling.

\Ours first randomly apply common image transformations (\eg, Crop, Flip, Rotate, \etc) to the original visual input $\mathcal{V}$,
This results in a transformed version of the visual input, $\mathcal{V}^{(T)}$.
\begin{equation}
\mathcal{V}^{(T)} = {T}(\mathcal{V}; \omega), \text{ where } \omega \in \Omega.
\label{eq:image_transformation}
\end{equation}
Here, $T$ represents a specific transformation function selected randomly from a set of image transformations.
The parameter $\omega$ represents the specific parameters of the transformation, drawn from a distribution $\Omega$ that governs the selection and nature of the transformation applied.
%
% \begin{equation}
% \eta_t \sim p_{\theta}(\eta_t | \mathcal{V}^{(T)}, \mathcal{T}, \eta_{<t}).
% \label{eq:transformed_sampling}
% \end{equation}
% However, using $\mathcal{V}^{(T)}$ alone rather harms the performance.

% During the decoding phase, rather than using \(\mathcal{V}^{(T)}\) alone---which we found to impair performance---we utilize both the original and transformed images.
During the decoding phase, we utilize both the original and transformed images.
% This dual-input approach significantly reduces the likelihood of hallucinatory outputs, as illustrated in~\cref{fig:motivation}\todo{fig 1로 수정?}, and improves the accuracy of the model’s predictions.
The sampling equation in~\cref{eq:original_sampling} is updated as follows:
\begin{equation}
\eta_t \sim p_{\theta}(\eta_t | \mathcal{V}, \mathcal{T}, \eta_{<t}) + \alpha p_{\theta}(\eta_t | \mathcal{V}^{(T)}, \mathcal{T}, \eta_{<t}).
\label{eq:ritual_sampling}
\end{equation}
% \begin{align}
% &p(\eta_t | \mathcal{V}, \mathcal{T}, \eta_{<t}) \\
% + &p(\eta_t | \mathcal{V}^{(T)}, \mathcal{T}, \eta_{<t})
% \end{align}
% \begin{align}
% &p(\eta_t | \mathcal{V}, \mathcal{T}) \\
% + &p(\eta_t | \mathcal{V}^{(T)}, \mathcal{T})
% \end{align}
% \begin{align}
% p(\eta_t | \mathcal{V}, \mathcal{T}) + p(\eta_t | \mathcal{V}^{(T)}, \mathcal{T})
% \end{align}
% \begin{align}
% \mathcal{T}:\\
% \eta_{<t}:
% \end{align}
Here, $\alpha$ is a balancing hyperparameter, adjusting the contribution of the transformed input relative to the original.

% In practice, we
\paragraph{Image transformations.}
We employ a predefined set of image transformations to enhance model robustness, divided into geometric and appearance transformations.
Geometric transformations, such as flipping, small random rotations, and cropping, simulate different viewing angles, orientations, and focus areas, enhancing the model’s ability to generalize across varied perspectives and object positioning.
Appearance transformations, including color jitter and Gaussian blur, adjust brightness, contrast, and saturation to account for lighting variations and sensor noise, increasing resilience to image imperfections.
Together, these transformations introduce meaningful variations that better prepare the model for real-world image scenarios, improving its flexibility and performance.

\vspace{\subsecmargin}\subsection{Adaptive Transformation Selection: \TitlePlus}
% \begin{equation}
% T \sim p_{\theta}(\eta | \mathcal{V}, \mathcal{T}, \textrm{Prompt}).
% \end{equation}
% \todo{식 불완전. 수정}
Despite the diverse views offered by random transformations by \Ours, the effectiveness of each transformation varies depending on the image, query, and task.
\Cref{tab:ritual_limitation} summarizes the performance of \Ours when employing individual augmentations.
% Notably, cropping hinders performance in counting tasks, while rotations and flips adversely affect the model’s ability to infer positional relationships between objects.
Gaussian Blur and Horizontal Flip improve counting and existence tasks, transformations like Crop degrade counting accuracy, and flips or rotations disrupt positional understanding. Color Jitter also negatively affect color-related tasks, while Gaussian Blur and Crop enhance them.

To further enhance reliability and address these inconsistencies, we propose \Oursplus, a self-adaptive extension of \Ours.
\Oursplus leverages LVLM self-feedback to evaluate the impact of each transformation within the specific context of the image and query.
Instead of relying on random augmentation, it selects transformations that are most effective in minimizing hallucinations and enhancing task-specific performance.
By dynamically tailoring augmentations to the requirements of the task, \Oursplus mitigates negative effects, such as the disruption of positional understanding or feature distortions, and ensures more robust and consistent results across diverse scenarios.\footnote{More details about \Oursplus are in Appendix.}

\begin{table}[t!]
\centering
\caption{
    \textbf{Impact of individual image transformations across various tasks on the MME-Hallucination benchmark~\cite{fu2024mme}.}
    Each transformation demonstrates varying degrees of effectiveness across different tasks, suggesting the need to carefully select transformations based on the specific image and task requirements.
}
\vspace{\abovetabcapmargin}
\scalebox{0.8}{
\begin{tabular}{clcccc}
    \toprule
    \multicolumn{2}{c}{\multirow{2}{*}{Method}} & \multicolumn{4}{c}{LLaVA 1.5~\cite{liu2023visual}}\\
    \arrayrulecolor{gray} \cmidrule(lr){3-6}
    & & Existence & Count  & Position & Color  \\
    \midrule
    \multicolumn{2}{l}{\textit{base}}          & 190.00    & 140.00 & 120.00   & 160.00 \\
    \arrayrulecolor{gray} \cmidrule(lr){1-6}
    \multirow{6}{*}{\rot{\textit{Transformation}}} & + Color Jitter    & 190.00\hspace{1mm}    & 130.00~\down & 126.67\up   & 143.33~\down \\
    & + Crop             & 190.00    & 123.33~\down & 128.33\up   & 170.00\up \\
    & + Gaussian Blur   & 195.00\up    & 146.67\up & 123.33\up   & 170.00\up \\
    & + Horizontal Flip & 195.00\up   & 158.33\up & 111.67~\down   & 165.00\up \\
    & + Rotation         & 190.00\hspace{1mm}    & 141.67\up & 116.67~\down   & 165.00\up \\
    & + Vertical Flip   & 190.00\hspace{1mm}   & 140.00\hspace{1mm} & 115.00~\down   & 160.00\hspace{1mm} \\
    \bottomrule
\end{tabular}
}
\vspace{\belowtabcapmargin}
\vspace{-2mm}
\label{tab:ritual_limitation}
\end{table}
\begin{table*}[ht!]
    \vspace{\abovetabcapmargin}
    \caption{
        \textbf{Results on POPE~\citep{li2023evaluating} benchmark.}
        \Ours consistently outperforms the contrastive decoding baselines: VCD~\cite{leng2023mitigating}, M3ID~\cite{favero2024multi}, and DoLa~\cite{chuang2023dola}.
        \Oursplus employs standard decoding but achieves performance comparable to OPERA~\cite{huang2023opera}, which uses beam search.
        \textit{Note: All baseline methods were reimplemented within our evaluation setup for fair comparison.}
        % Moreover, \Ours is shown to be compatible with both VCD and M3ID, leading to further performance improvements in most configurations.
        % \textbf{We reimplemented all baseline methods and produced results within our evaluation setup.}
        % As our approach is sampling-based, numerical differences from the original LLaVA~\cite{} may occur.
        % \jh{sampling 기반이라 original llava와 수치가 다를수 있다고 언급?}
    }
    \vspace{\belowtabcapmargin}
    \centering
    \small
    \setlength{\tabcolsep}{5pt} % base value: 6pt
    \scalebox{0.65}{
    \begin{tabular}{y{20}y{50}y{70}x{37}x{37}x{37}x{37}x{37}x{37}x{37}x{37}x{37}x{37}x{37}x{37}}
    \toprule
     & \multirow{2}{*}{\textbf{Setup}} & \multirow{2}{*}{\textbf{Method}} & \multicolumn{4}{c}{\textbf{LLaVA 1.5~\citep{liu2023visual}}} & \multicolumn{4}{c}{\textbf{InstructBLIP~\citep{dai2024instructblip}}} & \multicolumn{4}{c}{\textbf{mPLUG-Owl2~\citep{ye2024mplug}}} \\
    \arrayrulecolor{gray} \cmidrule(lr){4-7} \cmidrule(lr){8-11} \cmidrule(lr){12-15}
     &  &  & {{Acc.} {\up}} & {{Prec.} {\up}} & {{Rec.} {\up}} & {{F1} {\up}} & {{Acc.} {\up}} & {{Prec.} {\up}} & {{Rec.} {\up}} & {{F1} {\up}} & {{Acc.} {\up}} & {{Prec.} {\up}} & {{Rec.} {\up}} & {{F1} {\up}} \\
    \midrule
    \multirow{21}{*}{\rot{\textbf{\normalsize MS-COCO~\citep{lin2014microsoft}\qquad\qquad}}} & \multirow{9}{*}{Random} 
    & \textit{base} & 84.13 & 82.86 & 86.07 & 84.43 & 82.80 & 82.24 & 83.67 & 82.95 & 81.00 & 75.27 & \colorbox{SecondBest}{92.33} & 82.93 \\
     &  & VCD & 85.37 & 83.14 & \colorbox{SecondBest}{88.73} & 85.84 & 83.93 & 84.42 & 82.67 & 83.73 & 81.53 & 76.40 & 91.27 & 83.17 \\
     &  & M3ID & {86.00} & {85.11} & 87.27 & {86.18} & {84.37} & {84.62} & {84.00} & {84.31} & 80.90 & 75.29 & 92.00 & 82.81 \\
     &  & DoLa & 85.97          & 85.10          & 87.20          & 86.14          & 84.00          & 82.86          & 85.73    & 84.27 & 81.20 & 75.97 & 91.27 & 82.92 \\    
     
     \arrayrulecolor{gray!50}\cmidrule(lr){3-15}
     
     &  & \textbf{\Ours} & \colorbox{SecondBest}{88.87}	& \colorbox{Best}{89.23} & {88.40} & 
     \colorbox{SecondBest}{88.81} & \colorbox{Best}{88.83} & \colorbox{Best}{90.48} & \colorbox{Best}{86.80} & \colorbox{Best}{88.60} & \colorbox{SecondBest}{84.83} & \colorbox{SecondBest}{80.40} & 92.13 & \colorbox{SecondBest}{85.87} \\
     &  & \textbf{\Oursplus} & \colorbox{Best}{89.17} & \colorbox{SecondBest}{88.89} & \colorbox{Best}{89.53} & \colorbox{Best}{89.21} & \colorbox{SecondBest}{88.67} & \colorbox{SecondBest}{90.28} & \colorbox{SecondBest}{86.67} & \colorbox{SecondBest}{88.44} & \colorbox{Best}{85.57} & \colorbox{Best}{81.18} & \colorbox{Best}{92.60} & \colorbox{Best}{86.52} \\
     
     % &  & \textbf{\Ours}~\&~VCD & 89.07	& 89.49	& 88.53	& 89.01	& 89.30	& 90.85	& 87.40	& 89.09 &  &  &  &  \\
     % &  & \textbf{\Ours}~\&~M3ID & 89.00 & 89.85	& 87.93	& 88.88	& 88.93	& 91.13	& 86.27	& 88.63 &  &  &  &  \\
     % \arrayrulecolor{gray!50}\cmidrule(lr){3-15}
     \arrayrulecolor{gray!50}\cmidrule(lr){3-15}
     &  & OPERA (Beam) & 89.37          & 92.03          & 86.20          & 89.02          & 89.17          & 95.51          & 82.20          & 88.36 &  89.27 & 89.48 & 89.00 & 89.24  \\
     \arrayrulecolor{gray}\cmidrule(lr){2-15}
      & \multirow{9}{*}{Popular} & \textit{base} & 80.87 &	78.23 &	85.53 &	81.72 & 75.80	& 72.74 &	82.53 &	77.33 & 76.27 & 69.96 & \colorbox{SecondBest}{92.07} & 79.50 \\
     &  & VCD & 81.10 & 77.78 & {87.07} & 82.16 & {77.73} & {75.43} & 82.27 & 78.70 & 75.70 & 69.88 & 90.33 & 78.80 \\
     &  & M3ID & {82.83} & {79.62} & \colorbox{SecondBest}{88.27} & {83.72} & 77.30 & 74.10 & {83.93} & {78.71} & 76.50 & 70.23 & 92.00 & 79.65 \\
      &  & DoLa & 82.93 & 79.76 & \colorbox{SecondBest}{88.27} & 83.80 & 77.37 & 73.50 & 85.60 & 79.09 & 76.67 & 70.58 & 91.47 & 79.67 \\
      
     \arrayrulecolor{gray!50}\cmidrule(lr){3-15} 
     
     &  & \textbf{\Ours} & \colorbox{SecondBest}{85.83} & \colorbox{SecondBest}{84.17} & \colorbox{SecondBest}{88.27} & \colorbox{SecondBest}{86.17} & \colorbox{SecondBest}{81.97} & \colorbox{SecondBest}{78.90} & \colorbox{SecondBest}{87.27} & \colorbox{SecondBest}{82.87} & \colorbox{SecondBest}{80.43} & \colorbox{SecondBest}{74.64} & \colorbox{Best}{92.20} & \colorbox{SecondBest}{82.49} \\
     &  & \textbf{\Oursplus} & \colorbox{Best}{86.65} & \colorbox{Best}{85.35} & \colorbox{Best}{88.67} & \colorbox{Best}{86.98} & \colorbox{Best}{82.63} & \colorbox{Best}{79.65} & \colorbox{Best}{87.67} & \colorbox{Best}{83.47} & \colorbox{Best}{80.83} & \colorbox{Best}{75.62} & 91.00 & \colorbox{Best}{82.60} \\
     % &  & \textbf{\Ours}~\&~VCD & 85.77 & 83.89 & 88.53 & 86.15 & 82.83 & 80.16 & 87.27 & 83.56 &  &  &  &  \\
     % &  & \textbf{\Ours}~\&~M3ID & 85.37 & 83.60 & 88.00 & 85.74 & 81.90 & 78.98 & 86.93 & 82.77 &  &  &  &  \\
     % \arrayrulecolor{gray!50}\cmidrule(lr){3-15}
     \arrayrulecolor{gray!50}\cmidrule(lr){3-15}
     &  & OPERA (Beam) & 86.20 & 85.17 & 87.67 & 86.40 & 84.07 & 85.39 & 82.20 & 83.76 &  84.13 & 81.11 & 89.00 & 84.87  \\
     \arrayrulecolor{gray}\cmidrule(lr){2-15}
      & \multirow{9}{*}{Adversarial} & \textit{base} & 76.23 & 71.75 & 86.53 & 78.45 & 75.40 & 71.60 & 84.20 & 77.39 & 73.20 & 66.88 & 91.93 & 77.43 \\
     &  & VCD & 75.60 & 70.78 & 87.20 & 78.14 & {76.80} & {73.62} & 83.53 & {78.26} & 73.23 & 67.26 & 90.53 & 77.18 \\
     &  & M3ID & {77.70} & {73.23} & {87.33} & {79.66} & 76.03 & 72.48 & {83.93} & 77.79 & 72.57 & 66.28 & 91.87 & 77.00 \\
     &  & DoLa & 77.17 & 72.30 & 88.07 & 79.41 & 74.30 & 69.95 & 85.20 & 76.83 & 72.37 & 66.29 & 91.00 & 76.71 \\
     \arrayrulecolor{gray!50}\cmidrule(lr){3-15} 
     
     &  & \textbf{\Ours} & \colorbox{SecondBest}{78.80} & \colorbox{SecondBest}{74.43} & \colorbox{SecondBest}{87.73} & \colorbox{SecondBest}{80.54} & \colorbox{Best}{78.73} & \colorbox{SecondBest}{74.57} & \colorbox{Best}{87.20} & \colorbox{Best}{80.39} & \colorbox{SecondBest}{75.23} & \colorbox{SecondBest}{68.88} & \colorbox{Best}{92.07} & \colorbox{SecondBest}{78.80} \\
     &  & \textbf{\Oursplus} &  \colorbox{Best}{79.37} &  \colorbox{Best}{74.62} &  \colorbox{Best}{89.00} &  \colorbox{Best}{81.18} &  \colorbox{SecondBest}{78.63} &  \colorbox{Best}{74.70} &  \colorbox{SecondBest}{86.60} &  \colorbox{SecondBest}{80.21} & \colorbox{Best}{75.57} & \colorbox{Best}{69.24} & \colorbox{SecondBest}{92.00} & \colorbox{Best}{79.02} \\
     % &  & \textbf{\Ours}~\&~VCD & 79.60 & 75.26 & 88.20 & 81.22 & 79.07 & 74.89 & 87.47 & 80.69 &  &  &  &  \\
     % &  & \textbf{\Ours}~\&~M3ID & 79.20 & 74.83 & 88.00 & 80.88 & 78.93 & 75.06 & 86.67 & 80.45 &  &  &  &  \\
     % \arrayrulecolor{gray!50}\cmidrule(lr){3-15}
     
     \arrayrulecolor{gray!50}\cmidrule(lr){3-15}
     &  & OPERA (Beam) & 81.07 & 77.44 & 87.67 & 82.24 & 81.83 & 81.60 & 82.20 & 81.90 &  80.00 & 75.42 & 89.00 & 81.65  \\
     \arrayrulecolor{gray}\midrule
     \multirow{21}{*}{\rot{\textbf{\normalsize A-OKVQA~\citep{schwenk2022okvqa}\qquad\qquad}}} & \multirow{9}{*}{Random} & \textit{base} & 81.73 & 76.53 & 91.53 & 83.36 & 81.13 & 78.03 & 86.67 & 82.12 & 78.13 & 70.87 & 95.53 & 81.37 \\
     &  & VCD & 81.83 & 75.74 & 93.67 & 83.76 & 82.00 & {79.38} & 86.47 & 82.77 & 77.70 & 70.42 & 95.53 & 81.07 \\
     &  & M3ID & {83.57} & {77.86} & {93.80} & {85.09} & {82.33} & 77.81 & 90.47 & 83.66 & 78.23 & 70.73 & \colorbox{Best}{96.33} & 81.57 \\
     &  & DoLa & 83.23 & 77.47 & 93.73 & 84.83 & 82.17 & 78.17 & 89.27 & 83.35 & 77.67 & 70.38 & 95.53 & 81.05 \\
     
     \arrayrulecolor{gray!50}\cmidrule(lr){3-15} 
     &  & \textbf{\Ours} & \colorbox{SecondBest}{85.17} & \colorbox{SecondBest}{79.79} & \colorbox{Best}{94.20} & \colorbox{SecondBest}{86.40} & \colorbox{SecondBest}{87.13} & \colorbox{SecondBest}{83.92} & \colorbox{Best}{91.87} & \colorbox{SecondBest}{87.71} & \colorbox{SecondBest}{80.20} & \colorbox{SecondBest}{73.02} & \colorbox{SecondBest}{95.80} & \colorbox{SecondBest}{82.87} \\
     &  & \textbf{\Oursplus} & \colorbox{Best}{85.43} & \colorbox{Best}{80.15} &\colorbox{Best}{94.20} & \colorbox{Best}{86.81} & \colorbox{Best}{87.40} & \colorbox{Best}{84.42} & \colorbox{SecondBest}{91.73} & \colorbox{Best}{87.92} & \colorbox{Best}{80.37} & \colorbox{Best}{73.35} & 95.40 & \colorbox{Best}{82.93} \\
     
     % &  & \textbf{\Ours}~\&~VCD & 85.10 & 79.93 & 93.73 & 86.28 & 86.77 & 83.57 & 91.53 & 87.37 &  &  &  &  \\
     % &  & \textbf{\Ours}~\&~M3ID & 85.93 & 80.62 & 94.60 & 87.06 & 87.17 & 84.35 & 91.27 & 87.67 &  &  &  &  \\
     % \arrayrulecolor{gray!50}\cmidrule(lr){3-15}
     \arrayrulecolor{gray!50}\cmidrule(lr){3-15}
     &  & OPERA (Beam) & 86.80 & 82.90 & 92.73 & 87.54 & 89.97 & 90.75 & 89.00 & 89.87 &  86.57 & 82.17 & 93.40 & 87.43  \\
     \arrayrulecolor{gray}\cmidrule(lr){2-15}
      & \multirow{9}{*}{Popular} & \textit{base} & 76.67 & {70.51} & 91.67 & 79.71 & 75.67 & 70.97 & 86.87 & 78.12 & 71.27 & 64.43 & 94.93 & 76.77 \\
     &  & VCD & 74.70 & 68.12 & 92.87 & 78.59 & {76.50} & {71.69} & 87.60 & {78.85} & 71.07 & 64.21 & 95.20 & 76.69 \\
     &  & M3ID & {76.80} & 70.20 & {93.13} & {80.06} & 75.60 & 70.40 & {88.33} & 78.36 & 69.57 & 62.80 & \colorbox{SecondBest}{96.00} & 75.93 \\
     &  & DoLa & 76.47 & 69.79 & 93.33 & 79.86 & 76.93 & 71.15 & 90.60 & 79.71 & 71.10 & 64.22 & 95.27 & 76.72 \\
     
     \arrayrulecolor{gray!50}\cmidrule(lr){3-15}
     
     &  & \textbf{\Ours} & \colorbox{SecondBest}{78.83} & \colorbox{SecondBest}{71.99} & \colorbox{SecondBest}{94.40} & \colorbox{SecondBest}{81.68} & \colorbox{SecondBest}{78.73} & \colorbox{SecondBest}{72.83} & \colorbox{SecondBest}{91.67} & \colorbox{SecondBest}{81.17} & \colorbox{SecondBest}{74.20} & \colorbox{Best}{66.96} & 95.53 & \colorbox{SecondBest}{78.74} \\
     &  & \textbf{\Oursplus} & \colorbox{Best}{79.13} & \colorbox{Best}{72.30} & \colorbox{Best}{94.47} & \colorbox{Best}{81.91} & \colorbox{Best}{79.00} & \colorbox{Best}{72.92} & \colorbox{Best}{92.27} & \colorbox{Best}{81.46} & \colorbox{Best}{74.37} & \colorbox{SecondBest}{66.93} & \colorbox{Best}{96.33} & \colorbox{Best}{78.98} \\
     % &  & \textbf{\Ours}~\&~VCD & 79.17 & 72.40 & 94.27 & 81.90 & 78.83 & 72.75 & 92.20 & 81.33 &  &  &  &  \\
     % &  & \textbf{\Ours}~\&~M3ID & 79.63 & 72.83 & 94.53 & 82.27 & 79.20 & 73.42 & 91.53 & 81.48 &  &  &  &  \\
     % \arrayrulecolor{gray!50}\cmidrule(lr){3-15}
     \arrayrulecolor{gray!50}\cmidrule(lr){3-15}
     &  & OPERA (Beam) & 79.60 & 73.44 & 92.73 & 81.97 & 82.60 & 78.90 & 89.00 & 83.65 &  80.90 & 74.72 & 93.40 & 83.02  \\

     \arrayrulecolor{gray}\cmidrule(lr){2-15}
      & \multirow{9}{*}{Adversarial} & \textit{base} & 67.40 & 61.78 & 91.27 & 73.68 & 68.00 & 63.08 & 86.80 & 73.06 & 64.83 & 59.15 & \colorbox{SecondBest}{95.87} & 73.16 \\
     &  & VCD & 67.43 & 61.48 & 93.33 & 74.13 & \colorbox{SecondBest}{70.67} & \colorbox{Best}{65.24} & {88.47} & {75.10} & \colorbox{Best}{66.43} & \colorbox{Best}{60.39} & 95.53 & \colorbox{SecondBest}{74.00} \\
     &  & M3ID & {68.10} & {61.99} & {93.60} & {74.58} & 69.57 & {64.21} & 88.40 & 74.39 & 65.13 & 59.33 & \colorbox{Best}{96.27} & 73.41 \\
     &  & DoLa & 68.03 & 62.02 & 93.07 & 74.43 & 68.50 & 62.94 & 90.00 & 74.07 & 65.73 & 59.91 & 95.13 & 73.52 \\
     \arrayrulecolor{gray!50}\cmidrule(lr){3-15}
     
     &  & \textbf{\Ours} & \colorbox{SecondBest}{68.57} & \colorbox{SecondBest}{62.26} & \colorbox{SecondBest}{94.27} & \colorbox{SecondBest}{74.99} & {70.27} & 64.15 & \colorbox{SecondBest}{91.87} & \colorbox{SecondBest}{75.55} & 65.93 & 59.99 & 95.67 & 73.74 \\
     &  & \textbf{\Oursplus} & \colorbox{Best}{68.80} & \colorbox{Best}{62.51} & \colorbox{Best}{94.47} & \colorbox{Best}{75.23} & \colorbox{Best}{70.97} & \colorbox{SecondBest}{64.74} & \colorbox{Best}{92.07} & \colorbox{Best}{76.03} & \colorbox{SecondBest}{66.20} & \colorbox{SecondBest}{60.12} & \colorbox{Best}{96.27} & \colorbox{Best}{74.01} \\
     % &  & \textbf{\Ours}~\&~VCD & 68.80 & 62.48 & 94.13 & 75.11 & 71.00 & 64.72 & 92.33 & 76.10 &  &  &  &  \\
     % &  & \textbf{\Ours}~\&~M3ID & 68.77 & 62.42 & 94.33 & 75.13 & 69.30 & 63.43 & 91.13 & 74.80 &  &  &  &  \\
     % \arrayrulecolor{gray!50}\cmidrule(lr){3-15}
     \arrayrulecolor{gray!50}\cmidrule(lr){3-15}
     &  & OPERA (Beam) & 70.00 & 63.75 & 92.73 & 75.56 & 74.53 & 69.03 & 89.00 & 77.75 &  71.17 & 64.65 & 93.40 & 76.41
  \\
  
     \arrayrulecolor{gray}\midrule
     \multirow{21}{*}{\rot{\textbf{\normalsize GQA~\citep{hudson2019gqa}\qquad\qquad}}} & \multirow{9}{*}{Random} & \textit{base} & 81.23 & 75.42 & 92.67 & 83.16 & 79.93 & 76.73 & 85.93 & 81.07 & 80.00 & 74.04 & 92.40 & 82.21 \\
     &  & VCD & 81.50 & 74.78 & {95.07} & 83.71 & {81.83} & {79.03} & 86.67 & {82.67} & 81.60 & \colorbox{Best}{77.56} & 88.93 & 82.86 \\
     &  & M3ID & {82.83} & {76.64} & 94.47 & {84.62} & 80.57 & 76.77 & {87.67} & 81.85 & 80.93 & 74.95 & 92.93 & 82.98 \\
     &  & DoLa & 83.70 & 77.70 & 94.53 & 85.29 & 81.57 & 77.90 & 88.13 & 82.70 & 78.67 & 73.19 & 90.47 & 80.92 \\

     \arrayrulecolor{gray!50}\cmidrule(lr){3-15}
     &  & \textbf{\Ours} & \colorbox{SecondBest}{{86.10}} & \colorbox{SecondBest}{80.30} & \colorbox{SecondBest}{95.67} & \colorbox{SecondBest}{87.31} & \colorbox{SecondBest}{84.87} & \colorbox{SecondBest}{82.52} & \colorbox{SecondBest}{88.47} & \colorbox{SecondBest}{85.39} & \colorbox{SecondBest}{82.10} & 76.10 & \colorbox{SecondBest}{93.60} & \colorbox{SecondBest}{83.95} \\
     &  & \textbf{\Oursplus} & \colorbox{Best}{86.77} & \colorbox{Best}{81.00} & \colorbox{Best}{96.40} & \colorbox{Best}{88.03} & \colorbox{Best}{85.43} & \colorbox{Best}{83.20} & \colorbox{Best}{88.80} & \colorbox{Best}{85.91} & \colorbox{Best}{82.60} & \colorbox{SecondBest}{76.66} & \colorbox{Best}{93.73} & \colorbox{Best}{84.34} \\
     % &  & \textbf{\Ours}~\&~VCD & {86.03}	&80.21 & 95.67 & 87.26 & 84.97 & 82.40 & 88.93 & 85.54 &  &  &  &  \\
     % &  & \textbf{\Ours}~\&~M3ID & 86.30 & 80.64 & 95.53 & 87.46 & 85.00 & 82.94 & 88.13 & 85.46 &  &  &  &  \\
     % \arrayrulecolor{gray!50}\cmidrule(lr){3-15}
     \arrayrulecolor{gray!50}\cmidrule(lr){3-15}
     &  & OPERA (Beam) & 87.07 & 82.25 & 94.53 & 87.97 & 87.70 & 90.02 & 84.80 & 87.33 &  86.27 & 85.65 & 87.13 & 86.38  \\     
     \arrayrulecolor{gray}\cmidrule(lr){2-15}
      & \multirow{9}{*}{Popular} & \textit{base} & 72.50 & 65.85 & 93.47 & 77.27 & 72.73 & 68.14 & 85.40 & 75.80 & 71.53 & 64.94 & 93.60 & {76.68} \\
     &  & VCD & 71.57 & 64.72 & {94.80} & 76.93 & 73.67 & 68.82 & 86.53 & 76.67 & 71.40 & 65.77 & 89.27 & 75.74 \\
     &  & M3ID & {72.83} & {66.04} & 94.00 & {77.58} & \colorbox{SecondBest}{74.57} & \colorbox{SecondBest}{69.45} & {87.73} & {77.53} & 71.50 & 65.06 & 92.87 & 76.52 \\
     &  & DoLa & 74.03 & 66.85 & 95.33 & 78.59 & 73.70 & 68.58 & 87.47 & 76.88 & 71.03 & 65.23 & 90.07 & 75.67 \\
     \arrayrulecolor{gray!50}\cmidrule(lr){3-15}

     &  & \textbf{\Ours} & \colorbox{SecondBest}{74.80} & \colorbox{SecondBest}{67.50} & \colorbox{SecondBest}{95.67} & \colorbox{SecondBest}{79.15} & {74.50} & {69.17} & \colorbox{SecondBest}{88.40} & \colorbox{SecondBest}{77.61} & \colorbox{SecondBest}{73.47} & \colorbox{SecondBest}{66.60} & \colorbox{SecondBest}{94.13} & \colorbox{SecondBest}{78.01} \\
     &  & \textbf{\Oursplus} & \colorbox{Best}{75.47} & \colorbox{Best}{68.32} & \colorbox{Best}{96.20} & \colorbox{Best}{79.90} & \colorbox{Best}{76.10} & \colorbox{Best}{70.49} & \colorbox{Best}{89.80} & \colorbox{Best}{78.98} & \colorbox{Best}{73.93} & \colorbox{Best}{66.95} & \colorbox{Best}{94.53} & \colorbox{Best}{78.39} \\
     % &  & \textbf{\Ours}~\&~VCD & 75.07 & 67.82 & 95.40 & 79.28 & 75.33 & 69.98 & 88.73 & 78.25 &  &  &  &  \\
     % &  & \textbf{\Ours}~\&~M3ID & 74.40 & 67.15 & 95.53 & 78.87 & 75.57 & 70.24 & 88.73 & 78.41 &  &  &  &  \\
     % \arrayrulecolor{gray!50}\cmidrule(lr){3-15}
     \arrayrulecolor{gray!50}\cmidrule(lr){3-15}
     &  & OPERA (Beam) & 75.50 & 68.47 & 94.53 & 79.42 & 78.77 & 75.67 & 84.80 & 79.97 &  76.60 & 71.97 & 87.13 & 78.83  \\
     \arrayrulecolor{gray}\cmidrule(lr){2-15}
      & \multirow{9}{*}{Adversarial} & \textit{base} & 67.63 & 61.68 & 93.13 & 74.21 & {69.57} & \colorbox{SecondBest}{64.80} & 85.67 & {73.79} & 68.73 & 62.60 & {93.07} & 74.85 \\
     &  & VCD & 67.47 & 61.38 & 94.20 & 74.33 & 69.43 & 64.76 & 85.27 & 73.61 & \colorbox{Best}{71.67} & \colorbox{Best}{65.98} & 89.47 & \colorbox{Best}{75.95} \\
     &  & M3ID & {68.13} & \colorbox{SecondBest}{61.88} & {94.47} & {74.78} & 68.90 & 64.06 & {86.13} & 73.47 & 68.23 & 62.29 & 92.40 & 74.42 \\
     &  & DoLa & 68.73 & 62.34 & 94.67 & 75.17 & 69.70 & 64.28 & 88.67 & 74.53 & 69.50 & \colorbox{SecondBest}{63.51} & 91.67 & 75.03 \\
     
     \arrayrulecolor{gray!50}\cmidrule(lr){3-15}
     &  & \textbf{\Ours} & \colorbox{SecondBest}{68.23} & {61.75} & \colorbox{SecondBest}{95.80} & \colorbox{SecondBest}{75.10} & \colorbox{SecondBest}{70.17} & {64.76} & \colorbox{SecondBest}{88.47} & \colorbox{SecondBest}{74.78} & 68.30 & 62.15 & \colorbox{SecondBest}{93.60} & 74.70 \\
     &  & \textbf{\Oursplus} & \colorbox{Best}{69.17} & \colorbox{Best}{62.42} & \colorbox{Best}{96.33} & \colorbox{Best}{75.75} & \colorbox{Best}{70.60} & \colorbox{Best}{65.04} & \colorbox{Best}{89.07} & \colorbox{Best}{75.18} & \colorbox{SecondBest}{69.63} & 63.19 & \colorbox{Best}{94.07} & \colorbox{SecondBest}{75.60} \\
     % &  & \textbf{\Ours}~\&~VCD & 69.00 & 62.39 & 95.67 & 75.53 & 70.23 & 64.81 & 88.53 & 74.84 &  &  &  &  \\
     % &  & \textbf{\Ours}~\&~M3ID & 68.80 & 62.29 & 95.27 & 75.33 & 71.00 & 65.32 & 89.53 & 75.53 &  &  &  &  \\
     % \arrayrulecolor{gray!50}\cmidrule(lr){3-15}
     \arrayrulecolor{gray!50}\cmidrule(lr){3-15}
     &  & OPERA (Beam) & 70.00 & 63.42 & 94.53 & 75.91 & 74.40 & 70.20 & 84.80 & 76.81 &  73.33 & 68.29 & 87.13 & 76.57  \\
    \bottomrule
    \end{tabular}
    }
    \label{tab:POPE}
\end{table*}

\vspace{\secmargin}\section{Experiments}
% \vspace{\secmargin}
\label{sec:experiments}

% \jh{alpha beta는 차이가 너무 심해보일거 같으면 그냥 parameter setting table로 만들어서 서플에 넣고 거기 보라고 하면 좀 숨길수 있으려나}
% \sm{data link나 license 같은것도 넣어야해서 공간보고 implementation detail 전부 supple로 넘겨도될듯. alpha beta는 확실히 서플로}

% \vspace{\subsecmargin}
\subsection{Evaluation Setup}
Throughout our experiments, we set hyperparameter configuration at $\alpha = 3$.
For random image transformation, we use flip (horizontal \& Vertical), rotate, color jitter, Gaussian blur, and crop.
% and $\beta = 0.1$ \jh{beta 설명하려면 cutoff 수식도 넣거나 아니면 걍 다 appendix로 빼야할듯?}.
In all experimental tables, \textit{base} refers to standard decoding, where the token is directly sampled from the softmax distribution.
% which directly samples the response token from the softmax distribution.
To encourage output diversity and avoid deterministic responses, we sample from a multinomial distribution rather than simply selecting the most probable output using $\text{argmax}$.
\footnote{
% We refer readers to~\cref{sec:appendix_implementation_details} for further implementation \& experimental details and additional results.
Further implementation \& experimental details are in Appendix.}
% ~\cref{sec:appendix_implementation_details}.}

\vspace{-2mm}
\vspace{\paramargin}\paragraph{LVLMs.}
We integrate \Ours with three state-of-the-art LVLMs: \textbf{LLaVA-1.5}~\citep{liu2023visual}, \textbf{InstructBLIP}~\citep{dai2024instructblip}, and \textbf{mPLUG-Owl2}~\cite{ye2024mplug}.
Both LLaVA-1.5 and InstructBLIP use Vicuna 7B~\citep{chiang2023vicuna} for language decoding.
LLaVA-1.5 utilizes two-layer MLP to align image and text modalities and InstructBLIP employs the Q-Former~\citep{li2023blip} with a fixed number of tokens (\eg, 32) to bridge visual and textual features efficiently.
mPLUG-Owl2, built on LLaMA 7B~\cite{touvron2023llama2}, combines a vision encoder with learnable queries and a modality-adaptive module to facilitate a shared semantic space between visual and textual modalities.
Note that \Ours is model-agnostic, and its adaptability extends beyond these LVLMs.
% to a wide range of off-the-shelf LVLMs.

% \jh{
% We employed our \Ours on two state-of-the-art LVLMs, LLaVA-1.5~\citep{liu2023visual} and InstructBLIP~\citep{dai2024instructblip}.
% Both models incorporate Vicuna 7B~\citep{} as their language decoding mechanism.
% LLaVA makes use of linear projection layers to align image and text modalities, while InstructBLIP adopts the Q-Former~\citep{} to effectively connect visual and textual features using just 32 tokens.
% Note that \Ours is model-agnostic, making it compatible with a wide range of off-the-shelf LVLMs.
% }

\vspace{-2mm}
\vspace{\paramargin}\paragraph{Baselines.}
Our method aims to reduce hallucinations in LVLMs by modifying model's decoding process without relying on external models, costly self-feedback mechanisms, or additional training. 
% \jh{OPERA 넣을거면 self-feedback 빼기?}
% Therefore, we select baseline methods that align with these criteria.
To align with these criteria, we select baseline methods that meet these requirements.
% Recent contrastive decoding methods meet these requirements, and we establish two primary approaches as our baselines:
Recent contrastive decoding methods fit well within this scope, and we establish two primary baselines: \textbf{VCD}~\citep{leng2023mitigating} and \textbf{M3ID}~\citep{favero2024multi}.
Both VCD and M3ID aim to mitigate object hallucinations by increasing the influence of the reference image over the language prior.
This is achieved by contrasting output distributions derived from both original and distorted visual inputs.
% \jh{
% We also include DoLa~\citep{chuang2023dola} as a baseline, which employs a novel decoding strategy by contrasting layers within the model, particularly focusing on the logits derived from both earlier and later layers of the transformer architecture.
We also include \textbf{DoLa}~\citep{chuang2023dola} as a baseline, which employs a novel decoding strategy that contrasts logits from earlier and later layers of the transformer architecture.
This amplifies factual knowledge stored in the upper layers while suppressing linguistic patterns from the lower layers that may lead to hallucinations.
% 
% Additionally, we report results from OPERA~\citep{huang2023opera}, which mitigates hallucinations in LVLMs via an over-trust penalty and retrospection allocation.
Additionally, we report results from \textbf{OPERA}~\citep{huang2023opera}, which mitigates hallucinations in LVLMs via an over-trust penalty and retrospection allocation.
% Although OPERA relies on a costly self-feedback mechanism, making it incompatible with our primary criteria, we include it for comparison purposes due to its effectiveness in reducing hallucinations.
% Note that OPERA used beam search instead of sampling during response generation, which contributes to its higher performance.
% While all other methods employ greedy decoding, OPERA uses beam search during response generation, contributing to its higher performance.
In contrast to all other methods, OPERA uses beam search during response generation, contributing to its higher performance.
We include it for comparison purposes due to its demonstrated effectiveness in reducing hallucinations.
% }
% We reproduced baselines within our evaluation setting.
All methods were reimplemented in our evaluation setup to ensure a fair comparison.

\begin{table*}[t!]
    \begin{minipage}[t!]{0.67\textwidth}
        \begin{center}
        \begin{small}
        % \vspace{\abovetabcapmargin}
        \captionof{table}{
            \textbf{Results on MME-Hallucination~\citep{fu2024mme}.}
        \Ours effectively mitigates hallucinations at both the object and attribute levels, outperforming contrastive decoding methods in Total Score.
        \Oursplus further enhances performance by adaptively selecting appropriate augmentations, leading to improved mitigation of hallucinations.
        % \todo{\Oursplus 설명 추가, 적절한 aug 골라서 성능 향상됐다든가}
        }
        \vspace{\belowtabcapmargin}
        \setlength{\tabcolsep}{6pt} % base value: 6pt
        \renewcommand{\arraystretch}{1.04}
        \scalebox{0.8}{
        \begin{tabular}{llx{55}x{55}x{55}x{55}x{55}}
        \toprule
        \multirow{2}{*}{\textbf{Model}} & \multirow{2}{*}{\textbf{Method}} & \multicolumn{2}{c}{\textbf{Object-level}} & \multicolumn{2}{c}{\textbf{Attribute-level}} & \multicolumn{1}{c}{\multirow{2}{*}{\textbf{\makecell{Total \\ Score}}}}\\
        \arrayrulecolor{gray} \cmidrule{3-4} \cmidrule{5-6} 
         & & Existence {\up} & Count {\up} & Position {\up} & Color {\up} & \\
        \arrayrulecolor{gray} \midrule
        
        % \multirow{6}{*}{\textbf{LLaVA 1.5}}
        \multirow{6}{*}{\rot{\textbf{LLaVA1.5}}}
        
        & \textit{base} & $173.75_{(\pm 4.79)}$ & $121.67_{(\pm 12.47)}$ & $117.92_{(\pm 3.69)}$ & $149.17_{(\pm 7.51)}$ & $562.50_{(\pm 3.96)}$ \\
         & VCD & $178.75_{(\pm 2.50)}$ & $126.25_{(\pm 10.40)}$ & $120.00_{(\pm 4.08)}$ & $150.83_{(\pm 11.01)}$ & $575.84_{(\pm 9.67)}$ \\
         & M3ID & $177.50_{(\pm 6.45)}$ & $124.17_{(\pm 10.93)}$ & $120.00_{(\pm 7.07)}$ & $152.92_{(\pm 5.67)}$ & $574.59_{(\pm 9.75)}$ \\ 
         & DoLa & $174.58_{(\pm 5.34)}$ & $122.09_{(\pm 11.73)}$ & \colorbox{SecondBest}{\raisebox{0pt}[6pt][0pt]{\makebox[52pt][c]{$122.09_{(\pm 2.10)}$}}} & $149.17_{(\pm 4.19)}$ & $567.92_{(\pm 13.63)}$ \\ 
         \arrayrulecolor{gray!50}\cmidrule(lr){2-7}
         & \textbf{\Ours} & \colorbox{SecondBest}{\raisebox{0pt}[6pt][0pt]{\makebox[52pt][c]{$ 187.50_{(\pm 2.89)}$}}} & \colorbox{SecondBest}{\raisebox{0pt}[6pt][0pt]{\makebox[52pt][c]{$ 139.58_{(\pm 7.62)}$}}} & \colorbox{Best}{\raisebox{0pt}[6pt][0pt]{\makebox[52pt][c]{$ 125.00_{(\pm 10.27)}$}}} & \colorbox{SecondBest}{\raisebox{0pt}[6pt][0pt]{\makebox[52pt][c]{$ 164.17_{(\pm 6.87 )}$}}} & \colorbox{SecondBest}{\raisebox{0pt}[6pt][0pt]{\makebox[52pt][c]{$ 616.25_{(\pm 20.38 )}$}}} \\
         & \textbf{\Oursplus} & \colorbox{Best}{\raisebox{0pt}[6pt][0pt]{\makebox[52pt][c]{$ 188.89_{(\pm 6.74)}$}}} & \colorbox{Best}{\raisebox{0pt}[6pt][0pt]{\makebox[52pt][c]{$ 145.55_{(\pm 2.55)}$}}} & $ 110.00_{(\pm 21.86)}$ & \colorbox{Best}{\raisebox{0pt}[6pt][0pt]{\makebox[52pt][c]{$ 173.89_{(\pm 10.58)}$}}} & \colorbox{Best}{\raisebox{0pt}[6pt][0pt]{\makebox[52pt][c]{$ 618.33_{(\pm 28.04)}$}}} \\
         % & \textbf{\Ours}+VCD & $ 185.00_{(\pm 4.08 )}$ & $ 140.84_{(\pm 4.41 )}$ & $ 125.00_{(\pm 7.07 )}$ & $ 165.83_{(\pm 6.46 )}$ & $ 616.67_{(\pm 11.14 )}$ \\
         % & \textbf{\Ours}+M3ID & $ 187.50_{(\pm 2.89 )}$ & $ 141.25_{(\pm 9.85 )}$ & $ 125.00_{(\pm 10.27 )}$ & $ 164.17_{(\pm 6.87 )}$ & $ 617.92_{(\pm 22.12  )}$ \\
         % \arrayrulecolor{gray!50}\cmidrule(lr){2-7}
         \arrayrulecolor{gray} \midrule
         
        % \multirow{6}{*}{\textbf{InstructBLIP}} 
        \multirow{6}{*}{\rot{\textbf{InstructBLIP}}}
        & \textit{base} & $ 160.42_{(\pm 5.16 )}$ & $ 79.17_{(\pm 8.22 )}$ & \colorbox{Best}{\raisebox{0pt}[6pt][0pt]{\makebox[52pt][c]{$ 79.58_{(\pm 8.54 )}$}}} & $ 130.42_{(\pm 17.34 )}$ & $ 449.58_{(\pm 24.09 )}$ \\
         & VCD & $ 158.75_{(\pm 7.25 )}$ & \colorbox{SecondBest}{\raisebox{0pt}[6pt][0pt]{\makebox[52pt][c]{$ 90.75_{(\pm 3.11 )}$}}} & $ 70.00_{(\pm 15.81 )}$ & $ 132.50_{(\pm 18.78 )}$ & $452.00 _{(\pm 31.57 )}$ \\
         & M3ID & $ 158.33_{(\pm 5.44 )}$ & \colorbox{Best}{\raisebox{0pt}[6pt][0pt]{\makebox[52pt][c]{$ 94.58_{(\pm 9.85 )}$}}} & $ 72.50_{(\pm 17.03 )}$ & $ 128.33_{(\pm 14.72 )}$ & $ 453.75_{(\pm 26.82 )}$ \\
         & DoLa & $162.08_{(\pm 5.34)}$ & $82.50_{(\pm 6.16)}$ & \colorbox{SecondBest}{\raisebox{0pt}[6pt][0pt]{\makebox[52pt][c]{$78.75_{(\pm 8.96)}$}}} & $135.42_{(\pm 10.49)}$ & \colorbox{SecondBest}{\raisebox{0pt}[6pt][0pt]{\makebox[52pt][c]{$458.75_{(\pm 11.25)}$}}} \\ 
         \arrayrulecolor{gray!50}\cmidrule(lr){2-7}
         & \textbf{\Ours} & \colorbox{SecondBest}{\raisebox{0pt}[6pt][0pt]{\makebox[52pt][c]{$ 182.50_{(\pm 6.45 )}$}}} & $74.58 _{(\pm 5.99 )}$ & $ 67.08_{(\pm 10.31 )}$ & \colorbox{SecondBest}{\raisebox{0pt}[6pt][0pt]{\makebox[52pt][c]{$ 139.17_{(\pm 0.96 )}$}}} & \colorbox{SecondBest}{\raisebox{0pt}[6pt][0pt]{\makebox[52pt][c]{$ 463.33_{(\pm 12.40 )}$}}} \\
         & \textbf{\Oursplus} & \colorbox{Best}{\raisebox{0pt}[6pt][0pt]{\makebox[52pt][c]{$ 187.22_{(\pm 5.09 )}$}}} & $88.89_{(\pm 13.47)}$ & $72.22_{(\pm 7.52)}$ & \colorbox{Best}{\raisebox{0pt}[6pt][0pt]{\makebox[52pt][c]{$ 148.33_{(\pm 10.93 )}$}}} & \colorbox{Best}{\raisebox{0pt}[6pt][0pt]{\makebox[52pt][c]{$ 496.67_{(\pm 4.41 )}$}}} \\
         % & \textbf{\Ours}+VCD & $ 185.00_{(\pm 4.08 )}$ & $ 75.00_{(\pm 7.07 )}$ & $ 62.50_{(\pm 6.46 )}$ & $141.67 _{(\pm 6.53  )}$ & $ 464.17_{(\pm 9.07 )}$ \\
         % & \textbf{\Ours}+M3ID & $182.50_{(\pm 6.45 )}$ & $ 74.58_{(\pm 2.84 )}$ & $ 63.33_{(\pm 11.55 )}$ & $140.42 _{(\pm2.10  )}$ & $460.83± _{(\pm 11.1 )}$ \\
         % \arrayrulecolor{gray!50}\cmidrule(lr){2-7}
         \arrayrulecolor{gray} \midrule
         
        % \multirow{6}{*}{\textbf{mPLUG-Owl2}}
        \multirow{6}{*}{\rot{\textbf{mPLUG-Owl2}}}
        & \textit{base} & $174.58_{(\pm 4.17)}$ & $155.42_{(\pm 10.03)}$ & $81.67_{(\pm 14.72)}$ & $141.25_{(\pm 13.29)}$ & $552.92_{(\pm 9.94)}$ \\
         & VCD & $170.00_{(\pm 0.00)}$ & $138.75_{(\pm 6.44)}$ & $81.25_{(\pm 12.65)}$ & $138.75_{(\pm 5.51)}$ & $528.75_{(\pm 12.50)}$ \\
         & M3ID & $176.25_{(\pm 4.79)}$ & $157.92_{(\pm 9.75)}$ & $81.67_{(\pm 14.72)}$ & $142.50_{(\pm 12.51)}$ & $558.33_{(\pm 10.28)}$
 \\ 
         & DoLa & $175.00_{(\pm 5.77)}$ & $151.67_{(\pm 5.61)}$ & \colorbox{SecondBest}{\raisebox{0pt}[6pt][0pt]{\makebox[52pt][c]{$82.09_{(\pm 14.17)}$}}} & $139.58_{(\pm 5.51)}$ & $548.33_{(\pm 8.92)}$
 \\
         \arrayrulecolor{gray!50}\cmidrule(lr){2-7}
         & \textbf{\Ours} & \colorbox{SecondBest}{\raisebox{0pt}[6pt][0pt]{\makebox[52pt][c]{$185.00_{(\pm 4.08)}$}}} & \colorbox{Best}{\raisebox{0pt}[6pt][0pt]{\makebox[52pt][c]{$159.58_{(\pm 13.57)}$}}} & $77.50_{(\pm 9.57)}$ & \colorbox{SecondBest}{\raisebox{0pt}[6pt][0pt]{\makebox[52pt][c]{$160.42_{(\pm 4.59)}$}}} & \colorbox{SecondBest}{\raisebox{0pt}[6pt][0pt]{\makebox[52pt][c]{$582.5_{(\pm 21.71)}$}}} \\
         & \textbf{\Oursplus} & \colorbox{Best}{\raisebox{0pt}[6pt][0pt]{\makebox[52pt][c]{$189.44_{(\pm 5.09)}$}}} & \colorbox{SecondBest}{\raisebox{0pt}[6pt][0pt]{\makebox[52pt][c]{$159.45_{(\pm 5.36)}$}}} & \colorbox{Best}{\raisebox{0pt}[6pt][0pt]{\makebox[52pt][c]{$83.33_{(\pm 20.48)}$}}} & \colorbox{Best}{\raisebox{0pt}[6pt][0pt]{\makebox[52pt][c]{$162.22_{(\pm 8.55)}$}}} & \colorbox{Best}{\raisebox{0pt}[6pt][0pt]{\makebox[52pt][c]{$594.45_{(\pm 39.48)}$}}} \\
         % & \textbf{\Ours}+VCD & & & & & \\
         % & \textbf{\Ours}+M3ID & & & & & \\
         % \arrayrulecolor{gray!50}\cmidrule(lr){2-7}
        \bottomrule
    \end{tabular}
        }
        \label{tab:MME_hallucination}
        \vspace{-3mm}
        \end{small}
        \end{center}    
    \end{minipage}
    \hfill
    \begin{minipage}[t!]{0.3\textwidth}
        \begin{center}
        \begin{small}
        % \vspace{\abovetabcapmargin}
        \captionof{table}{
            \textbf{Results on CHAIR~\citep{rohrbach2018object}.}
\Ours and \Oursplus significantly reduce object hallucinations in caption generation compared to VCD, M3ID, and DoLa.
The number of \textit{max new tokens} is set to 64.
        }
        \vspace{\belowtabcapmargin}
        \scalebox{0.7}{
%             \begin{tabular}{lcccc}
%             \toprule
%             Sigma & Acc. & Prec. & Rec. & F1 \\
%             \arrayrulecolor{gray} \midrule
%                                                           % \arrayrulecolor{gray} \midrule
% 0.5 & 83.77 & 83.61 & 84.00 & 83.80   \\
% 100 & 85.13 & 86.45 & 83.33 & 84.86   \\
%                       \bottomrule
% \end{tabular}
        \begin{tabular}{llx{37}x{37}}
\toprule
&\textbf{Method}&CHAIR$_S${\down}&CHAIR$_I${\down}\\
\arrayrulecolor{gray}\midrule
\multirow{9}{*}{\rot{\textbf{LLaVA1.5}}}&\textit{base}&26.2&9.3\\
&VCD&{22.4}&7.6\\
&M3ID&23.0&\colorbox{Best}{6.8}\\
&DoLa&23.2&7.8\\
% \arrayrulecolor{gray!50}\cmidrule(lr){2-4}
% &\textbf{\Ours}+VCD&20.0&6.8\\
% &\textbf{\Ours}+M3ID&18.0&5.7\\
\arrayrulecolor{gray!50}\cmidrule(lr){2-4}
&\textbf{\Ours}&\colorbox{SecondBest}{20.6}&\colorbox{SecondBest}{6.9}\\
&\textbf{\Oursplus}& \colorbox{Best}{19.6} & \colorbox{Best}{6.8} \\
\arrayrulecolor{gray!50}\cmidrule(lr){2-4}
&OPERA(beam)&23.0&7.5\\

\arrayrulecolor{gray}\midrule

\multirow{9}{*}{\rot{\textbf{InstructBLIP}}}&\textit{base}&28.6&10.3\\
&VCD&{27.2}&{9.1}\\
&M3ID&31.8&10.4\\
&DoLa&36.6&12.5\\
\arrayrulecolor{gray!50}\cmidrule(lr){2-4}
% \arrayrulecolor{gray!50}\cmidrule(lr){2-4}
% &\textbf{\Ours}+VCD&25.0&8.6\\
% &\textbf{\Ours}+M3ID&23.4&7.9\\
&\textbf{\Ours}&\colorbox{SecondBest}{26.0}&\colorbox{SecondBest}{8.8}\\
&\textbf{\Oursplus}& \colorbox{Best}{24.2} & \colorbox{Best}{8.0} \\
\arrayrulecolor{gray!50}\cmidrule(lr){2-4}
&OPERA(beam)&25.6&8.3\\

\arrayrulecolor{gray}\midrule
\multirow{9}{*}{\rot{\textbf{mPLUG-Owl2}}}&\textit{base}&25.8 & 8.4\\
&VCD & 24.0 & 7.8 \\
&M3ID& 22.8 & 7.3 \\
&DoLa& 26.2 & 8.5\\
\arrayrulecolor{gray!50}\cmidrule(lr){2-4}
% \arrayrulecolor{gray!50}\cmidrule(lr){2-4}
% &\textbf{\Ours}+VCD&25.0&8.6\\
% &\textbf{\Ours}+M3ID&23.4&7.9\\
&\textbf{\Ours}&\colorbox{SecondBest}{19.2}&\colorbox{SecondBest}{6.4}\\
&\textbf{\Oursplus}& \colorbox{Best}{18.0} & \colorbox{Best}{5.5} \\
\arrayrulecolor{gray!50}\cmidrule(lr){2-4}
&OPERA(beam)&  18.2 & 5.5 \\

\bottomrule
\end{tabular}
        }

\label{tab:CHAIR}
        \end{small}        
        \end{center}     
    \end{minipage}
    
    \vspace{-3mm}
\end{table*}

\vspace{-2mm}

\vspace{\paramargin}\paragraph{Benchmarks.}
\textbf{(1) POPE}~\citep{li2023evaluating} frames hallucination assessment as a binary classification task using yes/no questions about object presence (\eg, "Is there a dog in the image?").
It evaluates 500 MS-COCO images with questions based on actual objects or nonexistent objects.
The benchmark contains three subsets (random, popular, and adversarial), addressing object prevalence and co-occurrences.
\textbf{(2) MME}~\citep{fu2024mme} is a comprehensive LVLM benchmark assessing 14 subtasks, including object hallucination through tasks like object existence, count, position, and color.
These tasks are framed as binary yes/no questions.
\textbf{(3) CHAIR}~\citep{rohrbach2018object} evaluates the proportion of words in captions that correspond to actual objects in an image, using ground-truth captions and object annotations.
% The metric focuses on 80 objects.
It has two variants:
(i) per-sentence (CHAIR$_S$) is defined as $|\text{\{sentences with hallucinated objects\}}| / |\text{\{all sentences\}}|$.
(ii) per-instance (CHAIR$_I$) is defined as $|\text{\{hallucinated objects\}}| / |\text{\{all objects mentioned\}}|$.
We randomly select 500 images from the COCO~\citep{lin2014microsoft} validation set and conduct image captioning with the prompt "Please describe this image in detail".
% \jh{
% \textbf{(4) LLaVA-Bench~\citep{liu2023visual}} comprises 24 images with 60 questions covering indoor and outdoor scenes, memes, paintings, sketches, etc. The dataset aims to assess the model’s performance in challenging tasks and its adaptability to new domains.
% }

% \textbf{AMBER}~\citep{wang2023llm} provides a multi-dimensional assessment for both generative and discriminative tasks, addressing hallucination in object existence, attributes, and relationships.
% For generative tasks, it employs CHAIR, while for discriminative tasks, it calculates an F1 score.
% The AMBER score combines these metrics as follows: AMBER Score = ((1 − CHAIR) + F1) / 2.
% \jh{AMBER 없어서 생략하기}

% \todo{\textbf{LLaVA-Bench}~\citep{liu2023visual} features a collection of $24$ images, accompanying $60$ questions that span a range of contexts including indoor and outdoor scenes, memes, paintings, and sketches.
% This dataset is crafted to assess the capability of LVLMs in tackling more challenging tasks and their adaptability to new domains.}

\vspace{\subsecmargin}\subsection{Results}
\noindent\textbf{Results on POPE.}
\Cref{tab:POPE} compares various decoding-based hallucination mitigation methods on the POPE benchmark~\citep{li2023evaluating}, reporting results from the same sampling-based decoding approach (sampling from a multinomial distribution).
% , evaluated with three representative LVLMs: LLaVA 1.5~\citep{liu2023visual}, InstructBLIP~\citep{dai2024instructblip}, and mPLUG-Owl2~\citep{ye2024mplug}.
The results demonstrate that \Ours consistently outperforms standard decoding (\textit{base}) and contrastive decoding baselines~\citep{leng2023mitigating,favero2024multi,chuang2023dola},
% (VCD~\citep{leng2023mitigating}, M3ID~\citep{favero2024multi}, and DoLa~\citep{chuang2023dola})
across all datasets (MS-COCO~\citep{lin2014microsoft}, A-OKVQA~\citep{schwenk2022okvqa}, and GQA~\citep{hudson2019gqa}), setups (random, popular, and adversarial), and evaluation metrics, demonstrating its robustness in mitigating object hallucinations.
Moreover, \Oursplus achieves performance comparable to the beam search-based method OPERA~\cite{huang2023opera}, despite its simpler design.
This highlights the effectiveness of incorporating visual context from multiple perspectives.

\noindent\textbf{Results on MME-Hallucination.}
In~\Cref{tab:MME_hallucination}, we compare the results on the MME-hallucination benchmark~\citep{fu2024mme} to assess the model's effectiveness in reducing various types of hallucinations.
When combined with LLaVA-1.5~\citep{liu2023visual}, our approach consistently outperforms all counterparts across both object-level (Existence and Count) and attribute-level (Position and Color) evaluations.
With InstructBLIP~\citep{dai2024instructblip}, while other methods show slight advantages in specific metrics like Count and Position, our method still surpasses the baseline and other contrastive decoding methods in overall Total Score.
Some reductions in performance on Count and Position tasks can be attributed to certain transformations: for example, cropping can reduce visible object quantity, impacting the Count score, and flipping can alter spatial relationships, affecting Position score.
\Oursplus addresses these limitations by adaptively selecting the most suitable transformations based on self-feedback, thereby overcoming challenges associated with individual transformations and improving performance.
With mPLUG-Owl2, our method demonstrates strong performance as well, particularly excelling in Existence and Color tasks.
\begin{figure*}[t]
    \centering
    \includegraphics[width=.9\linewidth]{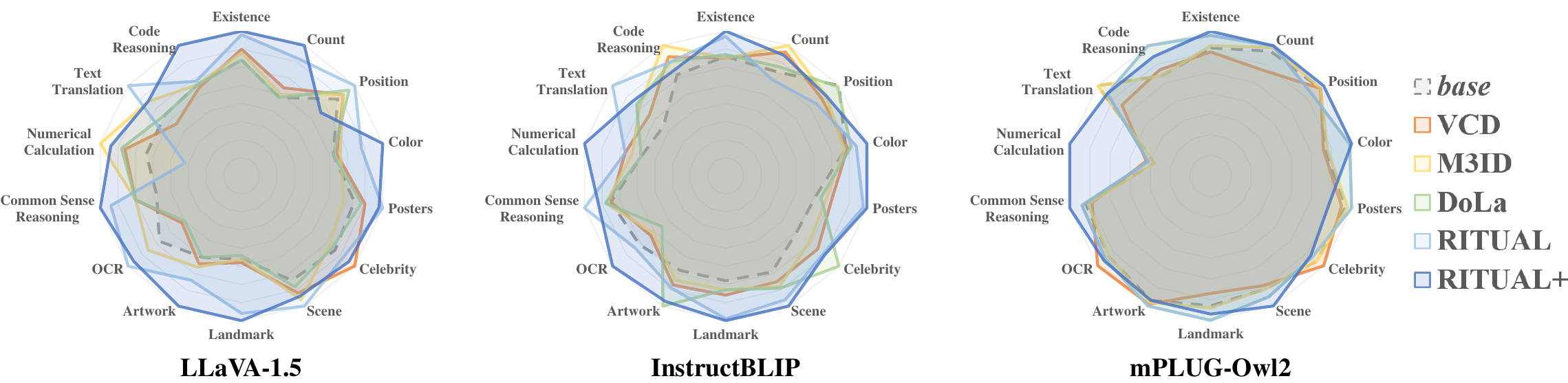}
    \vspace{-2mm}
    % \vspace{\abovefigcapmargin}
    \caption{
        \textbf{Comparison on MME-Fullset~\citep{fu2024mme}.}
        % When equipped with \Ours, LLaVA-1.5~\citep{liu2023visual} performs best in 12 out of 14 categories, while InstructBLIP~\citep{dai2024instructblip} excels in 8 categories and mPLUG-Owl2 excels in 9 categories.
        \Ours significantly enhances the general vision-language capabilities of LVLMs across wide range of tasks.
        When equipped with \Ours, LLaVA-1.5~\citep{liu2023visual} achieves top performance in 12 of the 14 categories, while InstructBLIP~\citep{dai2024instructblip} leads in 8 categories and mPLUG-Owl2~\cite{ye2024mplug} ranks highest in 9 categories.
        % These results demonstrate the effectiveness of \Ours and \Oursplus in providing balanced performance gains across diverse vision-language tasks.
        Detailed results are in Appendix.
        % ~\cref{sec:appendix_mme_fullset}.
        % \todo{general benchmark라고 언급.}
    }%
    \label{fig:MME_full}
    \vspace{\belowfigcapmargin}
\end{figure*}

\noindent\textbf{Results on CHAIR.}
% \cref{tab:CHAIR}
To evaluate hallucinations in generative tasks, we use the CHAIR benchmark, which compares objects in the image with objects in the generated text to measure hallucination levels.
For a fair comparison, we set the maximum number of new tokens to 64 across all methods.
% To assess the reduction of object existence hallucination, we use the CHAIR metrics, where the presence of objects in the description serves as the measurement criterion.
% Given the generative nature of the task, we limit the maximum number of new tokens to 64.
As shown in~\Cref{tab:CHAIR}, \Ours cosistently outperforms both the baseline and prior contrastive decoding approaches.
Specifically, with LLaVA 1.5, \Ours achieves CHAIR$_S$ and CHAIR$_I$ scores of 20.6 and 6.9, respectively, showing a substantial improvement over the baseline scores of 26.2 and 9.3.
Although M3ID slightly outperforms \Ours on CHAIR$_I$, \Ours delivers comparable results while significantly excelling in CHAIR$_S$.
For InstructBLIP, \Ours achieves the best results, with CHAIR$_S$ and CHAIR$_I$ scores of 26.0 and 8.8, marking a major advancement over the baseline scores of 28.6 and 10.3.
Similarly, with mPLUG-Owl2, \Ours records CHAIR$_S$ and CHAIR$_I$ scores of 19.2 and 6.4, outperforming the baseline scores of 25.8 and 8.4 by a large margin.
\Oursplus further enhances these results, demonstrating the value of adaptive transformation selection.
This adaptive approach not only benefits discriminative tasks but also proves effective for descriptive tasks that require a comprehensive understanding of the image content.
% \todo{mPLUG 결과 추가, \Oursplus 한문장 더 좋더라~}

\noindent\textbf{Results on MME-Fullset.}
The MME-Fullset~\citep{fu2024mme} serves as a comprehensive benchmark for assessing the general vision-language capabilities of LVLMs beyond hallucination reduction, covering 14 diverse categories and use cases.
As depicted in~\cref{fig:MME_full}, we evaluate the impact of different decoding methods on LVLM performance across these categories.
% MME-Fullset~\citep{fu2024mme} to assess the impact of decoding methods on the general ability of LVLMs. 
% \todo{hallucination을 넘어 general bench라고 약간 강조?}
% VCD, M3ID, and DoLa show improvements in certain tasks, with VCD excelling in object-level metrics and M3ID in reasoning and attribute-level metrics.
% Across 14 categories, across all LVLMs, adopting \Ours consistently achieve the highest scores across most tasks, demonstrating its effectiveness of \Ours in improving vision-language understanding.
Across all tested LVLMs, \Ours and \Oursplus consistently achieves the highest scores across most tasks, demonstrating its effectiveness in enhancing vision-language comprehension beyond hallucination mitigation.
By enriching the model’s understanding with diverse visual contexts, \Ours provides balanced performance gains across a wide range of tasks, establishing itself as a robust and flexible method for improving LVLM performance.
\Oursplus further enhances these results, showing that adaptive transformation selection improves performance even on more general tasks, confirming the benefit of tailored augmentation for varied use cases.
However, despite the additional visual information provided, some tasks still exhibit slightly lower performance due to inherent challenges within LVLMs, such as statistical biases and language priors.
% By enriching the model's visual capacity from diverse visual contexts, \Ours provides a balanced enhancement across a wide range of tasks, making it a versatile and robust method for improving LVLM performance.
% Despite 다양한 image에 대한 정보를 줘도 , some tasks may still exhibit lower performance due to the inherent challenges of LVLM의 근본적 한계 statistical bias and language priors같은.
% statistical bias and language priors affecting LVLMs. 

% By enriching the model's visual capacity from diverse visual contexts, \Ours boosts the overall performance of LVLMs, including hallucination mitigation.
% Despite these advancements, some tasks may still exhibit lower performance due to the inherent challenges of statistical bias and language priors affecting LVLMs. 

% \todo{remove, appendix에서 설명 -> Additionally, when combined with VCD and M3ID, \Ours further reduces the CHAIR score.}
% \jh{
% We randomly select 500 images from the COCO validation set and conduct image captioning with the prompt "Please describe this image in detail.". We limit the maximum number of new tokens to 64.
% % 
% We verify how much decoding methods reduce object existence hallucination by using CHAIR metrics.
% As shown in \cref{tab:CHAIR}, our \Ours surpasses the baseline and previous contrastive decoding approaches.
% }

\vspace{\subsecmargin}\subsection{Analysis}

\begin{table}[t!]
    \begin{minipage}[t]{0.44\linewidth}
        \begin{center}
        \begin{small}
        % \vspace{\abovetabcapmargin}
        \captionof{table}{
            % \textbf{Generated Text Quality.}
            % \Ours demonstrates a competitive level of text quality compared to other decoding methods.
            % \todo{scale 0 to 10}
            \textbf{GPT4-aided text quality evaluation.} Scores ranging from 1 to 10. 
        }
        \vspace{\belowtabcapmargin}
        \setlength{\tabcolsep}{5pt} % base value: 6pt
        \scalebox{0.78}{
            \begin{tabular}{y{30}x{42}x{36}}
            \toprule
            \multirow{2}{*}{Method} & \multicolumn{2}{c}{LLaVA1.5} \\
            \arrayrulecolor{gray}\cmidrule(lr){2-3}
            & Grammar{\up} & Fluency{\up} \\
            \arrayrulecolor{gray}\midrule
            \textit{base} & 9.804 & 9.432 \\
            VCD & 9.802 & 9.352 \\
            M3ID & 9.832 & 9.344 \\
            DoLa & 9.814 & 9.320 \\
            \Ours & 9.844 & 9.398 \\
            \Oursplus & 9.850 & 9.421 \\
            \arrayrulecolor{gray}\midrule
            \begin{tabular}[c]{@{}l@{}}OPERA\\ (beam)\end{tabular} & 9.828 & 9.308 \\
            \bottomrule
            \end{tabular}
        }
        \label{tab:generation quality}
        \end{small}
        \end{center}    
    \end{minipage}%
    \hfill
    \begin{minipage}[t]{0.52\linewidth}
        \begin{center}
        \begin{small}
        % \vspace{\abovetabcapmargin}
        \captionof{table}{
            \textbf{Comparison of performance and latency on POPE COCO random setup.}
        }
        \setlength{\tabcolsep}{5pt} % base value: 6pt
        \vspace{\belowtabcapmargin}
        \scalebox{0.72}{
            \begin{tabular}{y{30}x{25}x{25}x{35}}
            \toprule
            \multirow{3}{*}{Method} & \multicolumn{3}{c}{LLaVA1.5} \\
            \arrayrulecolor{gray}\cmidrule(lr){2-4}
            & Acc.{\up} & F1{\up} & \begin{tabular}[c]{@{}c@{}}Latency\\ (ms/token)\end{tabular} \\
            \arrayrulecolor{gray}\midrule
            \textit{base} & 84.13 & 84.43 & 21.96 \\
            VCD & 85.37 & 85.84 & 43.33 \\
            M3ID & 86.00 & 86.18 & 40.07 \\
            DoLa & 85.97 & 86.14 & 28.70 \\
            \Ours & 88.87 & 88.81 & 43.37 \\
            \Oursplus & 89.17 & 89.21 & 69.27  \\
            \arrayrulecolor{gray}\midrule
            \begin{tabular}[c]{@{}l@{}}OPERA\\ (beam)\end{tabular} & 89.37 & 89.02 & 308.48 \\
            \bottomrule
            \end{tabular}
        }
        \label{tab:latency}
        \end{small}        
        \end{center}     
    \end{minipage}
    \vspace{-5mm}
\end{table}

\noindent\textbf{Textual Quality.}
Since previous methods and \Ours modify the logits from the standard decoding strategy, there may be concerns about potentially compromising the quality of the generated text.
Therefore, we employed GPT-4-Turbo to assess the grammar and fluency of generated text from 500 samples of the CHAIR benchmark~\cite{rohrbach2018object} using the InstructBLIP~\cite{dai2024instructblip}.
As shown in~\cref{tab:generation quality}, our decoding method demonstrates text generation quality that is comparable to or exceeds that of the previous work in terms of grammar and fluency.
The results highlight the robustness and effectiveness of our method in generating grammatically correct and fluent text while also improving hallucination mitigation without compromising overall text generation quality.

\noindent\textbf{Latency.}
Contrastive decoding methods like VCD~\cite{leng2023mitigating} and M3ID~\cite{favero2024multi}, as well as \Ours, require performing the forward process twice to compare two probability distributions, doubling resource consumption.
\Cref{tab:latency} details the performance and speed comparison.
In our experiments, DoLa~\cite{chuang2023dola} has minimal overhead compared to normal decoding, with only a 1.3$\times$ increase in latency.
DoLa is faster than \Ours, but \Ours shows better performance.
Despite implementation differences such as beam search, OPERA~\cite{huang2023opera} achieves slightly higher accuracy than \Ours, but our method is significantly faster than OPERA. 
There are trade-offs among the methods, but \Ours offers clear advantages.
It is conceptually and implementation-wise simple, applicable to various methods, and delivers a favorable speed and performance trade-off.
Also, it can be complementarily used with other contrastive decoding methods.

% \todo{
% remove?
% % \input{figures/06description_case}
% \noindent\textbf{\Ours in descriptive task.}
% We demonstrate how \Ours is effective in descriptive tasks such as CHAIR in \Cref{fig:descriptive_task}. In the case of standard decoding, the model assigns the highest probability to the token '\textit{S}' at the current timestep \textit{t}, leading to the incorrect prediction of "Sherlock Holmes." In contrast, \Ours, which utilizes both the original and augmented images, effectively adjusts the probability distribution and selects the token '\textit{Det}' rather than '\textit{S}', resulting in the correct prediction of "Detective Conan." This highlights the advantage of leveraging augmented images for probability correction, thereby improving accuracy in visually ambiguous contexts.
% }

\begin{figure}[t!]
    \begin{minipage}[t]{0.48\linewidth}
        \begin{center}
        \begin{small}
        % \vspace{\abovetabcapmargin}
        \setlength{\tabcolsep}{5pt} % base value: 6pt
        \scalebox{1}{
            \includegraphics[width=\linewidth]{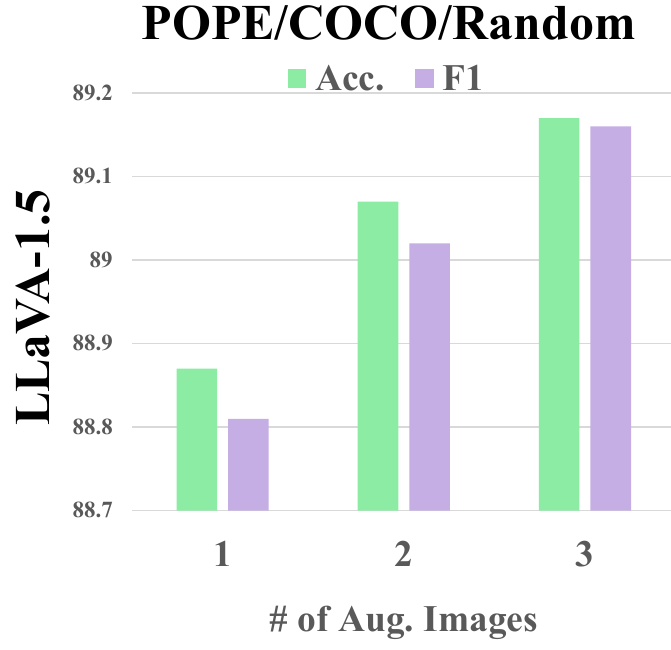} 
        }
        \vspace{\abovefigcapmargin}
        \caption{
            \textbf{Impact of the number of augmented images in \Ours.}
            % \todo{chart}
        }
        % \vspace{\belowfigcapmaxrgin}
        \label{fig:multiple_augmentation_main}
        \end{small}
        \end{center}    
    \end{minipage}%
    \hfill
    \begin{minipage}[t]{0.48\linewidth}
        \begin{center}
        \begin{small}
        % \vspace{\abovetabcapmargin}
        % \vspace{\belowtabcapmargin}
        \setlength{\tabcolsep}{5pt} % base value: 6pt
        \scalebox{1}{
            \includegraphics[width=\linewidth]{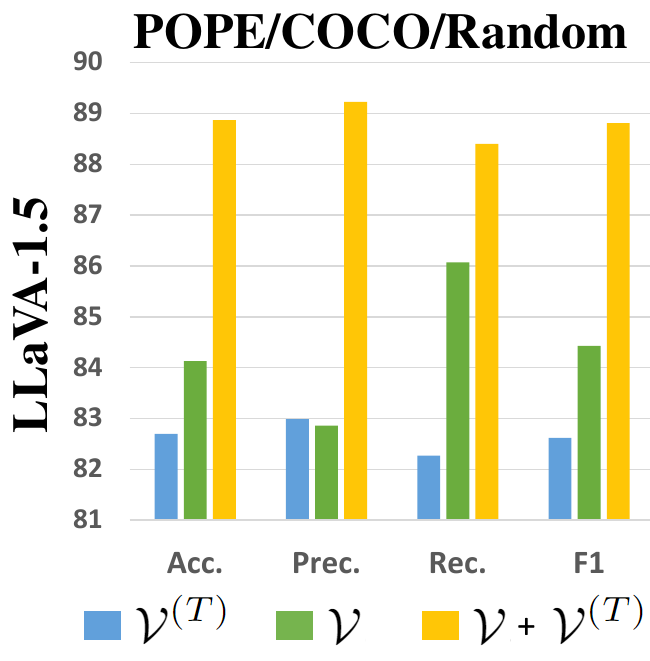} 
        }
        \vspace{\abovefigcapmargin}
        \vspace{-2mm}
        \caption{
            \textbf{
            Impact of combining original and transformed images.
            }
        }
        % \vspace{\belowfigcapmargin}
        \label{fig:transformed_only}
        \end{small}
        \end{center}    
    \end{minipage}%
    \vspace{-5mm}
\end{figure}

\noindent\textbf{Ablation of the number of augmented images.} 
To investigate whether increased exposure to diverse visual scenarios allows the model to better understand images and produce more robust responses, we conducted an ablation study by varying the number of augmented images in \Ours.
As shown in~\cref{fig:multiple_augmentation_main}, the performance slightly improves as more augmented images are used.
This improvement can be attributed to the richer visual context provided by the additional augmentations.
However, using multiple augmented images also introduces a trade-off, as it increases latency due to the additional computational load.\footnote{Detailed results are in Appendix.}

\noindent\textbf{Original \vs Transformed \vs Combined Images.} 
As shown in~\cref{fig:transformed_only}, the model’s performance declines when using only randomly transformed images ($\color{Blue}\mathcal{V}^{(T)}$) as input compared to using the original images ($\color{Green}\mathcal{V}$).
This drop in performance can likely be attributed to the introduction of visual artifacts and loss of essential cues, which disrupt the model’s contextual understanding.
In contrast, using both the original and transformed images together ($\color{Yellow}\mathcal{V} + \mathcal{V}^{(T)}$) significantly enhances the model’s performance.
This combined approach offers the model a richer, multiview representation, allowing it to leverage complementary perspectives from each view.
As a result, the model achieves better generalization, reduces hallucinated responses, and improves the likelihood of producing correct answers across various tasks.

\begin{wraptable}[9]{r}{0.4\linewidth}
\vspace{-7mm}
\begin{center}
\begin{small}
% \vspace{\abovetabcapmargin}
\captionof{table}{
\textbf{Compatibility w/ contrastive decoding.}
}
\vspace{-2mm}
\renewcommand{\arraystretch}{1} % line-height X 1.4
\setlength\tabcolsep{4pt} % base value: 6pt
\scalebox{0.92}{
    \begin{tabular}{ccc}
    \toprule
    \multirow{2}{*}{Method} & \multicolumn{2}{c}{LLaVA 1.5} \\
    \arrayrulecolor{gray} \cmidrule(lr){2-3}
    & Acc.\up          & F1\up            \\
    \arrayrulecolor{gray} \cmidrule(lr){1-3}
    \Ours               & 88.87         & 88.81         \\
    \midrule    
    + VCD                   & 89.07         & 88.81         \\
    + M3ID                  & 89.00         & 88.88          \\
    \bottomrule
    \end{tabular}
}
\vspace{\belowtabcapmargin}
\label{tab:compat}
\end{small}
\end{center}
\end{wraptable}

\noindent\textbf{Compatibility with contrastive decoding methods.} 
As shown in~\Cref{tab:compat}, combining \Ours with contrastive decoding methods like VCD and M3ID yields additional performance gains, underscoring the compatibility and complementary strengths of these approaches.
While contrastive decoding helps reduce inherent language biases, \Ours broadens the model’s visual perception by exposing it to diverse transformations and perspectives.
This synergy effectively mitigates object hallucinations, leading to notable improvements in accuracy and F1 scores, demonstrating the potential of integrating diverse decoding strategies to enhance LVLM reliability and comprehension.\footnote{Full results are in Appendix.}
% ~\cref{sec:appendix}.}
% \todo{얘 sup 말고 sec으로 나오네}
% \Ours achieves further performance improvement when incorporated with contrastive decoding methods (VCD and M3ID), indicating compatibility.
% This synergy between contrastive decoding, which aims to reduce language biases, and our approach, which captures a broader range of visual contexts through varying fields of view, effectively mitigates object hallucinations.

%du: VCD와 M3ID가 그랬듯, contrastive decoding을 통하여 LVLMs의 visual information에 대한 dependency를 극대화하고 text input의 불필요한 bias를 줄이는 논리는 우리의 Ritual 위에서 동시에 존재할 수 있음. 

\begin{figure}[t!]
    \centering
    % \vspace{-1mm}
    \includegraphics[width=\linewidth]{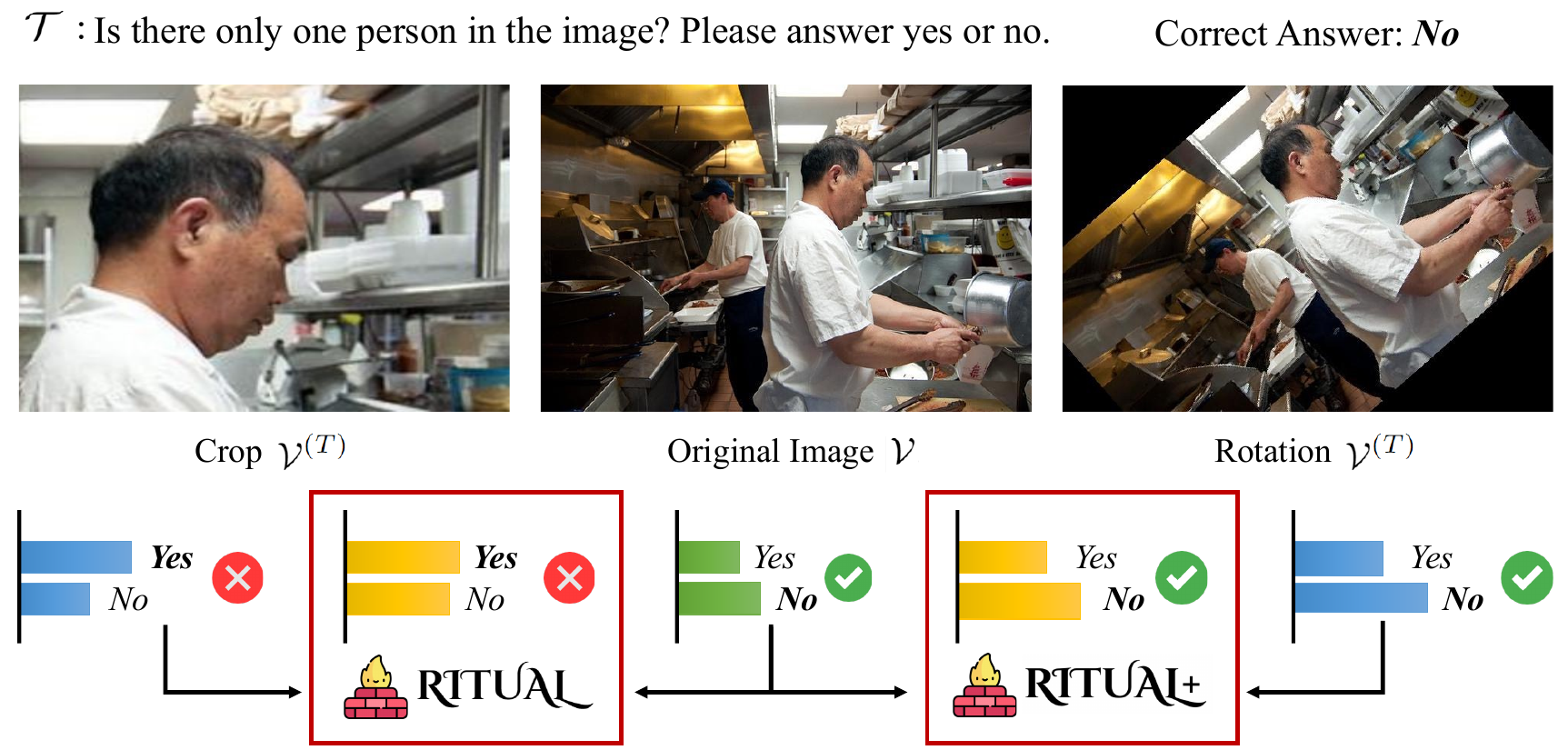}
    \vspace{\abovefigcapmargin}
    \vspace{-1mm}
    \caption{
        \textbf{Case study: \Title \vs \TitlePlus.}
        \Ours’s random transformations can miss key details like a person in the image, while \Oursplus adaptively selects certain transformation in context (\eg, rotation) to yield correct answers, accurately identifying multiple people in the images.
    }
    \label{fig:crop_failure}
    \vspace{\belowfigcapmargin}
\end{figure}

\noindent\textbf{Case study.} 
In \cref{fig:crop_failure}, we compare \Ours and \Oursplus in handling a query "Is there only one person in the image?".
\Ours, which applies transformations randomly, may occasionally lead to detrimental choices.
For example, performance may be impacted by the cropping area; in some cases, random cropping may inadvertently cut out important parts of the image, such as a person, resulting in poor outcomes.
In contrast, \Oursplus adapts to the query and image context, selecting transformations more strategically.
In this case, \Oursplus chooses rotation to interpret the image from a different angle, successfully identifying details that \Ours missed, leading to a correct response.

% \Oursplus demonstrates a modest improvement in performance by aligning transformations more closely with the specifics of each query.

% In future work, we aim to develop a more sophisticated mechanism that can more effectively determine the most suitable transformations based on the interplay between the image and its associated query.

% \begin{table*}[t!]

% \begin{minipage}[t!]{\textwidth}
%   \begin{minipage}[t!]{0.65\textwidth}
%     \centering
%     \includegraphics[width=.95\textwidth]{figures/failure cases.pdf}
%     \vspace{-1mm}
%     \captionof{figure}{\textbf{Case study on Crop image transformation.}
%         Performance can be affected by the cropping area.
%         The randomness of the selected region may sometimes lead to poor outcomes.
%     }
%   \label{fig:crop_failure}
%   \end{minipage}
%   \vspace{-1mm} 
%   \hfill
%   \begin{minipage}[t]{0.34\textwidth}
%     \vspace{-2cm}
%     \captionof{table}{
%         \textbf{Results of self-feedback augmentation selection on COCO random setup.}
%       }
%     \centering
%     \scalebox{0.9}{
%     \begin{tabular}{lcc}
%     \toprule
%     \textbf{\multirow{2}{*}{Method}} & \multicolumn{2}{c}{\textbf{LLaVA 1.5}} \\
%     \arrayrulecolor{gray} \cmidrule(lr){2-3}
%     & {{Acc.} {\up}} & {{F1} {\up}}\\
%     \midrule
%     \textit{base} & 84.13 & 84.43 \\
%     \textbf{\Ours} & \colorbox{SecondBest}{88.87} & \colorbox{SecondBest}{88.81} \\
%     \textbf{\Oursplus} & \colorbox{Best}{89.17} & \colorbox{Best}{89.21} \\
%     \bottomrule
%     \end{tabular}
%     }
%     \label{tab:self_feedback_main}
%     \end{minipage}
%   \end{minipage}
%    \vspace{-4mm} 
% \end{table*}

\vspace{\secmargin}\section{Conclusion}\vspace{\secmargin}
\label{sec:discussion}
We presented \OursEmoji, a simple decoding method that reduces hallucinations in LVLMs by incorporating randomly transformed images as complementary inputs.
To further enhance stability, \Oursplus adaptively selects transformations based on self-feedback, ensuring consistent performance across tasks.
Experiments show that both \Ours and \Oursplus outperform existing contrastive decoding methods on hallucination benchmarks and improve general vision-language understanding.
Our approach is training-free, model-agnostic, easy-to-implement, and requires no external models, yet it delivers strong performance.
This makes it a robust solution for enhancing LVLM accuracy and trustworthiness across diverse applications.
% We found that while relying solely on random image transformations can degrade performance, they contribute to mitigating hallucination when used in combination with the original image.
% Inspired by these findings, \Ours employs random image transformations to provide LVLMs with a broader visual context, thereby improving the model's robustness against hallucinatory outputs.
% \Ours significantly outperforms existing approaches on multiple hallucination benchmarks without requiring additional model training or complex external mechanisms.
% \todo{\Oursplus 내용 추가}
% \todo{remove, Moreover, \Ours is also compatible with existing contrastive decoding techniques, further enhancing performance.}
% In this paper, we treat the visual hallucination problems in Large Vision Language Models (LVLMs). 
% We found that relying solely on randomly transformed images leads to performance degradation because of visual artifacts, however, they contribute to mitigating hallucination when used in combination with the original image in probability distribution space.
% Inspired by this finding, we propose a novel hallucination mitigation method, \Ours, providing LVLMSs with more broad visual contexts along with randomly transformed images.
% Our approach improves the model's robustness by leveraging a varied view of visual input.
% Our extensive experiments show our superiority in reducing hallucination on various LVLMs and benchmarks.

% \vspace{-1mm}

\noindent\textbf{Limitations.}
% \todo{본문에 \Oursplus 넣었으니까 latency tradeoff, 등으로 수정하기?}
Like other contrastive decoding methods~\cite{leng2023mitigating,favero2024multi,chuang2023dola}, \Ours requires two forward passes, nearly doubling the latency compared to standard decoding.
\Oursplus, which involves additional self-feedback for adaptive transformation selection, requires three passes, resulting in approximately triple the latency.
This introduces a trade-off between improved hallucination mitigation and increased latency, which may impact usability in time-sensitive applications.

{\small
\bibliographystyle{ieeenat_fullname}
\bibliography{references}
}

\ifarxiv \clearpage \appendix \section*{\centering\LARGE Appendix} \addtocontents{toc}{\protect\setcounter{tocdepth}{2}}
\definecolor{linkcolor}{HTML}{000000}
% \newpage
% \section*{\centering\LARGE Appendix\todo{suppl로 이름 변경?}}
% \label{sec:appendix}
\tableofcontents
\newpage
\appendix
\definecolor{linkcolor}{HTML}{ED1C24}

\section{Extended Related Work}
\label{sec:appendix_related_work}
\vspace{\paramargin}\paragraph{Large Vision Language Models (LVLMs).}
Recent approaches to integrating visual and language modalities in LVLMs commonly leverage pre-trained uni-modal models.
They include an adaptive interface to bridge pre-trained visual encoders with Large Language Models (LLMs), facilitating efficient information synthesis across modalities.
These interfaces generally fall into two main categories:
(1) \textit{Learnable query-based methods}, exemplified by  Q-Former~\cite{li2023blip} in InstructBLIP~\cite{dai2024instructblip} and MiniGPT-4~\cite{zhu2023minigpt}, a set of learnable query tokens is employed to capture visual signals through cross-attention.
These tokens are optimized to distill the essential visual information and input it into the LLM for further processing.
mPLUG-Owl2~\cite{ye2024mplug} incorporates a visual abstractor that uses a predefined set of learnable queries to capture higher-level semantic features from images.
(2) \textit{Projection layer-based methods}, such as LLaVA~\cite{liu2023visual,liu2023improved} and Shikra~\cite{chen2023shikra}, use projection layers to transform visual features into the input space of LLMs.
This mapping ensures seamless integration between pre-trained visual representations and the LLMs, enabling the latter to interpret the visual content effectively.
Both strategies translate visual features into formats that the LLMs can understand.
Despite their efficacy, LVLMs still encounter challenges with hallucination, which we aim to mitigate in this work.
We specifically use three LVLMs, LLAVA, InstructBLIP, and mPLUG-Owl2, for experiments.

\begin{figure*}[t]
    \centering
    % \vspace{-1mm}
    \includegraphics[width=\textwidth]{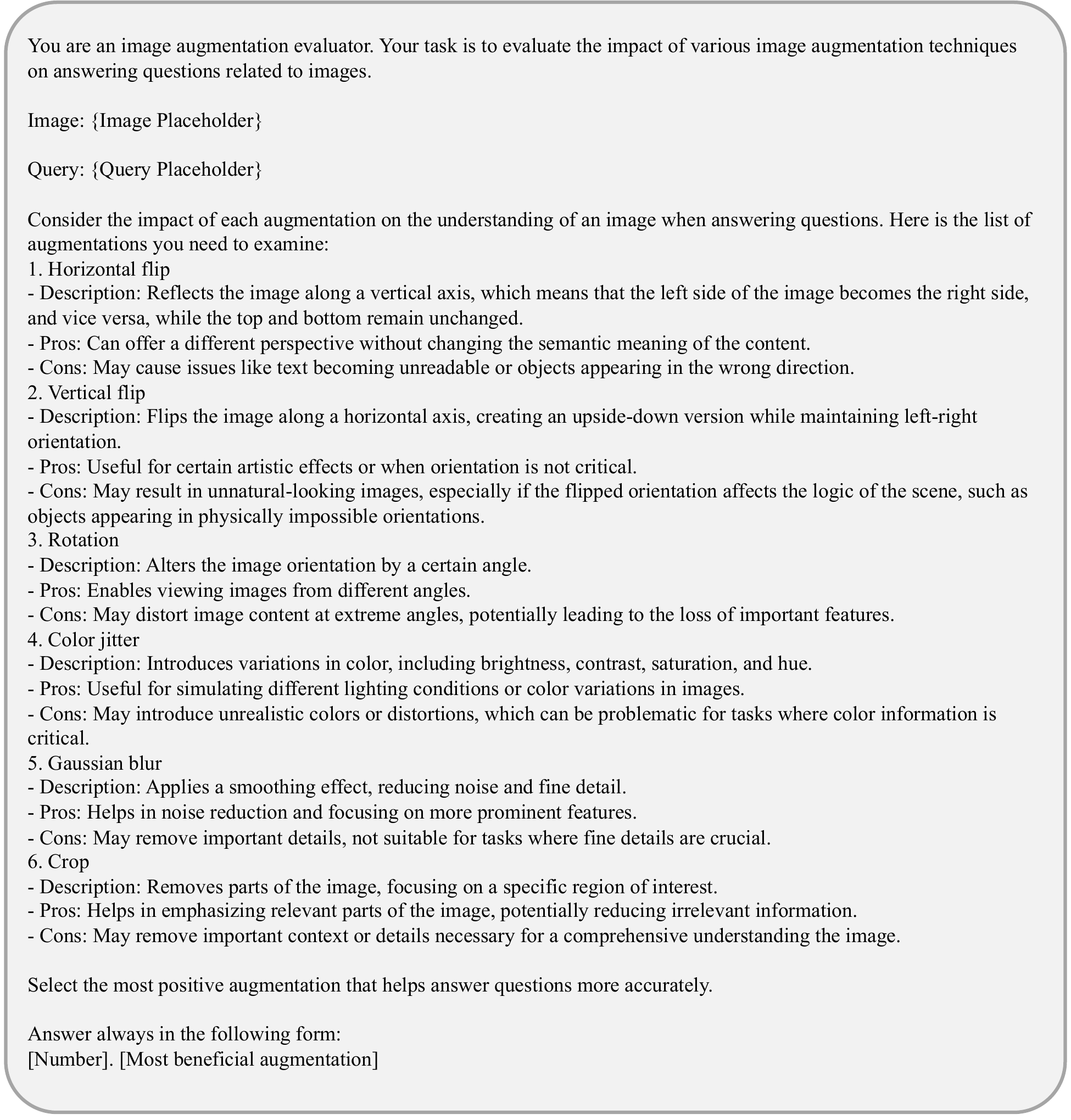}
    \vspace{\abovefigcapmargin}
    \caption{
    \textbf{Prompt for \TitlePlus.}
    }%
    \label{fig:ritual_plus}
    % \vspace{\belowfigcapmargin}
\end{figure*}

\vspace{\paramargin}\paragraph{Test-Time Augmentation (TTA).}
Test-Time Augmentation (TTA)~\cite{wang2018test,wang2019aleatoric,zhang2022memo,shanmugam2021better,perez2021enhancing} enhances model robustness and generalization during inference by utilizing multiple augmented versions of an input. By applying transformations such as rotations, flips, or noise, TTA reduces uncertainty and improves accuracy through prediction averaging or ensembling across these variations. This is especially beneficial for tasks with high input variability or noise, enabling the model to handle perturbations that could otherwise degrade performance.
By generating predictions for both the original and augmented inputs, TTA produces a more stable final output, mitigating the impact of noise and stabilizing predictions near decision boundaries~\cite{kimura2021understanding}. Unlike traditional ensembling~\cite{dietterich2000ensemble}, which requires multiple models, TTA leverages a single model, offering the benefits of an ensemble with minimal computational cost.
Our approach is similar to TTA in that we apply random transformations during inference. These augmentations broaden the model’s visual context, capturing a wider range of potential interpretations and reducing the risk of hallucinated outputs. By combining predictions from both the original and augmented inputs, we enhance robustness without additional training or complex architectures.

\begin{figure*}[h!]
    \centering
    % \vspace{-1mm}
    \includegraphics[width=\textwidth]{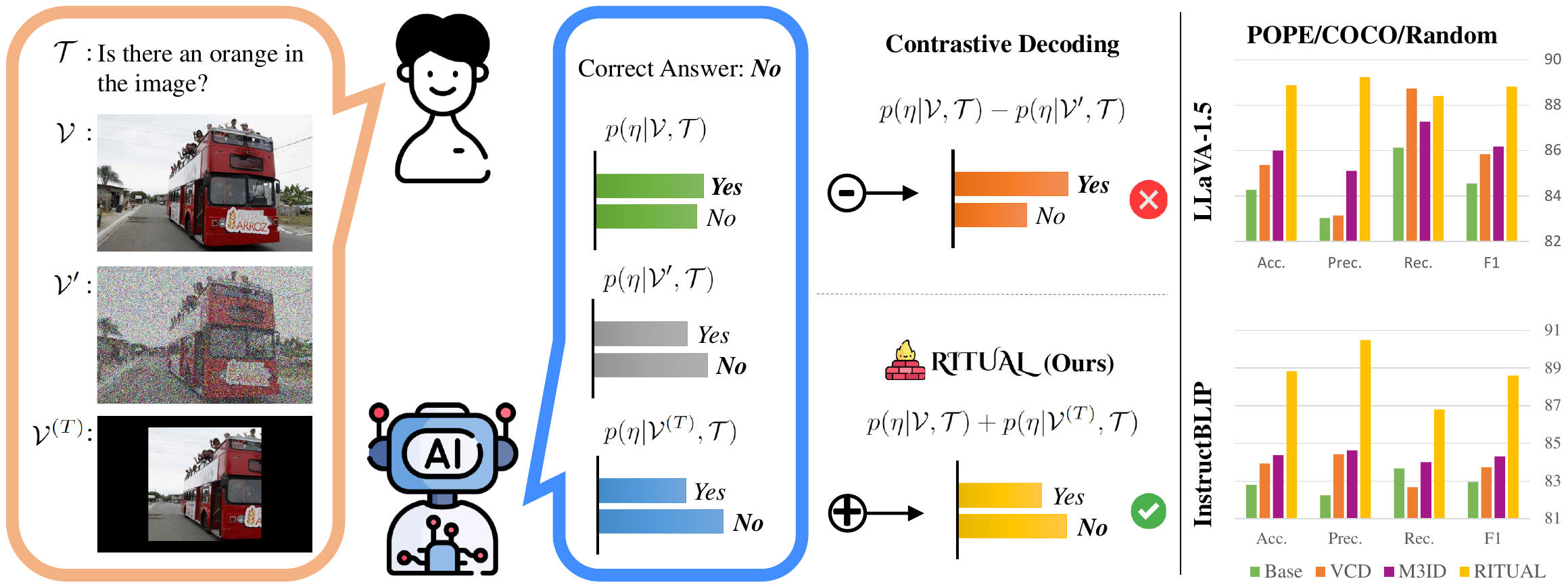}
    \vspace{\abovefigcapmargin}
    \caption{
    \textbf{Comparison of \TitleEmoji with contrastive decoding.}
    Unlike contrastive decoding methods~\citep{leng2023mitigating,favero2024multi}, which contrast the conditional probability given the original image ($\color{Green}\mathcal{V}$) to that given a diffused~\citep{leng2023mitigating} (or absent~\citep{favero2024multi}) image ($\color{Gray}\mathcal{V}'$), we leverage both the original image ($\color{Green}\mathcal{V}$) and a randomly transformed image ($\color{Blue}\mathcal{V}^{(T)}$) in a complementary manner.
    With latency similar to contrastive decoding, \Ours achieves state-of-the-art performance on multiple hallucination benchmarks.
    }%
    \label{fig:comparison}
    \vspace{\belowfigcapmargin}
    % \vspace{-2mm}
\end{figure*}

\section{Details of \TitlePlus}
\label{sec:appendix_self_feedback}
\Oursplus aims to address the limitations of random image transformations.
This extension is designed to minimize hallucinations and improve task-specific performance by dynamically tailoring image transformations to the query and task at hand.
% Below, we detail the workings of \Oursplus:

\noindent\textbf{Motivation.}
While the \Ours leverages random image transformations to provide diverse views, these transformations often have variable impacts on model predictions.
For example:
(1) Gaussian Blur obscures fine details;
% , making it unsuitable for tasks requiring precise recognition or small object boundaries.
(2) Crop reduces counting accuracy;
(3) Color Jitter negatively affects color-related tasks;
(4) Flips and Rotations disrupt positional understanding.
To mitigate these inconsistencies and improve overall reliability, \Oursplus employs a self-adaptive mechanism that evaluates and selects transformations based on their impact on the specific image-question pair.

\noindent\textbf{Key mechanism.}
The process begins with the LVLM receiving an input consisting of an image and a corresponding query, such as "How many objects are in the image?" Along with this input, the model is presented with a comprehensive list of potential image transformations.
Each transformation is described in detail, including its advantages and disadvantages.
For instance, Gaussian Blur can improve focus by reducing noise but may obscure fine details, while Crop might emphasize specific regions of interest but risks excluding essential information.

Using this information, the LVLM evaluates each transformation in the context of the given image and query.
It implicitly reasons through the pros and cons of the transformations, considering how they would affect its ability to generate an accurate response.
For example, in a counting task, Gaussian Blur might reduce noise and enhances focus on prominent features, while Crop could lead to errors by excluding parts of the image critical for the task.
Similarly, for positional reasoning tasks, the LVLM might reject transformations like Rotation or Flips, which could disrupt spatial orientation.
Once this implicit evaluation is complete, the LVLM selects the most suitable transformation.
% The decision is guided by the transformation’s ability to improve the model’s understanding of the image in relation to the query.
% For example, for tasks requiring clear object boundaries, the model might choose Gaussian Blur, while avoiding Color Jitter for tasks sensitive to color details.

% Gaussian Blur works by smoothing the image, reducing noise, and softening fine details. While this may seem counterintuitive for identifying precise object boundaries, the smoothing effect can actually make prominent features stand out by eliminating visual clutter caused by noise or irrelevant fine details. This can be particularly helpful in tasks where the LVLM struggles to distinguish objects due to excessive noise or pixel-level irregularities in the image.

This query-aware transformation selection ensures that transformations are not only tailored to the input but also aligned with the task requirements, improving reliability and reducing the potential for errors.
The structured reasoning process enables the model to adaptively select transformations that maximize task performance while minimizing disruptions caused by unsuitable transformations.

\noindent\textbf{Prompt design.}
As illustrated in the prompt provided to the LVLM (see~\cref{fig:ritual_plus}), the model uses the explicit descriptions of transformations and their effects to guide its reasoning.
By incorporating this self-adaptive approach, \Oursplus enhances the consistency and robustness of LVLM outputs, addressing the variability and unpredictability associated with random transformations.

% \todo{\Oursplus statistics?}

% Some image transformations may interfere with the model’s original predictions.
% To address this issue, we implemented a simple mechanism that allows the model to select an query-aware transformation through self-feedback.
% \todo{ritual plus prompt} in~\cref{fig:ritual_plus}
% \todo{
% the model receives an image-question pair along with a comprehensive description of transformations, after which it selects the most suitable transformation in a self-feedback manner.
% }
% Note that \Oursplus is the model with self-feedback transformation selection rather than random choice. 
% We compared the performance between \Ours and \Oursplus on POPE COCO setups in \Cref{tab:self_feedback}.
% \Oursplus declines in the popular setting while it achieves performance improvement in random and adversarial setups.
% Considering the computational complexity involved in the self-feedback process, the potential for performance improvement appears limited, suggesting the need for more advanced methodologies.

\section{\Title \vs Contrastive Decoding}
\label{sec:appendix_comparison_contrastive_decoding}
Contrastive decoding~\cite{leng2023mitigating, favero2024multi, zhang2024debiasing, wang2024mitigating, chuang2023dola} refines model outputs by contrasting two conditional probabilities: one more reliable and the other less reliable.
This is typically achieved by contrasting the conditional probability of textual responses given the original visual input with that given a distorted visual input.
The method aims to mitigate language biases or statistical priors, ensuring that responses are better grounded in the actual images, thereby reducing deviations from the visual truth.
% While beneficial, contrastive decoding does not fully resolve the misalignments between visual data and textual descriptions and can sometimes lead to the reinforcement of incorrect patterns.

While our approach also leverages two images as inputs, similar to contrastive decoding, we fundamentally differ in methodology.
Instead of negatively contrasting (subtracting) the two probability distributions, we integrate them in a complementary and positive manner.
Unlike contrastive decoding~\cite{leng2023mitigating, favero2024multi, wang2024mitigating}, which attributes hallucinations primarily to language biases or statistical priors, \Ours proposes that hallucinatory content may stem from the visual inputs themselves.
% By adopting a broader perspective on visual inputs, our method advocates for a multifaceted view of visual inputs.
% While our method uses the two images as inputs similar to contrastive decoding, our method does not negatively contrast (-) the two prob dist, rather we combine them in a positive manner.
% Our method is distinct from contrastive decoding~\cite{leng2023mitigating,favero2024multi, wang2024mitigating}, which attributes the causes of hallucinations to language bias or statistical priors.
% Instead, \Ours suggests that the source of hallucinatory content might actually reside within the images themselves, advocating for a multifaceted view of visual inputs.
The conceptual comparison is shown in~\cref{fig:comparison}.
In~\cref{sec:appendix_compatibility}, we also demonstrate that our method can be effectively combined with contrastive decoding techniques to achieve superior performance.
% \todo{compatability 내용 여기서 ref}

\section{Detailed Experimental Settings}
\label{sec:appendix_experimental_settings}

% exp detail
\noindent\textbf{POPE\footnote{\href{https://github.com/RUCAIBox/POPE}{https://github.com/RUCAIBox/POPE}}.}
We utilize the official benchmark from \cite{li2023evaluating}, which includes 3,000 question-answer pairs for each of the random, popular, and adversarial settings. We use the query template `Is there a \texttt{[object]} in the image?'. Here, \texttt{[object]} is selected randomly, from the most frequent objects in the dataset, or from objects that frequently co-occur with \texttt{[object]}, corresponding to the random, popular, and adversarial settings respectively.
We evaluate the performance based on whether the model-generated output contained the ground truth (`Yes' or `No') using accuracy, precision, recall, and average F1-score.

\noindent\textbf{MME\footnote{\href{https://github.com/BradyFU/Awesome-Multimodal-Large-Language-Models/tree/Evaluation}{https://github.com/BradyFU/Awesome-Multimodal-Large-Language-Models/tree/Evaluation}}.}
The MME~\cite{fu2024mme} dataset consists of 10 perception categories (existence, count, position, color, posters, celebrity, scene, landmark, artwork, OCR) and 4 recognition ones (commonsense reasoning, numerical calculation, text translation, code reasoning).
Each query is used with an image-related question followed by "Please answer yes or no."
We report the sum of accuracy at the query level and image level following the official implementation.

\noindent\textbf{CHAIR\footnote{\href{https://github.com/LisaAnne/Hallucination}{https://github.com/LisaAnne/Hallucination}}.}
We select 500 random images from the COCO~\cite{lin2014microsoft} validation set and generate the output using the query "Please describe this image in detail.". Due to the computational complexity, we restrict the \textit{max new tokens} to 64.
Following the M3ID~\cite{favero2024multi}, we report two assessment metrics, $C_s$ and $C_i$, which calculate the hallucination ratio per sentence and instance as follows:
\begin{align}
    &C_s  = \frac{|\text{\{sentences with hallucinated objects\}}|}{|\text{\{all sentences\}}|},\\
    &C_i = \frac{|\text{\{hallucinated objects\}}|}{|\text{\{all objects mentioned\}}|}. 
    \label{eq:chair_metric}
\end{align}

\noindent\textbf{LLaVA-Bench\footnote{\href{https://huggingface.co/datasets/liuhaotian/llava-bench-in-the-wild}{https://huggingface.co/datasets/liuhaotian/llava-bench-in-the-wild}}.}
The LLaVA-Bench~\cite{liu2023visual} dataset consists of 24 images along with 60 image-related questions.
This dataset is demanding as it has been collected from a variety of domains including diverse scenes, memes, paintings, sketches, and more.
We conduct qualitative case studies on this dataset to exhibit the efficacy of \Ours in challenging tasks and its adaptability to new domains.

\section{Further Implementation Details}
\label{sec:appendix_implementation_details}

\subsection{Image Transformations}

We set predefined six commonly used image transformations and randomly applied one of them for each image. We provide a concise description and implementation details below. We employ the Pytorch/Torchvision~\cite{paszke2019pytorch} implementation for transformation. 

\noindent\textbf{Horizontal flip} flips the image in the horizontal direction.

\noindent\textbf{Vertical flip} flips the image in the vertical direction.

\noindent\textbf{Rotation} rotates the image by a selected angle. The rotation angle is uniformly sampled from the \textit{degrees}=$(-180,+180)$.

\noindent\textbf{Color jitter} adjusts the brightness, contrast, saturation, and hue of the image.
We set \textit{brightness}=1, \textit{contrast}=1, \textit{saturation}=1, \textit{hue}=0.5.

\noindent\textbf{Gaussian blur} applies Gaussian blurring to the image with a chosen standard deviation \textit{sigma}. We set \textit{kernel$\_$size}=13 and \textit{sigma}=(1.5, 2.0).

\noindent\textbf{Random Resized Crop} randomly crops a region of the image and resizes it to a specified size. We set \textit{size}=336 as the same as the original data resize scale.

\subsection{Decoding Methods}

% Ours parameters
For a fair comparison, we adopt an adaptive plausible constraint based on the confidence level associated with the output distribution derived from the original visual inputs, following~\cite{li2022contrastive,leng2023mitigating}. The plausible constraint is defined as: 
% \begin{equation}
%     \mathcal{O}(\eta_{<t}) = \{ \eta_t \in \mathcal{O} : p_\theta\left(\eta_t \mid \mathcal{V}, \mathcal{T}, \eta_{<t}\right) \geq \beta \max _w p_\theta\left(w \mid v, x, y_{<t}\right)\}.\
% \label{eq:appendix_ritual_cd_sampling}
% \end{equation}
\begin{align}
    \mathcal{O}(\eta_{<t}) = \Big\{ \eta_t \in \mathcal{O} : &\; p_\theta\left(\eta_t \mid \mathcal{V}, \mathcal{T}, \eta_{<t}\right) \geq \beta \nonumber \\
    & \times \max _w p_\theta\left(w \mid v, x, y_{<t}\right) \Big\}. \label{eq:appendix_ritual_cd_sampling}
\end{align}
where $\mathcal{O}$ represents the output vocabulary of LVLM, and $\beta$ is a hyperparameter for the plausible constraint that adjusts truncation of the next token distribution.
The logits of tokens not in $\mathcal{O}$ are set $-\infty$, meaning that a larger $\beta$ retains only tokens with higher probabilities.
For all experiments, we set $\beta$= 0.1.
We configured the hyperparameter with a value of $\alpha=3$ in Eq. \textcolor{red}{4} by default.
% ~\cref{eq:ritual_sampling}
% VCD, M3ID parameters
Note that we reproduced VCD~\cite{leng2023mitigating} and M3ID~\cite{favero2024multi} under our experimental settings.
Specifically, we used the contrastive distribution of VCD as shown below:
% We use the contrastive distribution of VCD as shown in \cref{eq:appendix_vcd_sampling} and set the balancing parameter $\gamma$=2 and $\delta$=1, and the total noise step = 500 for generating the corrupted image $\mathcal{V}'$.
\begin{equation}
    \eta^{\rm VCD}_t \sim \gamma p_{\theta}(\eta_t | \mathcal{V}, \mathcal{T}, \eta_{<t}) - \delta p_{\theta}(\eta_t | \mathcal{V'}, \mathcal{T}, \eta_{<t}).
\label{eq:appendix_vcd_sampling}
\end{equation}
where $\mathcal{V'}$ represents a corrupted version of the original image $\mathcal{V}$.
We set the balancing parameters $\gamma = 2$, $\delta = 1$, and the total noise steps to 500 for generating $\mathcal{V'}$.

% Furthermore, we reproduced a key concept of M3ID, preventing conditioning dilution by introducing the unconditioned model as below:
For M3ID, a key concept to prevent conditioning dilution is reproduced by introducing an unconditioned model as follows:
% \begin{equation}
%     \eta^{M3ID}_t \sim p_{\theta}(\eta_t | \mathcal{V}, \mathcal{T}, \eta_{<t}) + \frac{1-e^{-\lambda t}}{e^{-\lambda t}} (p_{\theta}(\eta_t | \mathcal{V}, \mathcal{T}, \eta_{<t})-p_{\theta}(\eta_t |\mathcal{T}, \eta_{<t}))
% \end{equation}
\begin{align}
    \eta^{\rm M3ID}_t \sim p_{\theta}(\eta_t | \mathcal{V}, \mathcal{T}, \eta_{<t}) 
    + & \frac{1-e^{-\lambda t}}{e^{-\lambda t}} 
    \Big(p_{\theta}(\eta_t | \mathcal{V}, \mathcal{T}, \eta_{<t}) \nonumber \\
    & - p_{\theta}(\eta_t |\mathcal{T}, \eta_{<t}) \Big).
\end{align}
Here, $\lambda$ is a parameter balancing the conditioned and unconditioned models, set to 0.1 in our experiments.
% We set the $\lambda$, balancing parameter between conditioned model and unconditioned model, to $0.1$.
% 
For \Ours combined with contrastive decoding, we used a combined distribution:
\begin{align}
    \zeta\eta^{(T)}_t + \eta^{D}_t, \quad \text{where} \{{\rm VCD}, {\rm M3ID}\} \in D,
\end{align}
and $\eta^{(T)}_t =p_{\theta}(\eta_t | \mathcal{V}^{(T)}, \mathcal{T}, \eta_{<t})$. 
In this setup, we set $\gamma=1$, $\delta=0.1$, and $\zeta=3$ for \Ours with VCD, and $\lambda=0.1$ and $\zeta=3.5$ for \Ours with M3ID.
% When we use \Ours and contrastive decoding, we used combined distribution as $\zeta\eta^{(T)}_t + \eta^{D}_t$ where $\{{\rm VCD}, {\rm M3ID}\} \in D$.
% 
% 
% Furthermore, we used combined distribution to utilize \Ours with contrastive decoding as follows:
% \begin{equation}
%     \eta_t \sim p_{\theta}(\eta_t | \mathcal{V}, \mathcal{T}, \eta_{<t}) + \alpha p_{\theta}(\eta_t | \mathcal{V}^{(T)}, \mathcal{T}, \eta_{<t}) - \gamma p_{\theta}(\eta_t | \mathcal{V}', \mathcal{T}, \eta_{<t}) ,
% \label{eq:appendix_ritual_cd_sampling}
% \end{equation}
% where $\mathcal{V}'$ is diffused or absent image for contrastive decoding and $\gamma$ is the balancing hyperparameter.
% 
% Furthermore, we used combined distribution to ensure the compatibility of \Ours with contrastive decoding (\Ours+VCD and \Ours+M3ID).
% In this case, we set $\alpha=3$ for VCD and $\alpha=3.5$ for M3ID.
% exp env
%
In the case of DoLa~\cite{chuang2023dola}, we select the first bucket of candidate layers.
For OPERA~\cite{huang2023opera}, we set the scale factor to 50, the threshold to 15, the number of attention candidates to 5, penalty weights to 1, and the number of beams to 5.
The code is implemented in Python 3.10 with PyTorch 2.0.1~\cite{paszke2019pytorch}, and all experiments are conducted with a single NVIDIA RTX 3090 GPU.
% \todo{여기 parameter를 쓰기가 애매해서 걍 아예 숨기고 리부탈때 물어보면 말할까 싶기도, 같이 썼을때 parameter를 물어볼까 싶기도 하고 수식으로 나타내기 애매한 상태, 할거면 VCD/M3ID 각각 나눠서 적어야할듯}

\section{Additional Experiments}
\label{sec:appendix_additional_experiments}

\subsection{Random Image Transformation \vs Single Image Transformation}
\label{sec:appendix_random_vs_regular}
\begin{table}[ht!]
    \vspace{\abovetabcapmargin}
    \caption{
    \textbf{Performance of singular and random image transformations on POPE COCO benchmark.}
    }
    \vspace{\belowtabcapmargin}
    \centering
    \small
    \setlength{\tabcolsep}{4pt} % base value: 6pt
    \scalebox{0.8}{
    \begin{tabular}{y{10}y{40}y{65}x{30}x{30}x{30}x{30}}
        \toprule
         & \multirow{2}{*}{\textbf{Setup}} & \multirow{2}{*}{\textbf{Transformation}} & \multicolumn{4}{c}{\textbf{LLaVA 1.5~\citep{liu2023visual}}}  \\
        \arrayrulecolor{gray} \cmidrule(lr){4-7}
         &  &  & {{Acc.}{\up}} & {{Prec.}{\up}} & {{Rec.}{\up}} & {{F1}{\up}}  \\
        \midrule
        \multirow{21}{*}{\rot{\textbf{\normalsize MS-COCO~\citep{lin2014microsoft}\qquad\qquad}}} & \multirow{7}{*}{Random} 
        & Horizontal Flip & \colorbox{Best}{89.50} & \colorbox{SecondBest}{89.95} & \colorbox{Best}{88.93} & \colorbox{Best}{89.44}
     \\
         &  & Vertical Flip & 88.60 & 88.76 & \colorbox{SecondBest}{88.40} & 88.58
     \\
         &  & Rotate & \colorbox{SecondBest}{88.90} & 89.56 & 88.07 & \colorbox{SecondBest}{88.81}
     \\
         &  & Color Jitter & 88.83 & \colorbox{Best}{89.98} & 87.40 & 88.67
     \\
         &  & Gaussian Blur & 88.77 & 89.48 & 87.87 & 88.66
     \\
         &  & Crop & 88.47 & 89.36 & 87.33 & 88.33
     \\
         \arrayrulecolor{gray!50}\cmidrule(lr){3-7}
         &  & Random Selection & 88.87 & 89.23 & 88.40 & 88.58 \\
         \arrayrulecolor{gray}\cmidrule(lr){2-7}
          & \multirow{7}{*}{Popular} & Horizontal Flip & 85.60 & 83.21 & \colorbox{Best}{89.20} & 86.10
    \\
         &  & Vertical Flip & 85.23 & 83.05 & \colorbox{SecondBest}{88.53} & 85.71
     \\
         &  & Rotate & \colorbox{Best}{86.20} & \colorbox{SecondBest}{84.67} & 88.40 & \colorbox{Best}{86.50}
     \\
         &  & Color Jitter & \colorbox{Best}{86.20} & \colorbox{Best}{84.90} & 88.07 & \colorbox{SecondBest}{86.45}
     \\
         &  & Gaussian Blur & 84.93 & 83.29 & 87.40 & 85.30
     \\
         &  & Crop & \colorbox{SecondBest}{85.70} & 84.62 & 87.27 & 85.92
     \\
        \arrayrulecolor{gray!50}\cmidrule(lr){3-7}
        &   & Random Selection & 85.83 & 84.17 & 88.27 & 86.17 \\
         \arrayrulecolor{gray}\cmidrule(lr){2-7}
          & \multirow{7}{*}{Adversarial} & Horizontal Flip & \colorbox{SecondBest}{79.50} & 74.65 & \colorbox{Best}{89.33} & \colorbox{SecondBest}{81.34}
     \\
         &  & Vertical Flip & 79.10 & 74.65 & 88.13 & 80.83
     \\
         &  & Rotate & \colorbox{Best}{79.73} & \colorbox{SecondBest}{75.06} & \colorbox{SecondBest}{89.07} & \colorbox{Best}{81.46}
     \\
         &  & Color Jitter & 78.70 & 74.47 & 87.33 & 80.39 \\
         &  & Gaussian Blur & 78.73 & 74.19 & 88.13 & 80.56
     \\
         &  & Crop & 79.37 & \colorbox{Best}{75.48} & 87.00 & 80.83
     \\
        \arrayrulecolor{gray!50}\cmidrule(lr){3-7}
        &   & Random Selection & 78.80 & 74.43 & 87.73 & 80.54 \\
        \bottomrule
    \end{tabular}
    }
    \label{tab:random_augmentation_selection}
    \vspace{-2mm}
\end{table}

To generate transformed images $\mathcal{V}^{(T)}$, we randomly apply one of six image transformations: horizontal flip, vertical flip,  rotate, color jitter, Gaussian blur, or crop. 
We compare this random selection with a method that only adopts specific transformations rather than making a random choice.
As shown in~\Cref{tab:random_augmentation_selection}, the effectiveness of each transformation varies significantly depending on the POPE evaluation setup.
For instance, in the popular setup, applying color jitter exclusively achieves the best results across most metrics (Acc: 86.20, Prec: 84.90, Rec: 88.07, F1: 86.45).
In contrast, the same transformation delivers the poorest results in the adversarial setup, where it leads to lower F1 scores (80.39).
Similarly, transformations like horizontal flip, rotation, and Gaussian blur also demonstrate inconsistent impacts, being effective in one context while detrimental in another.
These results underscore the variability and task-specific nature of transformations.
The same transformation can yield either beneficial or harmful outcomes depending on the specific image-query pair and evaluation scenario.
To address this inherent randomness and its potential drawbacks, we propose \Oursplus, a self-adaptive framework that dynamically selects the most suitable transformation based on task requirements and feedback.
% , ensuring more reliable and consistent performance across diverse setups.
% The same transformation may have varying effects, beneficial or detrimental, based on the specific image and query.
% Therefore, we have chosen to use random selection as our primary method.
% \todo{이런 randomness 제거하기 위해 ritual+ 제안했다?}

\subsection{\Title on Larger LVLMs}
\label{sec:appendix_larger_models}
\begin{table}[ht!]
\vspace{\abovetabcapmargin}
\caption{\textbf{Results of 13B models on POPE COCO benchmark.}}
\vspace{\belowtabcapmargin}
\centering
\resizebox{\linewidth}{!}{
\begin{tabular}{llcccccccc}
\toprule
\multicolumn{1}{l}{\multirow{2}{*}{Setup}} & \multicolumn{1}{l}{\multirow{2}{*}{Method}} & \multicolumn{4}{c}{LLaVA-1.5 (13B)}                               & \multicolumn{4}{c}{InstructBLIP (13B)}                            \\ \cmidrule(lr){3-6} \cmidrule(lr){7-10}
\multicolumn{1}{l}{}                       & \multicolumn{1}{l}{}                        & Acc.           & Prec.          & Rec.           & F1             & Acc.           & Prec.          & Rec.           & F1             \\ \hline
\multirow{5}{*}{Random}                    & \textit{base}                                    & 82.70          & 78.73          & 89.60          & 83.82          & 80.10          & 75.21          & 89.80          & 81.86          \\
& VCD                                         & 82.97          & 79.00          & 89.80          & 84.06          & \colorbox{SecondBest}{82.83}    & \colorbox{Best}{78.65} & 90.13          & \colorbox{SecondBest}{84.00}    \\
& M3ID                                        & \colorbox{SecondBest}{84.53}    & \colorbox{SecondBest}{80.51}    & \colorbox{SecondBest}{91.13}    & \colorbox{SecondBest}{85.49}    & 81.57          & 76.56          & \colorbox{SecondBest}{91.00}    & 83.16          \\
& \Ours                                      & \colorbox{Best}{87.03} & \colorbox{Best}{83.69} & \colorbox{Best}{92.00} & \colorbox{Best}{87.65} & \colorbox{Best}{84.87} & \colorbox{SecondBest}{78.49}    & \colorbox{Best}{96.07} & \colorbox{Best}{86.39} \\ \hline
\multirow{5}{*}{Popular}                   & \textit{base}                                    & 80.93          & 76.95          & 88.33          & 82.25          & 75.80          & 70.14          & 89.87          & 78.78          \\
& VCD                                         & 80.23          & 75.58          & 89.33          & 81.88          & \colorbox{SecondBest}{77.43}    & \colorbox{Best}{71.56} & 91.07          & \colorbox{SecondBest}{80.14}    \\
& M3ID                                        & \colorbox{SecondBest}{81.57}    & \colorbox{SecondBest}{76.92}    & \colorbox{SecondBest}{90.20}    & \colorbox{SecondBest}{83.03}    & 76.43          & 70.22          & \colorbox{SecondBest}{91.80}    & 79.57          \\
& \Ours                                      & \colorbox{Best}{84.57} & \colorbox{Best}{80.20} & \colorbox{Best}{91.80} & \colorbox{Best}{85.61} & \colorbox{Best}{78.43} & \colorbox{SecondBest}{71.23}    & \colorbox{Best}{95.40} & \colorbox{Best}{81.56} \\ \hline
\multirow{5}{*}{Adversarial}               & \textit{base}                                    & 75.90          & 70.76          & 88.27          & 78.55          & 71.47          & \colorbox{SecondBest}{65.48}    & 90.80          & 76.09          \\
& VCD                                         & 75.63          & 69.83          & 90.27          & 78.74          & \colorbox{Best}{73.33} & \colorbox{Best}{67.45} & 90.20          & \colorbox{SecondBest}{77.18}    \\
& M3ID                                        & \colorbox{Best}{78.77} & \colorbox{Best}{73.09} & \colorbox{SecondBest}{91.07}    & \colorbox{Best}{81.09} & 71.40          & 65.29          & \colorbox{SecondBest}{91.40}    & 76.17          \\
& \Ours                                      & \colorbox{SecondBest}{77.93}    & \colorbox{SecondBest}{71.75}    & \colorbox{Best}{92.13} & \colorbox{SecondBest}{80.68}    & \colorbox{SecondBest}{72.37}    & 65.37          & \colorbox{Best}{95.13} & \colorbox{Best}{77.49} 
\\
\bottomrule
\end{tabular}
}
\label{tab:13B models}
\vspace{-2mm}
\end{table}

We report the results of the LLaVA-v1.5-13B and InstructBLIP-13B models on the POPE benchmark using the COCO dataset in \Cref{tab:13B models}. \Ours achieves the best overall performance across most metrics and settings, particularly excelling in the random and popular dataset types. Although its performance slightly falls short of VCD and M3ID under the adversarial setting, its superiority in other types suggests its robustness and effectiveness.

% \input{tables/15mplug_larger}
% Moreover, we extend our experiments to the additional larger LVLM, mPLUG-owl2~\cite{ye2024mplug}. As shown in \Cref{tab:mplug}. our proposed \Ours demonstrates the best performance in most cases on the POPE benchmark, similar to its success with LLaVa and InstructBLIP. This highlights the versatility and robustness of our approach across different LVLMs.

\begin{table*}[t!]
    \vspace{\abovetabcapmargin}
    \caption{
        \textbf{Compatibility with contrastive decoding on POPE benchmark~\cite{li2023evaluating}.}
    }
    \vspace{\belowtabcapmargin}
    \centering
    \small
    \setlength{\tabcolsep}{5pt}
    \scalebox{0.8}{
    \begin{tabular}{y{20}y{45}y{45}x{35}x{35}x{35}x{35}x{35}x{35}x{35}x{35}}
    \toprule
     & \multirow{2}{*}{\textbf{Setup}} & \multirow{2}{*}{\textbf{Method}} & \multicolumn{4}{c}{\textbf{LLaVA 1.5~\citep{liu2023visual}}} & \multicolumn{4}{c}{\textbf{InstructBLIP~\citep{dai2024instructblip}}} \\
    \arrayrulecolor{gray} \cmidrule(lr){4-7} \cmidrule(lr){8-11}
     &  &  & {{Acc.} {\up}} & {{Prec.} {\up}} & {{Rec.} {\up}} & {{F1} {\up}} & {{Acc.} {\up}} & {{Prec.} {\up}} & {{Rec.} {\up}} & {{F1} {\up}} \\
    \midrule
    \multirow{9}{*}{\rot{\textbf{\normalsize MS-COCO~\citep{lin2014microsoft}\quad}}} & \multirow{3}{*}{Random} 
     & \textbf{\Ours} & {88.87}	& {89.23} & \colorbox{SecondBest}{88.40} & {88.81} & {88.83} & {90.48} & \colorbox{SecondBest}{86.80} & {88.60} \\
     \arrayrulecolor{gray!50}\cmidrule(lr){3-11}
     &  & +VCD & \colorbox{Best}{89.07}	& \colorbox{SecondBest}{89.49}	& \colorbox{Best}{88.53}	& \colorbox{Best}{89.01}	& \colorbox{Best}{89.30}	& \colorbox{SecondBest}{90.85}	& \colorbox{Best}{87.40}	& \colorbox{Best}{89.09} \\
     &  & +M3ID & \colorbox{SecondBest}{89.00} & \colorbox{Best}{89.85}	& 87.93	& \colorbox{SecondBest}{88.88}	& \colorbox{SecondBest}{88.93}	& \colorbox{Best}{91.13}	& 86.27	& \colorbox{SecondBest}{88.63} \\
     \arrayrulecolor{gray}\cmidrule(lr){2-11}
      & \multirow{3}{*}{Popular} & \textbf{\Ours} & \colorbox{Best}{85.83} & \colorbox{Best}{84.17} & \colorbox{SecondBest}{88.27} & \colorbox{Best}{86.17} & \colorbox{SecondBest}{81.97} & {78.90} & \colorbox{Best}{87.27} & \colorbox{SecondBest}{82.87} \\
     \arrayrulecolor{gray!50}\cmidrule(lr){3-11} 
     &  & +VCD & \colorbox{SecondBest}{85.77} & \colorbox{SecondBest}{83.89} & \colorbox{Best}{88.53} & \colorbox{SecondBest}{86.15} & \colorbox{Best}{82.83} & \colorbox{Best}{80.16} & \colorbox{Best}{87.27} & \colorbox{Best}{83.56} \\
     &  & +M3ID & 85.37 & 83.60 & 88.00 & 85.74 & 81.90 & \colorbox{SecondBest}{78.98} & \colorbox{SecondBest}{86.93} & 82.77 \\
     \arrayrulecolor{gray}\cmidrule(lr){2-11}
      & \multirow{3}{*}{Adversarial} & \textbf{\Ours} & {78.80} & {74.43} & {87.73} & {80.54} & {78.73} & {74.57} & \colorbox{SecondBest}{87.20} & {80.39} \\
     \arrayrulecolor{gray!50}\cmidrule(lr){3-11} 
     &  & +VCD & \colorbox{Best}{79.60} & \colorbox{Best}{75.26} & \colorbox{Best}{88.20} & \colorbox{Best}{81.22} & \colorbox{Best}{79.07} & \colorbox{SecondBest}{74.89} & \colorbox{Best}{87.47} & \colorbox{Best}{80.69} \\
     &  & +M3ID & \colorbox{SecondBest}{79.20} & \colorbox{SecondBest}{74.83} & \colorbox{SecondBest}{88.00} & \colorbox{SecondBest}{80.88} & \colorbox{SecondBest}{78.93} & \colorbox{Best}{75.06} & 86.67 & \colorbox{SecondBest}{80.45} \\
     \arrayrulecolor{gray}\midrule
     \multirow{9}{*}{\rot{\textbf{\normalsize A-OKVQA~\citep{schwenk2022okvqa}\quad}}} & \multirow{3}{*}{Random}  & \textbf{\Ours} & \colorbox{SecondBest}{85.17} & {79.79} & \colorbox{SecondBest}{94.20} & \colorbox{SecondBest}{86.40} & \colorbox{SecondBest}{87.13} & \colorbox{SecondBest}{83.92} & \colorbox{Best}{91.87} & \colorbox{Best}{87.71} \\
     \arrayrulecolor{gray!50}\cmidrule(lr){3-11} 
     &  & +VCD & 85.10 & \colorbox{SecondBest}{79.93} & 93.73 & 86.28 & 86.77 & 83.57 & \colorbox{SecondBest}{91.53} & 87.37 \\
     &  & +M3ID & \colorbox{Best}{85.93} & \colorbox{Best}{80.62} & \colorbox{Best}{94.60} & \colorbox{Best}{87.06} & \colorbox{Best}{87.17} & \colorbox{Best}{84.35} & 91.27 & \colorbox{SecondBest}{87.67} \\
     \arrayrulecolor{gray}\cmidrule(lr){2-11}
      & \multirow{3}{*}{Popular} & \textbf{\Ours} & {78.83} & {71.99} & \colorbox{SecondBest}{94.40} & {81.68} & {78.73} & \colorbox{SecondBest}{72.83} & \colorbox{SecondBest}{91.67} & {81.17} \\
     \arrayrulecolor{gray!50}\cmidrule(lr){3-11}
     &  & +VCD & \colorbox{SecondBest}{79.17} & \colorbox{SecondBest}{72.40} & 94.27 & \colorbox{SecondBest}{81.90} & \colorbox{SecondBest}{78.83} & 72.75 & \colorbox{Best}{92.20} & \colorbox{SecondBest}{81.33} \\
     &  & +M3ID & \colorbox{Best}{79.63} & \colorbox{Best}{72.83} & \colorbox{Best}{94.53} & \colorbox{Best}{82.27} & \colorbox{Best}{79.20} & \colorbox{Best}{73.42} & 91.53 & \colorbox{Best}{81.48} \\
     \arrayrulecolor{gray}\cmidrule(lr){2-11}
      & \multirow{3}{*}{Adversarial} & \textbf{\Ours} & {68.57} & {62.26} & \colorbox{SecondBest}{94.27} & {74.99} & \colorbox{SecondBest}{70.27} & \colorbox{SecondBest}{64.15} & \colorbox{SecondBest}{91.87} & \colorbox{SecondBest}{75.55} \\
     \arrayrulecolor{gray!50}\cmidrule(lr){3-11}
     &  & +VCD & \colorbox{Best}{68.80} & \colorbox{Best}{62.48} & 94.13 & \colorbox{SecondBest}{75.11} & \colorbox{Best}{71.00} & \colorbox{Best}{64.72} & \colorbox{Best}{92.33} & \colorbox{Best}{76.10} \\
     &  & +M3ID & \colorbox{SecondBest}{68.77} & \colorbox{SecondBest}{62.42} & \colorbox{Best}{94.33} & \colorbox{Best}{75.13} & 69.30 & 63.43 & 91.13 & 74.80 \\
     \arrayrulecolor{gray}\midrule
     \multirow{9}{*}{\rot{\textbf{\normalsize GQA~\citep{hudson2019gqa}\quad}}} & \multirow{3}{*}{Random} & \textbf{\Ours} & {\colorbox{SecondBest}{86.10}} & \colorbox{SecondBest}{80.30} & \colorbox{Best}{95.67} & \colorbox{SecondBest}{87.31} & {84.87} & \colorbox{SecondBest}{82.52} & \colorbox{SecondBest}{88.47} & \colorbox{SecondBest}{85.39} \\
     \arrayrulecolor{gray!50}\cmidrule(lr){3-11}
     &  & +VCD & {86.03}	&80.21 & \colorbox{Best}{95.67} & 87.26 & \colorbox{SecondBest}{84.97} & 82.40 & \colorbox{Best}{88.93} & 85.54 \\
     &  & +M3ID & \colorbox{Best}{86.30} & \colorbox{Best}{80.64} & \colorbox{SecondBest}{95.53} & \colorbox{Best}{87.46} & \colorbox{Best}{85.00} & \colorbox{Best}{82.94} & 88.13 & \colorbox{SecondBest}{85.46} \\
     \arrayrulecolor{gray}\cmidrule(lr){2-11}
      & \multirow{3}{*}{Popular} & \textbf{\Ours} & \colorbox{SecondBest}{74.80} & \colorbox{SecondBest}{67.50} & \colorbox{Best}{95.67} & \colorbox{SecondBest}{79.15} & {74.50} & {69.17} & \colorbox{SecondBest}{88.40} & {77.61} \\
     \arrayrulecolor{gray!50}\cmidrule(lr){3-11}
     &  & +VCD & \colorbox{Best}{75.07} & \colorbox{Best}{67.82} & 95.40 & \colorbox{Best}{79.28} & \colorbox{SecondBest}{75.33} & \colorbox{SecondBest}{69.98} & 88.73 & \colorbox{SecondBest}{78.25} \\
     &  & +M3ID & {74.40} & 67.15 & \colorbox{SecondBest}{95.53} & 78.87 & \colorbox{Best}{75.57} & \colorbox{Best}{70.24} & \colorbox{Best}{88.73} & \colorbox{Best}{78.41} \\
     \arrayrulecolor{gray}\cmidrule(lr){2-11}
      & \multirow{3}{*}{Adversarial} & \textbf{\Ours} & {68.23} & {61.75} & \colorbox{Best}{95.80} & {75.10} & {70.17} & {64.76} & {88.47} & {74.78} \\
     \arrayrulecolor{gray!50}\cmidrule(lr){3-11}
     &  & +VCD & \colorbox{Best}{69.00} & \colorbox{Best}{62.39} & \colorbox{SecondBest}{95.67} & \colorbox{Best}{75.53} & \colorbox{SecondBest}{70.23} & \colorbox{SecondBest}{64.81} & \colorbox{SecondBest}{88.53} & \colorbox{SecondBest}{74.84} \\
     &  & +M3ID & \colorbox{SecondBest}{68.80} & \colorbox{SecondBest}{62.29} & 95.27 & \colorbox{SecondBest}{75.33} & \colorbox{Best}{71.00} & \colorbox{Best}{65.32} & \colorbox{Best}{89.53} & \colorbox{Best}{75.53} \\
    \bottomrule
    \end{tabular}
    }
    \label{tab:compatibility_pope}
\end{table*}
\begin{table*}[t]
    \begin{minipage}[t!]{0.66\textwidth}
    % \vspace{\abovetabcapmargin}
    \captionof{table}{
        \textbf{Compatibility with contrastive decoding on MME-Hallucination benchmark~\citep{fu2024mme}.}
    }
    \vspace{\belowtabcapmargin}
    \centering
    \small 
    \setlength{\tabcolsep}{2pt} % base value: 6pt
    \scalebox{0.76}{
    \begin{tabular}{y{60}y{45}x{60}x{60}x{60}x{60}x{60}}
        \toprule
        \multirow{2}{*}{\textbf{Model}} & \multirow{2}{*}{\textbf{Method}} & \multicolumn{2}{c}{\textbf{Object-level}} & \multicolumn{2}{c}{\textbf{Attribute-level}} & \multicolumn{1}{c}{\multirow{2}{*}{\textbf{\makecell{Total \\ Score}}}}\\
        \arrayrulecolor{gray} \cmidrule{3-4} \cmidrule{5-6} 
         & & Existence {\up} & Count {\up} & Position {\up} & Color {\up} & \\
        \arrayrulecolor{gray} \midrule
        \multirow{3}{*}{\textbf{LLaVA 1.5}}
         & \textbf{\Ours} & \colorbox{Best}{$187.50_{(\pm 2.89)}$} & $ 139.58_{(\pm 7.62)}$ & \colorbox{Best}{$ 125.00_{(\pm 10.27)}$} & \colorbox{SecondBest}{$ 164.17_{(\pm 6.87 )}$} & $ 616.25_{(\pm 20.38 )}$ \\
         \arrayrulecolor{gray!50}\cmidrule(lr){2-7}
         & +VCD & $ 185.00_{(\pm 4.08 )}$ & \colorbox{SecondBest}{$ 140.84_{(\pm 4.41 )}$} & \colorbox{Best}{$ 125.00_{(\pm 7.07 )}$} & \colorbox{Best}{$ 165.83_{(\pm 6.46 )}$} & \colorbox{SecondBest}{$ 616.67_{(\pm 11.14 )}$} \\
         & +M3ID & \colorbox{Best}{$ 187.50_{(\pm 2.89 )}$} & \colorbox{Best}{$ 141.25_{(\pm 9.85 )}$} & \colorbox{Best}{$ 125.00_{(\pm 10.27 )}$} & \colorbox{SecondBest}{$ 164.17_{(\pm 6.87 )}$} & \colorbox{Best}{$ 617.92_{(\pm 22.12  )}$} \\
        \arrayrulecolor{gray} \midrule
        \multirow{3}{*}{\textbf{InstructBLIP}} 
         & \textbf{\Ours} & \colorbox{SecondBest}{$ 182.50_{(\pm 6.45 )}$} & \colorbox{SecondBest}{$74.58 _{(\pm 5.99 )}$} & \colorbox{Best}{$ 67.08_{(\pm 10.31 )}$} & $ 139.17_{(\pm 0.96 )}$ & \colorbox{SecondBest}{$ 463.33_{(\pm 12.40 )}$} \\
         \arrayrulecolor{gray!50}\cmidrule(lr){2-7}
         & +VCD & \colorbox{Best}{$ 185.00_{(\pm 4.08 )}$} & \colorbox{Best}{$ 75.00_{(\pm 7.07 )}$} & $ 62.50_{(\pm 6.46 )}$ & \colorbox{Best}{$141.67 _{(\pm 6.53  )}$} & \colorbox{Best}{$ 464.17_{(\pm 9.07 )}$} \\
         & +M3ID & \colorbox{SecondBest}{$182.50_{(\pm 6.45 )}$} & \colorbox{SecondBest}{$ 74.58_{(\pm 2.84 )}$} & \colorbox{SecondBest}{$ 63.33_{(\pm 11.55 )}$} & \colorbox{SecondBest}{$140.42 _{(\pm2.10  )}$} & $460.83 _{(\pm 11.1 )}$ \\
        \bottomrule
    \end{tabular}
    }
    \label{tab:compatibility_mme}
    \end{minipage}
    \hfill
    \begin{minipage}[t!]{0.31\textwidth}
    % \vspace{\abovetabcapmargin}
    \captionof{table}{
        \textbf{Compatibility with contrastive decoding on CHAIR benchmark~\citep{rohrbach2018object}.}
    }
    \vspace{\belowtabcapmargin}
    \vspace{-3mm}
    \begin{center}
    \begin{small}
    \setlength{\tabcolsep}{4pt} % base value: 6pt
    \scalebox{0.78}{
    \begin{tabular}{llx{37}x{37}}
        \toprule
         & \textbf{Method} & CHAIR$_S${\down} & CHAIR$_I${\down} \\
        \arrayrulecolor{gray} \midrule
        \multirow{3}{*}{\textbf{LLaVA 1.5}}
         & \textbf{\Ours} & 20.6 & 6.9 \\
         \arrayrulecolor{gray!50}\cmidrule(lr){2-4}
         & +VCD & \colorbox{SecondBest}{20.0} & \colorbox{SecondBest}{6.8} \\
         & +M3ID & \colorbox{Best}{18.0} & \colorbox{Best}{5.7} \\
        \arrayrulecolor{gray} \midrule
        \multirow{3}{*}{\textbf{InstructBLIP}}
         & \textbf{\Ours} & 26.0 & 8.8 \\
         \arrayrulecolor{gray!50}\cmidrule(lr){2-4}
         & +VCD & \colorbox{SecondBest}{25.0} & \colorbox{SecondBest}{8.6} \\
         & +M3ID & \colorbox{Best}{23.4} & \colorbox{Best}{7.9} \\
        \bottomrule
    \end{tabular}
    }
    \end{small}
    \end{center}
    \label{tab:compatibility_chair}
    \end{minipage}
    \vspace{-3mm}
\end{table*}

\subsection{Compatibility of \Title with Contrastive Decoding Methods}
\label{sec:appendix_compatibility}
As shown in~\Cref{tab:compatibility_pope,tab:compatibility_mme,tab:compatibility_chair}, \Ours yields further performance improvement when incorporated with contrastive decoding methods, such as VCD~\citep{leng2023mitigating} and M3ID~\citep{favero2024multi}, across various benchmarks.
This compatibility demonstrates a synergy between the two approaches.
While contrastive decoding primarily mitigates language biases by contrasting conditional probabilities, \Ours enriches visual understanding by leveraging transformations to capture diverse visual contexts.
Together, these methods effectively address the problem of object hallucinations and improve model grounding.

 % (VCD and M3ID), indicating compatibility.
% This synergy between contrastive decoding, which aims to reduce language biases, and our approach, which captures a broader range of visual contexts through varying fields of view, effectively mitigates object hallucinations.

\subsection{Effect of $\alpha$ in \Title}
\label{sec:appendix_alpha}
\begin{table}[ht!]
\begin{center}
        \vspace{\abovetabcapmargin}
        \captionof{table}{
            \textbf{Impact of $\alpha$ on POPE~\citep{li2023evaluating} COCO random benchmark.}
            Based on the results, we set $\alpha=3$ as the default. 
        }
        \vspace{\belowtabcapmargin}
        \setlength{\tabcolsep}{6pt} % base value: 6pt
        \scalebox{0.87}{
        \begin{tabular}{x{24}x{32}x{32}x{32}x{32}}
            \toprule
            \multirow{2}{*}{$\alpha$} & \multicolumn{4}{c}{LLaVA 1.5~\citep{liu2023visual}} \\
            \arrayrulecolor{gray} \cmidrule(lr){2-5}
            & {Acc.{\up}} & {Prec.{\up}} & {Rec.{\up}} & {F1{\up}} \\
            \arrayrulecolor{gray} \midrule
            0~(\textit{base}) & 84.13 & 82.86 & 86.07 & 84.43 \\
            \arrayrulecolor{gray!50}\midrule
            0.5 & 87.73 & 87.04 & \colorbox{Best}{88.67} & 87.85 \\
            1 & 88.00 & 87.70 & \colorbox{SecondBest}{88.40} & 88.05 \\
            1.5 & 88.53 & 88.74 & 88.27 & 88.50 \\
            2 & 88.50 & 89.05 & 87.80 & 88.42 \\
            2.5 & 88.27 & 88.68 & 87.73 & 88.20 \\
            3 & \colorbox{Best}{88.87} & \colorbox{SecondBest}{89.23} & \colorbox{SecondBest}{88.40} & \colorbox{Best}{88.81} \\
            3.5 & \colorbox{SecondBest}{88.67} & \colorbox{Best}{89.40} & 87.73 & \colorbox{SecondBest}{88.56} \\
            \bottomrule
        \end{tabular}
        }

        \label{tab:ablation}
        \end{center}
        
    \vspace{-3mm}
\end{table}
In \Cref{tab:ablation}, we conduct an ablation study on the hyperparameter $\alpha$ in Eq. \textcolor{red}{4}, which adjusts the ratio between the output logits of the model conditioned on the original image $\mathcal{V}$ and the transformed image $\mathcal{V^{(T)}}$.
We vary $\alpha$ from 0 (standard decoding) to 3.5 on the POPE COCO random setting.
Our method consistently outperforms the baseline across a broad spectrum of $\alpha$ values, with accuracy improvement ranging from $+3.60$ to $+4.74$.
This demonstrates that our approach is robust and effective regardless of the specific hyperparameter value chosen.
Based on these results, we set $\alpha = 3$ as the default value.

\subsection{Impact of One-Word Constraint}
\label{sec:appendix_oneword}
\begin{table}[t!]
\vspace{\abovetabcapmargin}
\caption{\textbf{Impact of the one-word constraint on POPE COCO random benchmark.}
In constrained setup, we use additional query "Please answer this question in one word.".}
\vspace{\belowtabcapmargin}
\centering
\resizebox{\linewidth}{!}{
\begin{tabular}{clccccc}
\toprule
 & & \multicolumn{5}{c}{LLaVA 1.5~\citep{liu2023visual}} \\
\arrayrulecolor{gray} \cmidrule(lr){3-7}
\begin{tabular}[c]{@{}c@{}}One word\\ Constraint\end{tabular} & Method & \begin{tabular}[c]{@{}c@{}}Yes\\ Ratio\end{tabular} & \multicolumn{1}{l}{Acc.} & Prec. & Rec.  & F1    \\
\hline
\multirow{2}{*}{\cmark}                                         & \textit{base} & 39.90                                               & 83.29                    & \colorbox{Best}{92.13} & 72.80 & 81.33 \\
& VCD      & 40.97                                               & \colorbox{SecondBest}{87.73}                    & 91.42 & 83.28 & \colorbox{SecondBest}{87.16} \\
\hline
\multirow{5}{*}{\xmark}                                         & \textit{base} & 51.87                                               & 84.13                    & 82.86 & 86.07 & 84.43 \\
& VCD      & 53.37                                               & 85.37                    & 83.14 & \colorbox{Best}{88.73} & 85.84 \\
& M3ID     & \colorbox{SecondBest}{50.97}                                               & 86.00                    & 85.11 & 87.27 & 86.18 \\
& DoLa     & 51.23                                               & 85.97                    & 85.10 & 87.20 & 86.14 \\
& \Ours   & \colorbox{Best}{49.53}                                               & \colorbox{Best}{88.87}                    & 89.23 & \colorbox{SecondBest}{88.40} & \colorbox{Best}{88.81}
\\
\bottomrule
\end{tabular}

}
\vspace{-0.05cm}
\label{tab:performance gap}
\end{table}

One of our primary baseline methods, VCD~\cite{leng2023mitigating}, introduces an additional instruction at the end of each question: "Please answer this question with one word".
As shown in~\Cref{tab:performance gap}, this constraint biases the model towards shorter, more definitive answers, with a notable inclination towards "No" (resulting in a ~60\% No ratio).
In contrast, our evaluation setup removes this “one-word” constraint, allowing the model to generate more detailed responses that include explanations. This approach results in a more balanced distribution of “Yes” and “No” answers (approximately 50\% each). Rather than limiting the output to a single word for simplicity in evaluation, our method assesses whether the response contains a “Yes” or “No” alongside a supporting explanation.
Despite this adjustment, \Ours achieves the best performance on the primary metric, F1, highlighting its effectiveness.
% By removing the "one word" constraint, we aim to capture more nuanced and contextually rich responses from the model, which we believe provides a more comprehensive assessment of its capabilities.
Note that since there is no official implementation of M3ID, we reimplemented the method and reported its results based on our settings.
% To provide more context, we have included a~\Cref{tab:performance gap} that presents the performance metrics under different settings with the respective "Yes" ratios.

\subsection{Effect of Transformation Intensity on Model Performance}
\label{sec:appendix_gaussian_noise}
\begin{table}[ht!]
    \centering
    \vspace{\abovetabcapmargin}
    \caption{\textbf{Performance of \Title with Gaussian noise at different noise steps and VCD with Gaussian blur at different sigma values on POPE COCO random benchmark.}}
    \vspace{\belowtabcapmargin}
    \begin{small}
    \setlength{\tabcolsep}{6pt}
    \begin{subtable}[t]{\linewidth}
        \centering
        \caption{\textbf{\Title w/ Gaussian noise.}}
        \begin{tabular}{lcccc}
            \toprule
            \multirow{2}{*}{Noise Step} & \multicolumn{4}{c}{LLaVA 1.5~\citep{liu2023visual}} \\
            \arrayrulecolor{gray} \cmidrule(lr){2-5}
             & Acc. & Prec. & Rec. & F1 \\
            \midrule
            50 & 89.37 & 91.04 & 87.33 & 89.15   \\
            999 & 81.47 & 75.85 & 92.33 & 83.28   \\
            \bottomrule
        \end{tabular}
        \label{tab:ritual_w_noise}
    \end{subtable}%
    \vspace{1mm}
    \begin{subtable}[t]{\linewidth}
        \centering
        \caption{\textbf{VCD w/ Gaussian blur.}}
        \begin{tabular}{lcccc}
            \toprule
            \multirow{2}{*}{Sigma} & \multicolumn{4}{c}{LLaVA 1.5~\citep{liu2023visual}} \\
            \arrayrulecolor{gray} \cmidrule(lr){2-5}
             & Acc. & Prec. & Rec. & F1 \\
            \midrule
            0.5 & 83.77 & 83.61 & 84.00 & 83.80   \\
            100 & 85.13 & 86.45 & 83.33 & 84.86   \\
            \bottomrule
        \end{tabular}
        \label{tab:vcd_w_blur}
    \end{subtable}
    \end{small}
\end{table}

% \begin{table}[t!]
%     \begin{center}
%         \begin{small}
%         % \vspace{\abovetabcapmargin}
%         \captionof{table}{
%         \textbf{\Ours with Gaussian noise.}
%         }
%         \vspace{\belowtabcapmargin}
%         \setlength{\tabcolsep}{6pt} % base value: 6pt
%         \scalebox{1}{
%         \begin{tabular}{lcccc}
%             \toprule
%             Noise Step & Acc. & Prec. & Rec. & F1 \\
%             \arrayrulecolor{gray} \midrule
%             50 & 89.37 & 91.04 & 87.33 & 89.15   \\
%             999 & 81.47 & 75.85 & 92.33 & 83.28   \\
%             \bottomrule
%         \end{tabular}
%         \label{tab:ritual_w_noise}
%         }
%         \end{small}
%     \end{center}
%     % \vfill
%     \begin{center}
%         \begin{small}
%         \vspace{\abovetabcapmargin}
%         \captionof{table}{
%         \textbf{VCD with Gaussian blur.}
%         }
%         % \vspace{\belowtabcapmargin}
%         \scalebox{1}{
%         \begin{tabular}{lcccc}
%             \toprule
%             Sigma & Acc. & Prec. & Rec. & F1 \\
%             \arrayrulecolor{gray} \midrule
%             0.5 & 83.77 & 83.61 & 84.00 & 83.80   \\
%             100 & 85.13 & 86.45 & 83.33 & 84.86   \\
%             \bottomrule
%         \end{tabular}
        
%         }
%         \label{tab:vcd_w_blur}
%         \end{small}        
%     \end{center}     
%     \vspace{-3mm}
% \end{table}

In our work, we use standard image transformations (\eg, crop, flip, rotate, color jitter, and Gaussian blur) to enhance model robustness by generating diverse views~\cite{chen2020simple, grill2020bootstrap}.
The key principle is that applying these transformations at an \textit{appropriate intensity} creates diverse perspectives while preserving the underlying semantics of the image.

% \todo{
% We are curious about how Gaussian noise used in VCD~\cite{leng2023mitigating} 가 image transformation으로써 적용되면 효과가 어떨지? 다른 transformation이랑 비슷한 효과가 있을지?
% VCD에서는 이게 negatively impact the understanding of image contents하는데 이게 우리 방식에 적용되었을때 어떻게 효과가 다를지?
% VCD는 두 이미지를 contrastive하게 decoding (두 이미지에서 얻어진 prob dist를 뺌)하지만 우리는 complementary decoding하기 (두 이미지에서 얻어진 prob dist를 더함) 때문에 상반되는 방식임.
% }

Contrastive decoding methods, such as VCD~\cite{leng2023mitigating}, leverage Gaussian noise to distort images and contrast probability distributions between the original and distorted versions.
VCD applies high-intensity noise (\eg, diffusion noise steps of 500 or 999, where 1000 steps typically reduce an image to near-complete Gaussian noise).
In contrast, \Ours employs low to moderate-intensity transformations, combining the probability distributions of the original and transformed images in a complementary manner.

\noindent\textbf{\Title w/ Gaussian noise.}
To explore how Gaussian noise, as used in VCD, performs as an image transformation, we applied it within the \Ours framework on the POPE-COCO-random setup (\Cref{tab:ritual_w_noise}).
At low noise intensities (e.g., noise step = 50), Gaussian noise effectively generates diverse perspectives while preserving the image’s semantic integrity, leading to enhanced performance.
However, at high noise intensities (e.g., noise step = 999), the transformation overly distorts the image, degrading performance by obscuring its content.
These results highlight the dependency of Gaussian noise’s efficacy on its intensity: low levels promote beneficial diversity, whereas excessive noise impairs understanding.
%
% These results indicate that the efficacy of Gaussian noise as a transformation depends on its intensity.
% Light noise can provide beneficial diversity, while excessive noise distorts the image and harms performance.
% Similarly, while Gaussian blur can also act as a diverse view generator at moderate levels, excessive blur leads to image distortion and performance degradation.
%

% Gaussian noise at low intensity (noise step=50) acts as a form of multiview transformation, leading to positive outcomes.
% However, with a noise step of 999, the image became excessively distorted, impairing performance.

% the summary is that low-intensity Gaussian noise can serve as an effective transformation.
% However, as the noise level increases, it shifts from providing beneficial diversity to distorting the image, which negatively impacts performance.
% In brief, Gaussian blur can distort the image if applied too strongly, just as Gaussian noise can serve as a diverse view generator if applied lightly.
% It all comes down to the intensity.
% Applying Gaussian noise at a low level can indeed offer a diverse perspective without compromising the image's semantics. Conversely, excessive Gaussian blur can distort the image.

% The reason why Gaussian noise in VCD distorts the image rather than acting as a useful transformation boils down to the intensity of the application.

\noindent\textbf{VCD w/ Gaussian blur.}
We also evaluate VCD with Gaussian blur at different sigma values (\Cref{tab:vcd_w_blur}).
% As sigma increases, the blur becomes stronger, causing greater image distortion.
%
Low sigma value (sigma = 0.5) introduces minimal blur while preserving image semantics, whereas high sigma value (sigma = 100) causes significant distortion.
VCD contrasts the probability distributions of the original and distorted images to reduce language prior influence and enhance visual grounding.
Stronger blur shifts the focus to visual content, mitigating object hallucination and improving performance in visually grounded tasks.
% Since VCD operates by contrasting ~~~~
% high sigma Gaussian blur encourages the model to prioritize visual content over textual cues.
% This helps VCD mitigate object hallucination by emphasizing the visual component.
%
% In VCD, increased image distortion (via stronger blur) enhances the model’s focus on visual information over language, producing better results in scenarios where object hallucination needs to be reduced.

% In VCD, this increased distortion enhances the model's focus on the visual part of the image relative to the language part, helping to mitigate object hallucination.
% The stronger the image distortion, the greater the emphasis on the visual component.
% As a result, when the sigma value is set to 100, the distortion is more pronounced than at sigma 0.5, leading to a more substantial effect in VCD.

% In conclusion, both Gaussian noise and blur can provide diverse perspectives when applied moderately.
% However, if applied excessively, they are more likely to be perceived as distortions.

% While both Gaussian noise and Gaussian blur can act as effective image transformations when applied at moderate levels, excessive application of either introduces distortions that negatively impact the image semantics.
% The distinction lies in their intensity: light Gaussian noise or blur fosters diversity, while high-intensity applications distort the image and hinder understanding.

\subsection{Impact of the Number of Augmented Images in \Title}
\label{sec:multiple augmented images appendix}
\begin{table}[ht!]
\vspace{\abovetabcapmargin}
\caption{\textbf{Impact of the number of augmented images in \Title on POPE COCO benchmark.}}
\vspace{\belowtabcapmargin}
\centering
\resizebox{\linewidth}{!}{
\begin{tabular}{lccccc}
\toprule
\multirow{2}{*}{Setup}       & \multirow{2}{*}{\begin{tabular}[c]{@{}l@{}}\# of Aug.\\ Images\end{tabular}} & \multicolumn{4}{c}{LLaVA-1.5~\citep{liu2023visual}} \\ \cmidrule(lr){3-6}
&                                                                              & Acc.  & Prec. & Rec.  & F1    \\ \hline
\multirow{3}{*}{Random}      & 1                                                                            & 88.87 & 89.23 & 88.40 & 88.81 \\
& 2                                                                            & \colorbox{SecondBest}{89.07} & \colorbox{Best}{89.38} & \colorbox{SecondBest}{88.67} & \colorbox{SecondBest}{89.02} \\
& 3                                                                            & \colorbox{Best}{89.17} & \colorbox{SecondBest}{89.25} & \colorbox{Best}{89.07} & \colorbox{Best}{89.16} \\ \hline

\multirow{3}{*}{Popular}     & 1                                                                            & \colorbox{SecondBest}{85.83} & \colorbox{Best}{84.17} & \colorbox{SecondBest}{88.27} & \colorbox{SecondBest}{86.17} \\
& 2                                                                            & 85.37 & 83.85 & 87.60 & 85.69 \\
& 3                                                                            & \colorbox{Best}{86.20} & \colorbox{SecondBest}{84.11} & \colorbox{Best}{89.27} & \colorbox{Best}{86.61} \\ \hline

\multirow{3}{*}{Adversarial} & 1                                                                            & 78.80 & 74.43 & 87.73 & 80.54 \\
& 2                                                                            & \colorbox{Best}{79.10} & \colorbox{SecondBest}{74.56} & \colorbox{Best}{88.33} & \colorbox{Best}{80.87} \\
& 3                                                                            & \colorbox{SecondBest}{79.07} & \colorbox{Best}{74.63} & \colorbox{SecondBest}{88.07} & \colorbox{SecondBest}{80.80}
\\
\bottomrule
\end{tabular}
}
\label{tab:multiple augmentated images}
\end{table}

As shown in \Cref{tab:multiple augmentated images}, we found that performances slightly improve with the addition of more augmented images. This improvement is likely due to the increased variety of views available for the same scene, enhancing the model's generalization ability. However, it is important to note that this also leads to increased computational overhead due to the necessity of additional forward passes. Using multiple augmented images can indeed contribute to performance improvement, but it comes with the inherent trade-off of increased latency due to the additional computational cost.

\subsection{Detailed Performance on MME-Fullset}
\label{sec:appendix_mme_fullset}
\begin{table*}[t]
\caption{\textbf{Results on MME-Fullset~\citep{fu2024mme}.}}
\vspace{\belowtabcapmargin}
\centering
\small 
\setlength\tabcolsep{2pt} % base value: 6pt
\scalebox{0.64}{
\def\arraystretch{0.9}
\newcolumntype{K}{!{\color{white}\ }c}
\begin{tabular}{ccKKKKKKKKKKKKKKKKKK}
\toprule
\multirow{2}{*}{Task} & \multirow{2}{*}{Category} & \multicolumn{6}{c}{\textbf{LLaVA 1.5}~\citep{liu2023visual}} & \multicolumn{6}{c}{\textbf{InstructBLIP}~\citep{dai2024instructblip}} & \multicolumn{6}{c}{\textbf{mPLUG-Owl2}~\citep{ye2024mplug}}
\\ \arrayrulecolor{gray} \cmidrule(lr){3-8} \cmidrule(lr){9-14} \cmidrule(lr){15-20} 
& & \textit{base} & VCD & M3ID & DoLa & \Ours & \Oursplus & \textit{base} & VCD & M3ID & DoLa & \Ours & \Oursplus & \textit{base} & VCD & M3ID & DoLa & \Ours & \Oursplus \\ \midrule \vspace{0.1cm}
\multirow{18}{*}{\rot{\textbf{Perception}}}
& Existence & \shortstack{173.75\\ $_{(\pm4.79)}$} & \shortstack{{178.75}\\ $_{(\pm2.5)}$} & \shortstack{177.50\\ $_{(\pm6.45)}$} & \shortstack{174.58\\ $_{(\pm5.34)}$} & \cellcolor{SecondBest}\shortstack{{187.50} \\ $_{(\pm2.89)}$} & \cellcolor{Best}\shortstack{188.89\\ $_{(\pm6.74)}$} & \shortstack{{160.42}\\ $_{(\pm5.16)}$} & \shortstack{158.75\\ $_{(\pm7.25)}$} & \shortstack{158.33\\ $_{(\pm5.44)}$} & \shortstack{{162.08}\\ $_{(\pm5.34)}$} & \cellcolor{SecondBest}\shortstack{{182.50}\\ $_{(\pm6.45)}$} & \cellcolor{Best}\shortstack{{187.20}\\ $_{(\pm5.09)}$} & \shortstack{174.58 \\ $_{(\pm 4.17)}$} & \shortstack{170.00 \\ $_{(\pm 0.00)}$} & \shortstack{176.25 \\ $_{(\pm 4.79)}$} & \shortstack{175.00 \\ $_{(\pm 5.77)}$} & \cellcolor{SecondBest}\shortstack{185.00 \\ $_{(\pm 4.08)}$} & \cellcolor{Best}\shortstack{189.44 \\ $_{(\pm 5.09)}$} 
\\ \vspace{0.1cm}
& Count & \shortstack{121.67\\ $_{(\pm12.47)}$} & \shortstack{{126.25}\\ $_{(\pm10.4)}$} & \shortstack{124.17\\ $_{(\pm10.93)}$} & \shortstack{122.09\\ $_{(\pm11.73)}$} & \cellcolor{SecondBest}\shortstack{{139.58}\\ $_{(\pm7.62)}$} & \cellcolor{Best}\shortstack{{145.55}\\ $_{(\pm2.55)}$} & \shortstack{79.17\\ $_{(\pm8.22)}$} & \cellcolor{SecondBest}\shortstack{{90.75}\\ $_{(\pm3.11)}$} & \cellcolor{Best}\shortstack{{94.58}\\ $_{(\pm9.85)}$} & \shortstack{82.50\\ $_{(\pm6.16)}$} & \shortstack{74.58\\ $_{(\pm5.99)}$} & \shortstack{{88.89}\\ $_{(\pm13.47)}$} & \shortstack{155.42 \\ $_{(\pm 10.03)}$} & \shortstack{138.75 \\ $_{(\pm 6.44)}$} & \shortstack{157.92 \\ $_{(\pm 9.75)}$} & \shortstack{151.67 \\ $_{(\pm 5.61)}$} & \cellcolor{Best}\shortstack{159.58 \\ $_{(\pm 13.57)}$} & \cellcolor{SecondBest}\shortstack{159.45 \\ $_{(\pm 5.36)}$} 
\\ \vspace{0.1cm}
& Position & \shortstack{117.92\\ $_{(\pm3.69)}$} & \shortstack{{120.00}\\ $_{(\pm4.08)}$} & \shortstack{{120.00}\\ $_{(\pm7.07)}$} & \cellcolor{SecondBest}\shortstack{{122.09}\\ $_{(\pm2.10)}$} & \cellcolor{Best}\shortstack{{125.00}\\ $_{(\pm10.27)}$} & \shortstack{110.00 \\ $_{(\pm 21.86)}$} & \cellcolor{Best}\shortstack{{79.58}\\ $_{(\pm8.54)}$} & \shortstack{70.00\\ $_{(\pm15.81)}$} & \cellcolor{SecondBest}\shortstack{{72.50}\\ $_{(\pm17.03)}$} & \shortstack{78.75\\ $_{(\pm8.96)}$} & \shortstack{67.08\\ $_{(\pm10.31)}$} & \shortstack{72.22 \\ $_{(\pm 7.52)}$} & \shortstack{81.67 \\ $_{(\pm 14.72)}$} & \shortstack{81.25 \\ $_{(\pm 12.65)}$} & \shortstack{81.67 \\ $_{(\pm 14.72)}$} & \cellcolor{SecondBest}\shortstack{82.09 \\ $_{(\pm 14.17)}$} & \shortstack{77.50 \\ $_{(\pm 9.57)}$} & \cellcolor{Best}\shortstack{83.33 \\ $_{(\pm 20.48)}$} 
\\ \vspace{0.1cm}
& Color & \shortstack{149.17\\ $_{(\pm7.51)}$} & \shortstack{150.83\\ $_{(\pm11.01)}$} & \shortstack{{152.92}\\ $_{(\pm5.67)}$} & \shortstack{{149.17}\\ $_{(\pm4.19)}$} & \cellcolor{SecondBest}\shortstack{{164.17}\\ $_{(\pm6.87)}$} & \cellcolor{Best}\shortstack{173.89 \\ $_{(\pm 10.58)}$} & \shortstack{130.42\\ $_{(\pm17.34)}$} & \shortstack{{132.5}\\ $_{(\pm18.78)}$} & \shortstack{128.33\\ $_{(\pm14.72)}$} & \shortstack{{135.42}\\ $_{(\pm10.49)}$} & \cellcolor{SecondBest}\shortstack{{139.17}\\ $_{(\pm0.96)}$} & \cellcolor{Best}\shortstack{148.33 \\ $_{(\pm 10.93)}$} & \shortstack{141.25 \\ $_{(\pm 13.29)}$} & \shortstack{138.75 \\ $_{(\pm 5.51)}$} & \shortstack{142.50 \\ $_{(\pm 12.51)}$} & \shortstack{139.58 \\ $_{(\pm 5.51)}$} & \cellcolor{SecondBest}\shortstack{160.42 \\ $_{(\pm 4.59)}$} & \cellcolor{Best}\shortstack{162.22 \\ $_{(\pm 8.55)}$} 
\\ \vspace{0.1cm}
& Posters & \shortstack{124.24\\ $_{(\pm3.36)}$} & \shortstack{{129.34}\\ $_{(\pm4.11)}$} & \shortstack{120.49\\ $_{(\pm8.23)}$} & \shortstack{127.98\\ $_{(\pm5.51)}$} & \cellcolor{Best}\shortstack{{135.46}\\ $_{(\pm0.94)}$} & \cellcolor{SecondBest}\shortstack{133.79 \\ $_{(\pm 2.27)}$} & \shortstack{101.96\\ $_{(\pm1.5)}$} & \shortstack{{114.29}\\ $_{(\pm7.07)}$} & \shortstack{110.54\\ $_{(\pm0.62)}$} & \shortstack{105.10\\ $_{(\pm3.41)}$} & \cellcolor{SecondBest}\shortstack{{139.46}\\ $_{(\pm4.85)}$} & \cellcolor{Best}\shortstack{142.97 \\ $_{(\pm 9.91)}$} & \cellcolor{SecondBest}\shortstack{154.08 \\ $_{(\pm 3.24)}$} & \shortstack{150.79 \\ $_{(\pm 5.53)}$} & \shortstack{154.76 \\ $_{(\pm 4.01)}$} & \shortstack{150.45 \\ $_{(\pm 3.94)}$} & \cellcolor{Best}\shortstack{158.39 \\ $_{(\pm 2.60)}$} & \shortstack{141.61 \\ $_{(\pm 11.33)}$} 
\\ \vspace{0.1cm}
& Celebrity & \shortstack{115.44\\ $_{(\pm3.98)}$} & \cellcolor{Best}\shortstack{{124.78}\\ $_{(\pm6.23)}$} & \shortstack{113.9\\ $_{(\pm4.85)}$} & \shortstack{115.00\\ $_{(\pm8.20)}$} & \shortstack{{120.07}\\ $_{(\pm1.88)}$} & \cellcolor{SecondBest}\shortstack{122.16 \\ $_{(\pm 2.94)}$} & \shortstack{105.22\\ $_{(\pm2.23)}$} & \shortstack{{128.31}\\ $_{(\pm5.14)}$} & \shortstack{119.05\\ $_{(\pm5.01)}$} & \cellcolor{Best}\shortstack{{150.74}\\ $_{(\pm2.15)}$} & \shortstack{{134.63}\\ $_{(\pm4.19)}$} & \cellcolor{SecondBest}\shortstack{136.37 \\ $_{(\pm 9.67)}$} & \cellcolor{SecondBest}\shortstack{152.16 \\ $_{(\pm 4.19)}$} & \cellcolor{Best}\shortstack{158.33 \\ $_{(\pm 3.56)}$} & \cellcolor{SecondBest}\shortstack{152.16 \\ $_{(\pm 3.51)}$} & \shortstack{144.70 \\ $_{(\pm 1.06)}$} & \shortstack{147.06 \\ $_{(\pm 4.12)}$} & \shortstack{145.49 \\ $_{(\pm 1.67)}$} 
\\ \vspace{0.1cm}
& Scene & \shortstack{147.44\\ $_{(\pm6.26)}$} & \shortstack{152.69\\ $_{(\pm2.46)}$} & \cellcolor{SecondBest}\shortstack{{155.94}\\ $_{(\pm2.83)}$} & \shortstack{150.94\\ $_{(\pm1.21)}$} & \cellcolor{Best}\shortstack{{159.75}\\ $_{(\pm2.79)}$} & \shortstack{154.75 \\ $_{(\pm 3.25)}$} & \shortstack{130.19\\ $_{(\pm3.9)}$} & \shortstack{140.56\\ $_{(\pm2.92)}$} & \shortstack{{145.31}\\ $_{(\pm5.78)}$} & \shortstack{{147.75}\\ $_{(\pm4.98)}$} & \cellcolor{SecondBest}\shortstack{{158.63}\\ $_{(\pm2.62)}$} & \cellcolor{Best}\shortstack{165.75 \\ $_{(\pm 7.94)}$} & \shortstack{153.75 \\ $_{(\pm 2.14)}$} & \shortstack{150.33 \\ $_{(\pm 2.74)}$} & \shortstack{154.33 \\ $_{(\pm 1.38)}$} & \shortstack{154.08 \\ $_{(\pm 2.08)}$} & \cellcolor{SecondBest}\shortstack{159.67 \\ $_{(\pm 1.38)}$} & \cellcolor{Best}\shortstack{168.92 \\ $_{(\pm 8.63)}$} 
\\ \vspace{0.1cm}
& Landmark & \shortstack{133.31\\ $_{(\pm4.73)}$} & \shortstack{{136.00}\\ $_{(\pm7.35)}$} & \shortstack{133.81\\ $_{(\pm5.84)}$} & \shortstack{132.31\\ $_{(\pm6.20)}$} & \cellcolor{SecondBest}\shortstack{{157.81}\\ $_{(\pm2.19)}$} & \cellcolor{Best}\shortstack{161.25 \\ $_{(\pm 4.44)}$} & \shortstack{118.13\\ $_{(\pm6.37)}$} & \shortstack{{131.06}\\ $_{(\pm3.71)}$} & \shortstack{127.06\\ $_{(\pm7.17)}$} & \shortstack{126.31\\ $_{(\pm3.68)}$} & \cellcolor{SecondBest}\shortstack{{150.69}\\ $_{(\pm1.39)}$} & \cellcolor{Best}\shortstack{152.25 \\ $_{(\pm 10.90)}$} & \shortstack{145.92 \\ $_{(\pm 5.38)}$} & \shortstack{136.08 \\ $_{(\pm 4.93)}$} & \shortstack{146.75 \\ $_{(\pm 4.42)}$} & \shortstack{140.83 \\ $_{(\pm 2.27)}$} & \cellcolor{Best}\shortstack{156.17 \\ $_{(\pm 3.26)}$} & \cellcolor{SecondBest}\shortstack{152.17 \\ $_{(\pm 16.96)}$} 
\\ \vspace{0.1cm}
& Artwork & \shortstack{107.31\\ $_{(\pm2.61)}$} & \shortstack{110.50\\ $_{(\pm0.79)}$} & \shortstack{{111.69}\\ $_{(\pm0.92)}$} & \shortstack{107.25\\ $_{(\pm7.95)}$} & \cellcolor{SecondBest}\shortstack{{117.31}\\ $_{(\pm2.23)}$} & \cellcolor{Best}\shortstack{126.92 \\ $_{(\pm 6.21)}$} & \shortstack{91.44\\ $_{(\pm5.61)}$} & \shortstack{{102.75}\\ $_{(\pm4.24)}$} & \shortstack{98.44\\ $_{(\pm3.91)}$} & \cellcolor{Best}\shortstack{{117.44}\\ $_{(\pm4.31)}$} & \shortstack{{103.94}\\ $_{(\pm6.95)}$} & \cellcolor{SecondBest}\shortstack{113.42 \\ $_{(\pm 12.00)}$} & \shortstack{128.92 \\ $_{(\pm 0.80)}$} & \cellcolor{SecondBest}\shortstack{131.25 \\ $_{(\pm 1.15)}$} & \shortstack{130.42 \\ $_{(\pm 0.29)}$} & \shortstack{129.75 \\ $_{(\pm 0.43)}$} & \cellcolor{Best}\shortstack{133.08 \\ $_{(\pm 2.32)}$} & \shortstack{128.92 \\ $_{(\pm 4.73)}$} 
\\ 
& OCR & \shortstack{107.50\\ $_{(\pm13.99)}$} & \shortstack{98.13\\ $_{(\pm7.18)}$} & \shortstack{{112.50}\\ $_{(\pm10.21)}$} & \shortstack{97.50\\ $_{(\pm10.80)}$} & \cellcolor{Best}\shortstack{{121.25}\\ $_{(\pm6.29)}$} & \cellcolor{SecondBest}\shortstack{119.17 \\ $_{(\pm 10.41)}$} & \shortstack{{90.63}\\ $_{(\pm6.88)}$} & \shortstack{81.25\\ $_{(\pm6.61)}$} & \shortstack{78.75\\ $_{(\pm17.85)}$} & \shortstack{73.13\\ $_{(\pm8.00)}$} & \cellcolor{SecondBest}\shortstack{{93.75}\\ $_{(\pm8.29)}$} & \cellcolor{Best}\shortstack{111.67 \\ $_{(\pm 3.82)}$} & \shortstack{102.50 \\ $_{(\pm 7.50)}$} & \cellcolor{Best}\shortstack{110.00 \\ $_{(\pm 12.99)}$} & \shortstack{102.50 \\ $_{(\pm 7.50)}$} & \shortstack{100.00 \\ $_{(\pm 4.33)}$} & \shortstack{105.00 \\ $_{(\pm 4.33)}$} & \cellcolor{SecondBest}\shortstack{105.83 \\ $_{(\pm 9.46)}$} 
\\ \arrayrulecolor{gray} \cmidrule(lr){1-20} \vspace{0.1cm}
\multirow{7}{*}{\rot{\textbf{Recognition}}}
& \shortstack{Commonsense\\Reasoning} & \shortstack{99.82\\ $_{(\pm9.39)}$} & \shortstack{{108.04}\\ $_{(\pm2.36)}$} & \shortstack{107.32\\ $_{(\pm10.13)}$} & \shortstack{107.32\\ $_{(\pm8.98)}$} & \cellcolor{SecondBest}\shortstack{{115.54}\\ $_{(\pm4.92)}$} & \cellcolor{Best}\shortstack{119.52 \\ $_{(\pm 6.87)}$} & \shortstack{92.68\\ $_{(\pm8.64)}$} & \shortstack{92.86\\ $_{(\pm6.20)}$} & \shortstack{{96.43}\\ $_{(\pm9.70)}$} & \shortstack{{96.43}\\ $_{(\pm1.31)}$} & \cellcolor{Best}\shortstack{{109.11}\\ $_{(\pm8.17)}$} & \cellcolor{SecondBest}\shortstack{100.83 \\ $_{(\pm 28.10)}$} & \shortstack{118.33 \\ $_{(\pm 6.63)}$} & \shortstack{115.24 \\ $_{(\pm 1.80)}$} & \shortstack{117.62 \\ $_{(\pm 5.45)}$} & \shortstack{118.10 \\ $_{(\pm 5.46)}$} & \cellcolor{SecondBest}\shortstack{121.19 \\ $_{(\pm 4.76)}$} & \cellcolor{Best}\shortstack{128.79 \\ $_{(\pm 5.49)}$} 
\\ \vspace{0.1cm}
& \shortstack{Numerical\\Calculation} & \shortstack{60.00\\ $_{(\pm12.42)}$} & \shortstack{{63.75}\\ $_{(\pm8.54)}$} & \cellcolor{Best}\shortstack{{68.75}\\ $_{(\pm7.22)}$} & \shortstack{{64.38}\\ $_{(\pm12.64)}$} & \shortstack{52.50\\ $_{(\pm8.9)}$} & \cellcolor{SecondBest}\shortstack{66.67 \\ $_{(\pm 13.77)}$} & \shortstack{56.88\\ $_{(\pm15.6)}$} & \cellcolor{SecondBest}\shortstack{{64.38}\\ $_{(\pm6.25)}$} & \shortstack{60.63\\ $_{(\pm19.51)}$} & \shortstack{56.88\\ $_{(\pm11.97)}$} & \shortstack{{63.75}\\ $_{(\pm9.24)}$} & \cellcolor{Best}\shortstack{83.33 \\ $_{(\pm 15.07)}$} & \shortstack{43.33 \\ $_{(\pm 16.07)}$} & \shortstack{46.67 \\ $_{(\pm 10.41)}$} & \shortstack{43.33 \\ $_{(\pm 16.07)}$} & \cellcolor{SecondBest}\shortstack{48.33 \\ $_{(\pm 20.36)}$} & \shortstack{45.83 \\ $_{(\pm 8.78)}$} & \cellcolor{Best}\shortstack{75.00 \\ $_{(\pm 10.00)}$} 
\\ \vspace{0.1cm}
& \shortstack{Text\\Translation} & \shortstack{81.88\\ $_{(\pm13.13)}$} & \shortstack{77.50\\ $_{(\pm8.90)}$} & \cellcolor{SecondBest}\shortstack{{87.50}\\ $_{(\pm10.61)}$} & \shortstack{{81.25}\\ $_{(\pm8.78)}$} & \cellcolor{Best}\shortstack{{93.75}\\ $_{(\pm10.51)}$} & \cellcolor{SecondBest}\shortstack{87.50 \\ $_{(\pm 0.00)}$} & \shortstack{56.88\\ $_{(\pm17.49)}$} & \shortstack{66.25\\ $_{(\pm6.61)}$} & \shortstack{{72.50}\\ $_{(\pm12.75)}$} & \shortstack{{74.38}\\ $_{(\pm10.48)}$} & \cellcolor{Best}\shortstack{{89.38} \\ $_{(\pm12.48)}$} & \cellcolor{SecondBest}\shortstack{76.67 \\ $_{(\pm 8.78)}$}& \cellcolor{Best}\shortstack{90.00 \\ $_{(\pm 7.50)}$} & \shortstack{76.67 \\ $_{(\pm 15.07)}$} & \cellcolor{Best}\shortstack{90.00 \\ $_{(\pm 7.50)}$} & \cellcolor{SecondBest}\shortstack{89.17 \\ $_{(\pm 15.07)}$} & \shortstack{84.17 \\ $_{(\pm 7.64)}$} & \shortstack{85.00 \\ $_{(\pm 16.39)}$} 
\\ 
& \shortstack{Code\\Reasoning} & \shortstack{{64.38}\\ $_{(\pm25.93)}$} & \shortstack{63.75\\ $_{(\pm25.86)}$} & \shortstack{{64.38}\\ $_{(\pm25.93)}$} & \shortstack{{64.38}\\ $_{(\pm29.04)}$} & \cellcolor{SecondBest}\shortstack{{65.00}\\ $_{(\pm10.21)}$} & \cellcolor{Best}\shortstack{73.33 \\ $_{(\pm 6.29)}$} & \shortstack{63.75\\ $_{(\pm11.27)}$} & \cellcolor{SecondBest}\shortstack{{72.50}\\ $_{(\pm20.31)}$} & \cellcolor{Best}\shortstack{{78.13}\\ $_{(\pm15.33)}$} & \shortstack{70.00\\ $_{(\pm7.91)}$} & \shortstack{66.19 \\ $_{(\pm 8.61)}$} & \shortstack{70.00\\ $_{(\pm4.08)}$} & \shortstack{60.00 \\ $_{(\pm 10.90)}$} & \shortstack{62.50 \\ $_{(\pm 17.50)}$} & \shortstack{60.00 \\ $_{(\pm 10.90)}$} & \shortstack{57.50 \\ $_{(\pm 7.50)}$} & \cellcolor{Best}\shortstack{71.67 \\ $_{(\pm 14.43)}$} & \cellcolor{SecondBest}\shortstack{67.50 \\ $_{(\pm 16.39)}$}
\\
\bottomrule
\end{tabular}
}
\vspace{0.2cm}
\label{tab:MME_full_apendix}
\end{table*}

\Cref{tab:MME_full_apendix} presents the results on the MME-Fullset benchmark~\citep{fu2024mme}.
We compare the decoding methods applied to several LVLMs, including LLaVA-1.5~\cite{liu2023visual}, InstructBLIP~\cite{dai2024instructblip}, and mPLUG-Owl2~\cite{ye2024mplug}.
Across all tested models, \Ours and \Oursplus demonstrate consistent and significant improvements on most task categories, showcasing its effectiveness in enhancing LVLMs' ability to accurately interpret and analyze general visual contents.
% \Ours and \Oursplus achieve the highest scores on the majority of tasks, underscoring their robustness and adaptability.
\Ours delivers significant performance gains by enhancing the models’ ability to interpret and analyze visual content accurately. \Oursplus further boosts results through adaptive transformation selection, showcasing its ability to tailor transformations for specific tasks and use cases.
% By enriching the model’s understanding with diverse visual contexts, \Ours provides balanced performance gains across a wide range of tasks, establishing itself as a robust and flexible method for improving LVLM performance.
% \Oursplus further enhances these results, showing that adaptive transformation selection improves performance even on more general tasks, confirming the benefit of tailored transformation for varied use cases.
In perception tasks, \Ours and \Oursplus outperform baseline methods in categories such as Existence, Count, and Landmark. In recognition tasks, they excel in Commonsense Reasoning and Text Translation, achieving top scores across multiple LVLMs.

\subsection{Confusion Matrices on POPE benchmark}
\label{sec:appendix_confusion_matrices}
\begin{table*}[t!]
    \caption{
        \textbf{Confusion matrices on POPE~\citep{li2023evaluating} benchmark.}
    }
    \vspace{\belowtabcapmargin}
    \centering
    \small
    \setlength{\tabcolsep}{4pt} % base value: 6pt
    \scalebox{0.8}{
    \begin{tabular}{y{20}y{45}y{65}x{35}x{35}x{35}x{35}x{35}x{35}x{35}x{35}x{35}x{35}}
    \toprule
     & \multirow{4}{*}{\textbf{Setup}} & \multirow{4}{*}{\textbf{Method}} & \multicolumn{5}{c}{\textbf{LLaVA 1.5~\citep{liu2023visual}}} & \multicolumn{5}{c}{\textbf{InstructBLIP~\citep{dai2024instructblip}}} \\
    \arrayrulecolor{gray} \cmidrule(lr){4-8} \cmidrule(lr){9-13}
     &  &  & TP {\up} & FP {\down} & TN {\up} & FN {\down} & {{Acc.} {\up}} & TP {\up} & FP {\down} & TN {\up} & FN {\down} & {{Acc.} {\up}} \\
    \midrule
    \multirow{12}{*}{\rot{\textbf{\normalsize MS-COCO~\citep{lin2014microsoft}\quad}}} & \multirow{4}{*}{Random} 
    & \textit{base} & 1291 & 267 & 1233 & 209 & 84.13 & 1255 & 271 & 1229 & 245 & 82.80 \\
     &  & VCD & 1331 & 270 & 1230 & 169 & 85.37 & 1240 & 222 & 1278 & 260 & 83.93
 \\
     &  & M3ID & 1309 & 229 & 1271 & 191 & 86.00 & 1260 & 229 & 1271 & 240 & 84.37
 \\
     &  & \textbf{\Ours} & 1326 & 160 & 1340 & 174 & 88.87 & 1302 & 137 & 1363 & 198 & 88.83
 \\
     % \arrayrulecolor{gray!50}\cmidrule(lr){3-13}
     % &  & \textbf{\Ours}+VCD & 1323 & 154 & 1346 & 177 & 89.07 & 1311 & 132 & 1368 & 189 & 89.30\\
     % &  & \textbf{\Ours}+M3ID & 1319 & 149 & 1351 & 181 & 89.00 & 1294 & 126 & 1374 & 206 & 88.93\\
     \arrayrulecolor{gray}\cmidrule(lr){2-13}
      & \multirow{4}{*}{Popular} & \textit{base} & 1283 & 357 & 1143 & 217 & 80.87 & 1238 & 464 & 1036 & 262 & 75.80
\\
     &  & VCD & 1306 & 373 & 1127 & 194 & 81.10 & 1234 & 402 & 1098 & 266 & 77.73
 \\
     &  & M3ID & 1324 & 339 & 1161 & 176 & 82.83 & 1259 & 440 & 1060 & 241 & 77.30
 \\
     &  & \textbf{Ours} & 1324 & 249 & 1251 & 176 & 85.83 & 1309 & 350 & 1150 & 191 & 81.97
 \\
     % \arrayrulecolor{gray!50}\cmidrule(lr){3-13}
     % &  & \textbf{\Ours}+VCD & 1328 & 255 & 1245 & 172 & 85.77 & 1309 & 324 & 1176 & 191 & 82.83\\
     % &  & \textbf{\Ours}+M3ID & 1320 & 259 & 1241 & 180 & 85.37 & 1304 & 347 & 1153 & 196 & 81.90\\
     \arrayrulecolor{gray}\cmidrule(lr){2-13}
      & \multirow{4}{*}{Adversarial} & \textit{base} & 1298 & 511 & 989 & 202 & 76.23 & 1263 & 501 & 999 & 237 & 75.40
 \\
     &  & VCD & 1308 & 540 & 960 & 192 & 75.60 & 1253 & 449 & 1051 & 247 & 76.80
 \\
     &  & M3ID & 1310 & 479 & 1021 & 190 & 77.70 & 1259 & 478 & 1022 & 241 & 76.03
 \\
     &  & \textbf{\Ours} & 1316 & 452 & 1048 & 184 & 78.80 & 1308 & 446 & 1054 & 192 & 78.73
 \\
     % \arrayrulecolor{gray!50}\cmidrule(lr){3-13}
     % &  & \textbf{\Ours}+VCD & 1323 & 435 & 1065 & 177 & 79.60 & 1312 & 440 & 1060 & 188 & 79.07 \\
     % &  & \textbf{\Ours}+M3ID & 1320 & 444 & 1056 & 180 & 79.20 & 1300 & 432 & 1068 & 200 & 78.93 \\
     \arrayrulecolor{gray}\midrule
     \multirow{12}{*}{\rot{\textbf{\normalsize A-OKVQA~\citep{schwenk2022okvqa}\quad}}} & \multirow{4}{*}{Random} & \textit{base} & 1373 & 421 & 1079 & 127 & 81.73 & 1300 & 366 & 1134 & 200 & 81.13
 \\
     &  & VCD & 1405 & 450 & 1050 & 95 & 81.83 & 1297 & 337 & 1163 & 203 & 82.00
 \\
     &  & M3ID & 1407 & 400 & 1100 & 93 & 83.57 & 1357 & 387 & 1113 & 143 & 82.33
 \\
     &  & \textbf{\Ours} & 1413 & 358 & 1142 & 87 & 85.17 & 1378 & 264 & 1236 & 122 & 87.13
 \\
     % \arrayrulecolor{gray!50}\cmidrule(lr){3-13}
     % &  & \textbf{\Ours}+VCD & 1406 & 353 & 1147 & 94 & 85.10 & 1373 & 270 & 1230 & 127 & 86.77 \\
     % &  & \textbf{\Ours}+M3ID & 1419 & 341 & 1159 & 81 & 85.93 & 1369 & 254 & 1246 & 131 & 87.17 \\
     \arrayrulecolor{gray}\cmidrule(lr){2-13}
      & \multirow{4}{*}{Popular} & \textit{base} & 1375 & 575 & 925 & 125 & 76.67 & 1303 & 533 & 967 & 197 & 75.67
 \\
     &  & VCD & 1393 & 652 & 848 & 107 & 74.70 & 1314 & 519 & 981 & 186 & 76.50
 \\
     &  & M3ID & 1416 & 551 & 949 & 84 & 78.83 & 1375 & 513 & 987 & 125 & 78.73
 \\
     &  & \textbf{\Ours} & 1416 & 551 & 949 & 84 & 78.83 & 1375 & 513 & 987 & 125 & 78.73
 \\
     % \arrayrulecolor{gray!50}\cmidrule(lr){3-13}
     % &  & \textbf{\Ours}+VCD & 1414 & 539 & 961 & 86 & 79.17 & 1383 & 518 & 982 & 117 & 78.83 \\
     % &  & \textbf{\Ours}+M3ID & 1418 & 529 & 971 & 82 & 79.63 & 1373 & 497 & 1003 & 127 & 79.20 \\
     \arrayrulecolor{gray}\cmidrule(lr){2-13}
      & \multirow{4}{*}{Adversarial} & \textit{base} & 1369 & 847 & 653 & 131 & 67.40 & 1302 & 762 & 738 & 198 & 68.00
 \\
     &  & VCD & 1400 & 877 & 623 & 100 & 67.43 & 1327 & 707 & 793 & 173 & 70.67
 \\
     &  & M3ID & 1404 & 861 & 639 & 96 & 68.10 & 1326 & 739 & 761 & 174 & 69.57
 \\
     &  & \textbf{\Ours} & 1414 & 857 & 643 & 86 & 68.57 & 1378 & 770 & 730 & 122 & 70.27
 \\
     % \arrayrulecolor{gray!50}\cmidrule(lr){3-13}
     % &  & \textbf{\Ours}+VCD & 1412 & 848 & 652 & 88 & 68.80 & 1385 & 755 & 745 & 115 & 71.00 \\
     % &  & \textbf{\Ours}+M3ID & 1415 & 852 & 648 & 85 & 68.77 & 1367 & 788 & 712 & 133 & 69.30 \\
     \arrayrulecolor{gray}\midrule
     \multirow{12}{*}{\rot{\textbf{\normalsize GQA~\citep{hudson2019gqa}\quad}}} & \multirow{4}{*}{Random} & \textit{base} & 1390 & 453 & 1047 & 110 & 81.23 & 1289 & 391 & 1109 & 211 & 79.93
 \\
     &  & VCD & 1426 & 481 & 1019 & 74 & 81.50 & 1300 & 345 & 1155 & 200 & 81.83
 \\
     &  & M3ID & 1417 & 432 & 1068 & 83 & 82.83 & 1315 & 398 & 1102 & 185 & 80.57
 \\
     &  & \textbf{\Ours} & 1435 & 352 & 1148 & 65 & 86.10 & 1327 & 281 & 1219 & 173 & 84.87
 \\
     % \arrayrulecolor{gray!50}\cmidrule(lr){3-13}
     % &  & \textbf{\Ours}+VCD & 1435 & 354 & 1146 & 65 & 86.03 & 1334 & 285 & 1215 & 166 & 84.97 \\
     % &  & \textbf{\Ours}+M3ID & 1433 & 344 & 1156 & 67 & 86.30 & 1322 & 272 & 1228 & 178 & 85.00 \\
     \arrayrulecolor{gray}\cmidrule(lr){2-13}
      & \multirow{4}{*}{Popular} & \textit{base} & 1402 & 727 & 773 & 98 & 72.50 & 1281 & 599 & 901 & 219 & 72.73
 \\
     &  & VCD & 1422 & 775 & 725 & 78 & 71.57 & 1298 & 588 & 912 & 202 & 73.67
 \\
     &  & M3ID & 1410 & 725 & 775 & 90 & 72.83 & 1316 & 579 & 921 & 184 & 74.57
 \\
     &  & \textbf{\Ours} & 1435 & 691 & 809 & 65 & 74.80 & 1326 & 591 & 909 & 174 & 74.50
 \\
     % \arrayrulecolor{gray!50}\cmidrule(lr){3-13}
     % &  & \textbf{\Ours}+VCD & 1431 & 679 & 821 & 69 & 75.07 & 1331 & 571 & 929 & 169 & 75.33 \\
     % &  & \textbf{\Ours}+M3ID & 1433 & 701 & 799 & 67 & 74.40 & 1331 & 564 & 936 & 169 & 75.57 \\
     \arrayrulecolor{gray}\cmidrule(lr){2-13}
      & \multirow{4}{*}{Adversarial} & \textit{base} & 1397 & 868 & 632 & 103 & 67.63 & 1285 & 698 & 802 & 215 & 69.57
 \\
     &  & VCD & 1413 & 889 & 611 & 87 & 67.47 & 1279 & 696 & 804 & 221 & 69.43
 \\
     &  & M3ID & 1417 & 873 & 627 & 83 & 68.13 & 1292 & 725 & 775 & 208 & 68.90
 \\
     &  & \textbf{\Ours} & 1437 & 890 & 610 & 63 & 68.23 & 1327 & 722 & 778 & 173 & 70.17
 \\
     % \arrayrulecolor{gray!50}\cmidrule(lr){3-13}
     % &  & \textbf{\Ours}+VCD & 1435 & 865 & 635 & 65 & 69.00 & 1328 & 721 & 779 & 172 & 70.23 \\
     % &  & \textbf{\Ours}+M3ID & 1429 & 865 & 635 & 71 & 68.80 & 1343 & 713 & 787 & 157 & 71.00 \\
    \bottomrule
    \end{tabular}
    }
    % \vspace{\abovetabcapmargin}
    % \vspace{\belowtabcapmargin}
    \label{tab:confusion_full}
\end{table*}

To analyze the performance of the model in detail, we report the confusion matrices in \Cref{tab:confusion_full} for the POPE benchmark. Notably, \Ours significantly improves True Negatives (TN) while maintaining a similar level of True Positives (TP) compared to existing contrastive decoding methods.
It implies that our method achieves the highest accuracy by significantly improving the identification of non-relevant instances compared to the baseline and previous methods.

\begin{figure*}[t!]
    \centering
    \begin{minipage}{\textwidth}
        \centering
        \includegraphics[width=\textwidth]{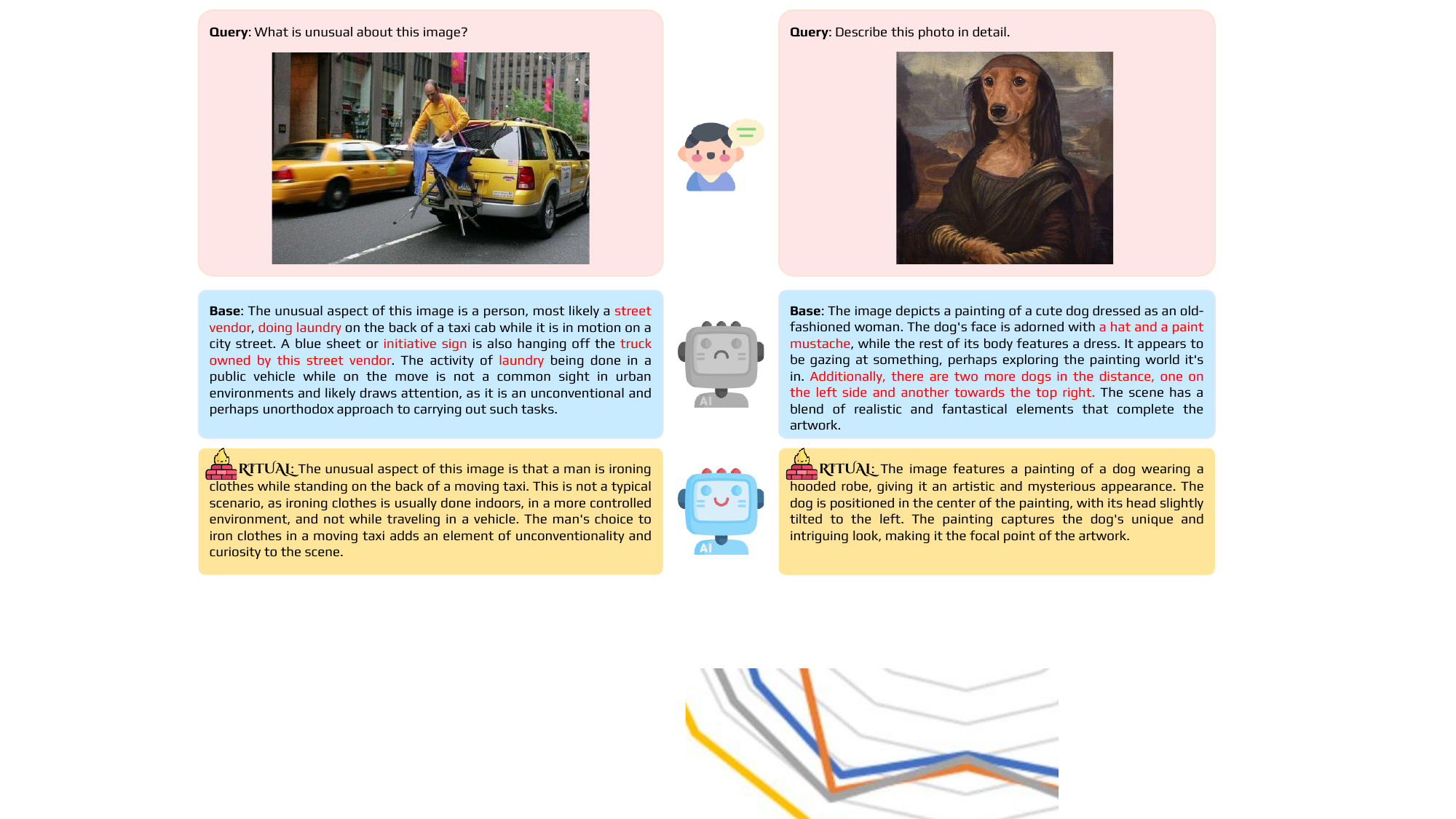}
        \vspace{\abovefigcapmargin}
        \caption{
        \textbf{Qualitative results on LLaVA-Bench~\citep{liu2023visual}.}
        Hallucinations are highlighted in {\color{red}{red}}.
        \Ours well understands ambiguous images and effectively mitigates hallucinations in outputs.
        }%
        \label{fig:llava_bench}
    \end{minipage}
    \begin{minipage}{0.88\textwidth}
        \centering
        \includegraphics[width=\textwidth]{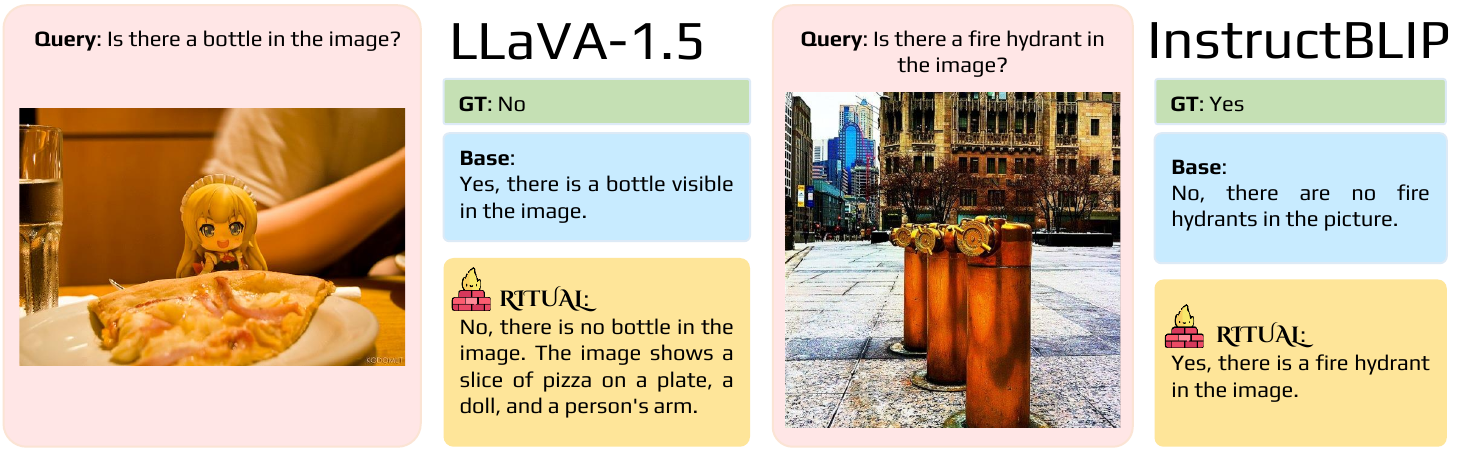}
        \vspace{\abovefigcapmargin}
        \caption{
        \textbf{Qualitative results on POPE~\citep{li2023evaluating}.}
        }%
        \label{fig:appendix_pope}
    \end{minipage}
    \begin{minipage}{0.88\textwidth}
        \centering
        \includegraphics[width=\textwidth]{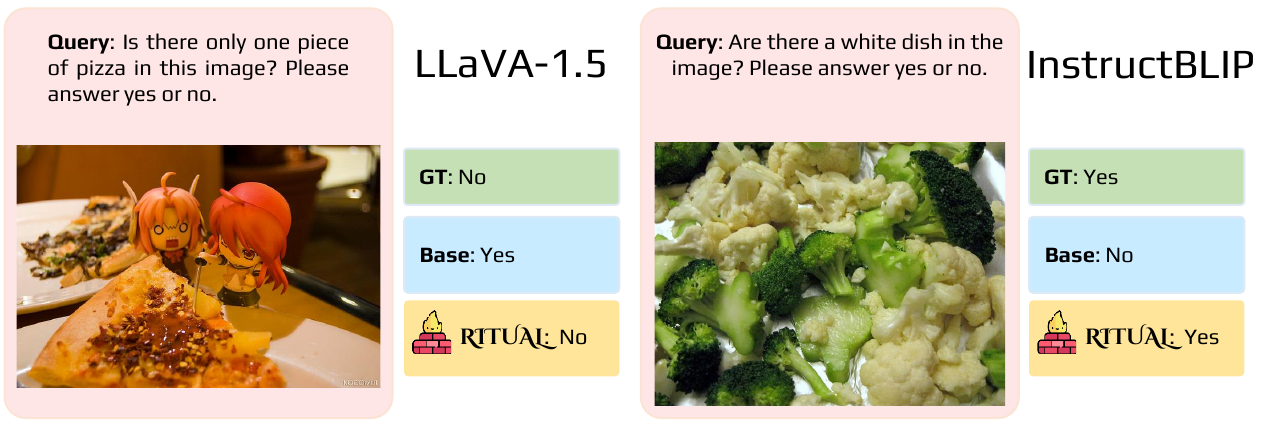}
        \vspace{\abovefigcapmargin}
        \caption{
        \textbf{Qualitative results on MME~\citep{fu2024mme}.}
        }%
        \label{fig:appendix_mme}
    \end{minipage}
\end{figure*}

\begin{figure*}[t!]
    \centering
    \includegraphics[width=\textwidth]{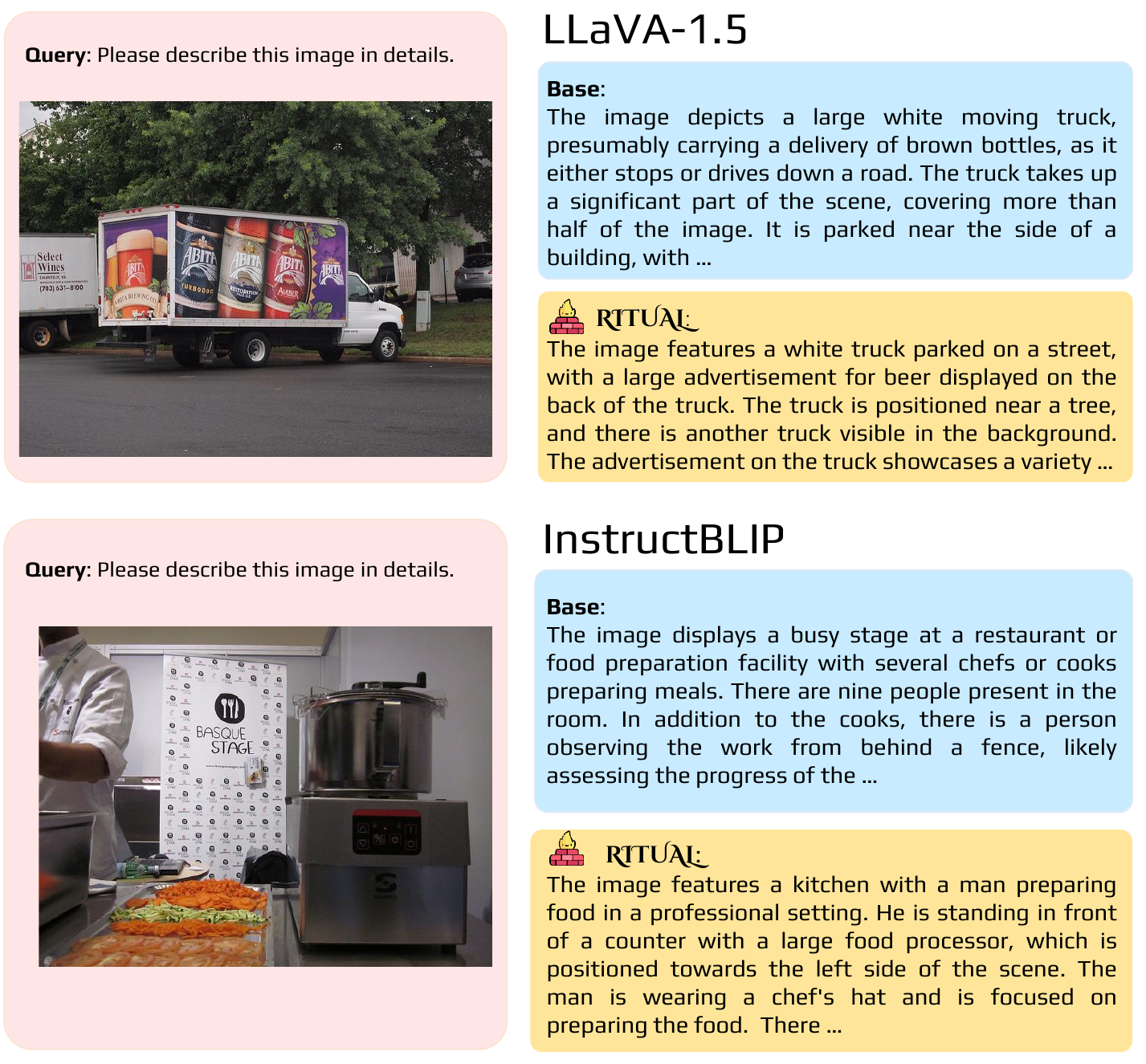}
    \vspace{\abovefigcapmargin}
    \caption{
    \textbf{Qualitative results on CHAIR~\citep{rohrbach2018object}.}
    }%
    \label{fig:appendix_chair}
    \vspace{\belowfigcapmargin}
\end{figure*}

\begin{figure*}[t!]
    \centering
    \includegraphics[width=\textwidth]{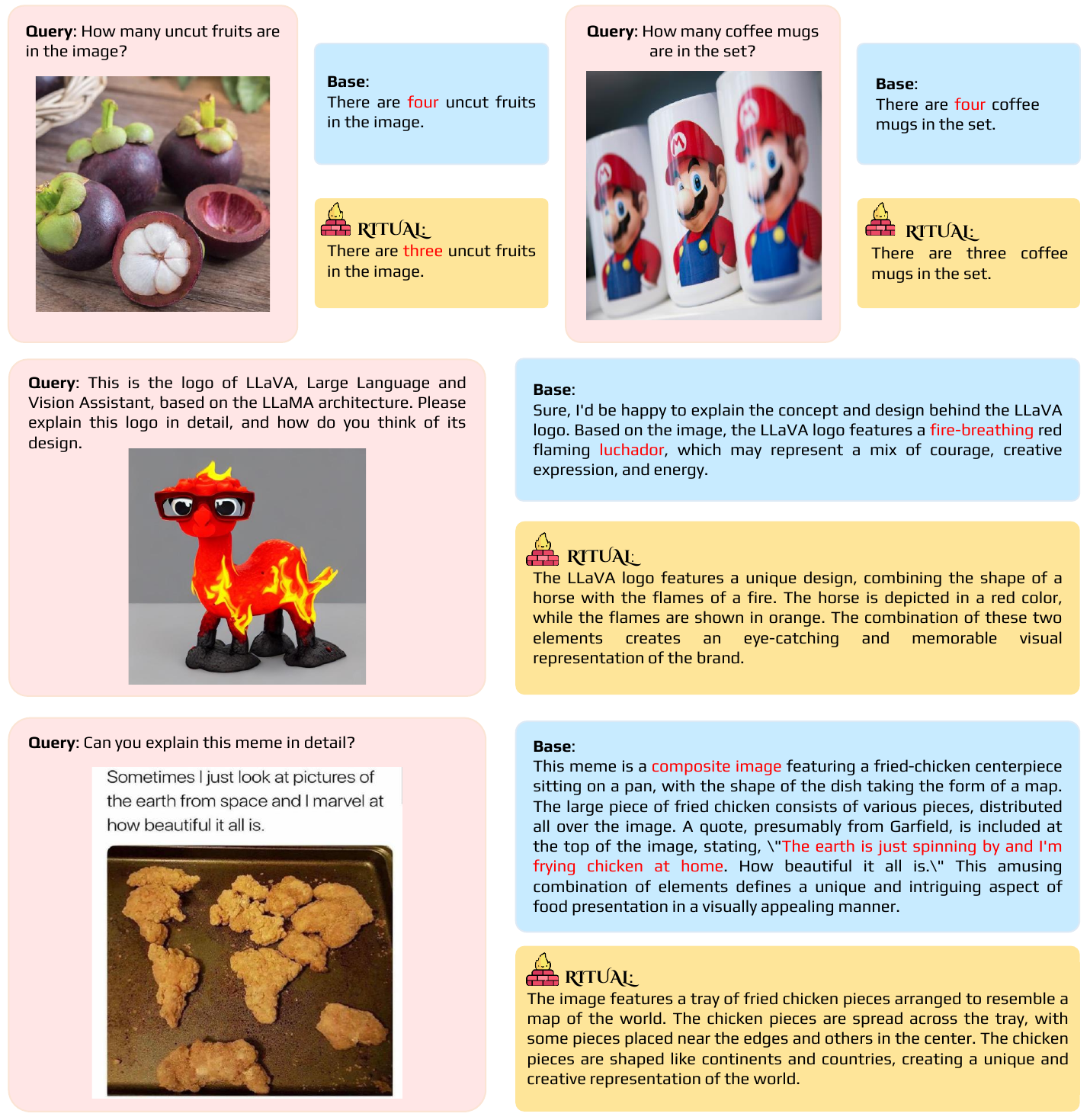}
    \vspace{\abovefigcapmargin}
    \caption{
    \textbf{Qualitative results on LLaVA-Bench~\citep{liu2023visual}.}
    Hallucinations are highlighted in \color{red}{red}.
    }%
    \label{fig:appendix_llava_bench}
    \vspace{\belowfigcapmargin}
\end{figure*}
\subsection{Qualitative Examples}
\label{sec:appendix_qualitative_examples}
We provide additional qualitative examples on POPE~\cite{li2023evaluating}, MME~\cite{fu2024mme}, CHAIR~\cite{rohrbach2018object}, and LLaVA-Bench~\cite{liu2023visual} in~\cref{fig:llava_bench,fig:appendix_pope,fig:appendix_mme,fig:appendix_chair,fig:appendix_llava_bench}.

\Cref{fig:llava_bench} presents two samples from the LLaVA-Bench~\cite{liu2023visual} with LLaVa-1.5~\cite{liu2023visual}, highlighting the differences between sentences generated by standard decoding (Base) and those produced by \Ours.
The results demonstrate that standard decoding often results in hallucinations, which can be effectively rectified by implementing \Ours.
For instance, in the left-hand image, the baseline model incorrectly identifies a `street vendor' and `initiative signs', neither of which are present in the image.
Additionally, it misinterprets `ironing' as `doing laundry'.
In the right-hand image, the baseline model hallucinates objects not present in the image, such as a `hat', `paint mustache', and `two more dogs'.
In contrast, our approach helps counteract these hallucinations, generating sentences that reflect a more accurate comprehension of the image.

% \section{License of Assets}
% \label{sec:appendix_license}
% POPE~\cite{li2023evaluating} is licensed under MIT License.
% CHAIR~\cite{rohrbach2018object} is made available under the BSD 2-Clause License.
% LLaVA-Bench is available under Apache-2.0 License.
% LLaVA~\cite{liu2023visual} is licensed under the Apache-2.0 License.
% InstructBLIP~\cite{dai2024instructblip} is under BSD-3-Clause License.
% PyTorch~\cite{paszke2019pytorch} is released under the Modified BSD License.

\section{Limitations}
\label{sec:appendix_limitations}
\Ours is a simple yet effective technique that improves model robustness against hallucinations.
However, it comes with the following limitations:

\begin{itemize}
    \item \textbf{Computational overhead}: \Ours necessitates running the model twice for each test image, resulting in higher inference time and computational demands. This can pose challenges in real-time or resource-constrained scenarios.
    \item \textbf{Diminishing returns}: Although \Ours offers noticeable performance gains, its benefits taper off with excessive or redundant transformations, which may introduce unnecessary complexity without significant improvements.
    % \item \textbf{Task-specific transformations}: The selection of transformations must align with the task at hand. For example, while horizontal flips are beneficial for natural image classification, they may introduce unrealistic artifacts in medical imaging or charts (\eg, flipping a chest X-ray).
    % \item \textbf{May not always effective for all models}: Some models, especially highly regularized or robust models, may not benefit as much from \Ours if they are already trained on extensive data augmentations during training.
\end{itemize}

 \fi

\end{document}